\documentclass[acmsmall]{acmart}
\usepackage{booktabs} 
\usepackage[ruled]{algorithm2e} 

\usepackage{url}
\usepackage{graphicx}
\usepackage{amsfonts}
\usepackage{amssymb}
\usepackage{amsthm}
\usepackage{amsmath}
\usepackage{color}
\usepackage{amsmath}
\usepackage{caption}
\usepackage[mathscr]{eucal}
\usepackage{multirow}
\usepackage{algorithmic}
\usepackage{epstopdf}
\usepackage{diagbox}
\usepackage{subfigure} 
\usepackage{enumitem}

\newtheorem{remark}{Remark}

\definecolor{ao}{rgb}{0.0, 0.5, 0.0}

\newcommand{\R}{\mathbb{R}}
\newcommand{\E}{\mathbb{E}}
\newcommand{\Hs}{\mathcal{H}}
\def\mL{\mathcal L}

\newcommand{\errapp}{\mathcal{E}_{\text{app}}}
\newcommand{\errest}{\mathcal{E}_{\text{est}}}

\newcommand{\X}{\mathcal{X}}

\newcommand{\Z}{\mathcal{Z}} 
\newcommand{\T}{T}
\newcommand{\D}{{D}}

\newcommand{\xtest}{x_{\text{test}}}

\newcommand{\dtrain}{{D}_{\text{train}}}

\newcommand{\dtest}{D_{\text{test}}}
\newcommand{\xii}{x_{i}}
\newcommand{\yii}{y_{i}}
\newcommand{\xyi}{(x_{i},y_{i})}

\usepackage{array}
\newcolumntype{L}[1]{>{\raggedright\let\newline\\\arraybackslash\hspace{0pt}}m{#1}}
\newcolumntype{C}[1]{>{\centering\let\newline \\\arraybackslash\hspace{0pt}}m{#1}}
\newcolumntype{R}[1]{>{\raggedleft\let\newline \\\arraybackslash\hspace{0pt}}m{#1}}
\SetAlFnt{\small}
\SetAlCapFnt{\small}
\SetAlCapNameFnt{\small}
\SetAlCapHSkip{0pt}
\IncMargin{-\parindent}

\begin{document}
\title[Generalizing from a Few Examples: A Survey on Few-Shot Learning]{Generalizing from a Few Examples: A Survey on Few-Shot Learning} 
\author{Yaqing Wang}
\email{ywangcy@connect.ust.hk}
\orcid{0000-0003-1457-1114}
\affiliation{%
	\department{Department of Computer Science and Engineering}
	\institution{Hong Kong University of Science and Technology}
}
\affiliation{%
\department{Business Intelligence Lab and National Engineering Laboratory of Deep Learning Technology and Application}
\institution{Baidu Research}	
}
\author{Quanming Yao}
\authornote{Corresponding Author}
\email{yaoquanming@4paradigm.com}
\orcid{0000-0001-8944-8618}
\affiliation{%
	\institution{4Paradigm Inc.}
}

\author{James T. Kwok}
\email{jamesk@cse.ust.hk}
\orcid{0000-0002-4828-8248}
\affiliation{%
	\department{Department of Computer Science and Engineering}
	\institution{Hong Kong University of Science and Technology}
}
\author{Lionel M. Ni}
\email{ni@ust.hk}
\orcid{0000-0002-2325-6215}
\affiliation{%
	\department{Department of Computer Science and Engineering}
	\institution{Hong Kong University of Science and Technology}
}
\begin{abstract}
Machine learning has been highly successful in data-intensive applications, but is
often hampered when the data set is small.
Recently, 
Few-Shot Learning (FSL) is proposed
to tackle this problem.
Using prior knowledge,
FSL can rapidly generalize to new tasks containing only a few samples with supervised information.
In this paper, we conduct a thorough survey 
to fully understand FSL.
Starting from a formal definition of FSL,
we distinguish FSL from several relevant machine learning problems. 
We then point out that 
the core issue in FSL is
that the empirical risk minimizer
is unreliable.
Based on how prior knowledge can be used to handle this core issue, 
we categorize FSL methods 
from three perspectives: 
(i) data, which uses prior knowledge to augment the supervised experience;
(ii) model, which uses prior knowledge to reduce the size of the hypothesis space; 
and (iii) algorithm, which uses prior knowledge to alter the search for the best
hypothesis in the given hypothesis space. 
With this taxonomy, we  review and 
discuss the pros and cons of each category.
Promising directions, in the aspects of the FSL problem setups, techniques, applications and theories, 
are also proposed
to provide insights for future research.\footnote{A list of references,
	which will be updated periodically, 
	can be found at
	\textit{\url{https://github.com/tata1661/FewShotPapers.git}}.}
\end{abstract}





\setcopyright{acmcopyright}
\acmJournal{CSUR}
\acmYear{2020} \acmVolume{1} \acmNumber{1} \acmArticle{1} \acmMonth{3} \acmPrice{15.00}\acmDOI{10.1145/3386252}

\maketitle

\section{Introduction}

``Can machines think?'' This is the question raised in Alan Turing's seminal paper entitled ``Computing Machinery and Intelligence'' 
\cite{turing1950computing}
in 1950. 
He made the statement that
``The idea behind digital computers may be explained by saying that these machines are
intended to carry out any operations which could be done by a human computer''.
In other words, the ultimate goal of machines is to be as intelligent as humans. 
In recent years, due to the emergence of powerful computing devices (e.g.,
GPU and distributed platforms), large data sets (e.g., ImageNet data
with 1000 classes
\cite{deng2009imagenet}), advanced models and algorithms (e.g., convolutional
neural networks (CNN) \cite{krizhevsky2012imagenet} and long short-term memory (LSTM) \cite{hochreiter1997long}), AI speeds up its pace to be like humans and defeats humans in many fields.
To name a few, 
AlphaGo \cite{silver2016mastering} defeats human champions in the ancient game of 
Go; and residual network (ResNet) \cite{he2016deep} obtains 
better classification performance than humans on ImageNet.
AI 
also supports the development of intelligent tools
in many aspects
of daily life,
such as voice assistants, 
search engines, 
autonomous driving cars, and industrial robots.

Albeit its prosperity, current AI techniques cannot rapidly generalize from a few examples.
The aforementioned successful AI applications rely on learning from large-scale data. 
In contrast, 
humans are capable of learning new tasks
rapidly 
by utilizing what they learned in the past.
For example, a child who learned how to add can rapidly transfer his knowledge to learn 
multiplication given a few examples (e.g., $2\times3=2+2+2$ and $1\times3=1+1+1$).
Another example is that given a few photos of a stranger, a child can easily identify the same person from a large number of photos.

Bridging this gap between AI and humans is an important direction.
It can be tackled by \textit{machine learning}, 
which is concerned with the question of how to construct computer programs that automatically improve with experience \cite{tom1997machine,mohri2018foundations}.
In order to learn from a limited number of examples with supervised information,  
a new machine learning paradigm called \textit{Few-Shot Learning} (FSL) \cite{fink2005object,fei2006one} is proposed. 
A typical example is character generation \cite{lake2015human}, in which computer
programs are asked to parse and generate 
new 
handwritten characters given a few examples. 
To handle this task, one can decompose
the characters into smaller parts transferable across characters, and then aggregate these smaller components into new characters. 
This is a way of learning like human \cite{lake2017building}. 
Naturally, FSL can also advance robotics \cite{craig2009introduction},
which develops machines that can replicate human actions. 
Examples include one-shot imitation 
\cite{wu2010towards}, 
multi-armed bandits \cite{duan2017one}, 
visual navigation \cite{finn2017model}, and continuous control \cite{yoon2018bayesian}.

Another classic FSL scenario is where examples with
supervised information are hard or impossible to acquire due to 
privacy, safety or ethic issues.
A typical example is drug discovery, which tries to discover properties of new molecules so as to identify useful ones 
as new drugs \cite{altae2017low}.
Due to possible toxicity, low activity, and low solubility, new molecules do not have many real biological records on clinical candidates.
Hence, it is important to learn effectively from a small  number of samples.
Similar examples where the target tasks do not have many examples include
FSL translation \cite{kaiser2017learning},  and
cold-start item recommendation \cite{vartak2017meta}. 
Through FSL, learning suitable models for these rare cases can become possible.

FSL can also help relieve the burden of collecting 
large-scale supervised data.
For example, although
ResNet \cite{he2016deep} outperforms humans on ImageNet,
each class needs to have sufficient labeled images 
which can be laborious
to collect. 
FSL can reduce the data gathering effort for data-intensive
applications. Examples include image classification \cite{vinyals2016matching}, image retrieval \cite{triantafillou2017few}, 
object tracking \cite{bertinetto2016learning}, 
gesture recognition \cite{pfister2014domain}, 
image captioning, visual question answering \cite{dong2018fast}, 
video event detection \cite{yan2015multi}, 
language modeling \cite{vinyals2016matching}, 
and  
neural architecture search \cite{brock2018smash}.

Driven by the academic goal for AI to approach humans and the industrial demand for
inexpensive learning, FSL
has drawn much recent attention and is now a hot topic. 
Many related machine learning approaches have been proposed, 
such as meta-learning 
\cite{santoro2016meta,finn2017model,ravi2017optimization}, embedding learning \cite{vinyals2016matching,bertinetto2016learning,sung2018learning} and generative modeling \cite{fei2006one,salakhutdinov2012one,edwards2017towards}.
However, currently, there is no work that provides an organized taxonomy to connect these 
FSL methods, explains why some methods work while others fail, nor discusses the
pros and cons of different approaches.
Therefore, in this paper, we conduct a survey on the FSL problem. 
In contrast,
the survey in \cite{shu2018small} only 
focuses on concept learning and experience learning for small samples. 

Contributions of this survey can be summarized as follows: 
\begin{itemize}
\item We give a formal definition on FSL, which naturally connects to the classic machine learning definition  in \cite{tom1997machine,mohri2018foundations}.
The definition is not only general enough to include existing FSL works,
but also specific enough to clarify what the goal of FSL is and how we can solve it.
This definition is helpful for setting future research targets in the FSL area.

\item We list the relevant learning problems for FSL with concrete examples,
clarifying their relatedness and differences with respect to FSL. These discussions can help better discriminate and position FSL among various learning problems.

\item We point out that the 
core issue of FSL supervised learning problem is the unreliable empirical risk minimizer, which is analyzed 
based on
error decomposition \cite{bottou2008tradeoffs} in machine learning.  
This provides insights to improve FSL methods in a more organized and systematic way.

\item We perform an extensive literature review, 
and organize them in an unified taxonomy from the perspectives of data, model and algorithm. 
We also present a summary of insights and a discussion on the pros and cons
of each category.
These can help establish a better understanding of FSL methods.

\item We propose promising future directions for FSL in the aspects of problem setup, techniques, applications and theories.
These insights are based on the weaknesses of the current development of FSL, with possible improvements to make in the future.
\end{itemize} 

\subsection{Organization of the Survey}
The remainder of this survey is organized as follows.
Section~\ref{sec:review} provides an overview for FSL,
including its formal definition, relevant learning problems, core issue,  
and a taxonomy of existing works in terms of data, model and algorithm.
Section~\ref{sec:data} is for methods that augment data to solve FSL problem.
Section~\ref{sec:model} is for methods that reduce the size of hypothesis space so as to make FSL feasible.
Section~\ref{sec:alg} is for methods that alter the search strategy of algorithm to
deal with the FSL problem.
In Section~\ref{sec:future}, we propose future directions for FSL in terms of problem setup, techniques, applications and theories.
Finally, the survey closes with conclusion in Section~\ref{sec:conclusion}.

\subsection{Notation and Terminology}
\label{sec:overview_notation}
Consider a learning task $\T$, FSL deals with
a data set 
$\D = \{\dtrain,\dtest\}$ consisting of a
training set
$\dtrain=\{\xyi\}_{i=1}^{I}$ where $I$ is small, and 
a testing set 
$\dtest=\{x^{\text{test}}\}$. 
Let $p(x,y)$ be the ground-truth joint probability distribution of input $x$ and output $y$, and 
$\hat{h}$ be the optimal hypothesis from $x$ to $y$.
FSL learns to discover $\hat{h}$ by fitting $\dtrain$ and testing on $\dtest$.
To approximate $\hat{h}$, the FSL
model
determines a hypothesis space $\Hs$ of hypotheses $h(\cdot;\theta)$'s, where 
$\theta$ denotes all the parameters used by $h$. 
Here, a parametric $h$ is used, as a nonparametric
model often requires large data sets, and thus not suitable for FSL.  
A FSL algorithm 
is an optimization strategy that searches $\Hs$ in order to find the $\theta$ that
parameterizes the best $h^*\in\Hs$.  
The FSL performance is measured by a loss function $\ell(\hat{y},y)$ defined over
the prediction $\hat{y}=h(x;\theta)$ and the observed output $y$. 

\section{Overview}
\label{sec:review}
In this section, we first provide a formal definition of the FSL problem in Section~\ref{sec:overview_problem_define} with concrete examples. 
To differentiate the FSL problem from relevant machine learning problems, we discuss their
relatedness and differences in Section~\ref{sec:overview_relevant_topic}.
In Section~\ref{sec:overview_core_issue}, we discuss the core issue that makes 
FSL difficult.
Section~\ref{sec:overview_taxonomy}
then presents a unified taxonomy 
according to how existing works handle the core issue.

\subsection{Problem Definition}
\label{sec:overview_problem_define}

As 
FSL is a sub-area in machine learning,
before giving the definition of FSL,
let us recall how machine learning is defined in the literature.

\begin{definition}[\textbf{Machine Learning} \cite{tom1997machine,mohri2018foundations}] 
	\label{def:machine_learn}
	A computer program is said to learn from experience $E$ with respect to some classes of task $T$ 
	and performance measure $P$ 
	if its performance can improve with $E$ on $T$ measured by $P$.
\end{definition}

For example, consider an image classification task ($T$), 
a machine learning program can improve its classification accuracy ($P$) through
$E$ obtained by training on a large number of labeled images (e.g., the ImageNet
data set \cite{krizhevsky2012imagenet}).
Another example is the recent computer program
AlphaGo \cite{silver2016mastering}, which has defeated the human champion in playing the ancient game of 
Go ($T$). 
It improves its winning rate ($P$) against opponents by training on a database ($E$) of around 30 million 
recorded moves of human experts as well as playing against itself repeatedly. 
These are summarized in Table~\ref{tab:ml_example}.

\begin{table}[ht]
	\caption{Examples of machine learning problems
		based on Definition~\ref{def:machine_learn}.}
	\footnotesize
	\begin{tabular}
		{C{100px} | C{180px} | C{65px} }
		\hline
		\multirow{1}{*}{task $T$}&      \multicolumn{1}{c|}{experience $E$}       &\multirow{1}{*}{performance $P$} \\ \hline
		image classification \cite{krizhevsky2012imagenet} &    large-scale labeled images for each class &    classification accuracy                                     \\ \hline
		the ancient game of Go \cite{silver2016mastering} &   a database containing around 30 million 
		recorded moves of human experts   and self-play records    &      winning rate                                  \\ \hline
	\end{tabular}
	\label{tab:ml_example}
\end{table}

Typical machine learning applications, as in the examples mentioned 
above,
require 
a lot of examples with supervised information.
However,
as mentioned in the introduction, 
this may be difficult or even not possible.
FSL is a special case of machine learning,
which targets at obtaining good learning performance 
given limited supervised information provided in the 
training set $\dtrain$, which consists of examples of inputs $\xii$'s along with their corresponding output $\yii$'s \cite{bishop2006pattern}. 
Formally, we define FSL in Definition~\ref{def:fsl}.

\begin{definition}	
	\label{def:fsl}
	\textbf{\textit{Few-Shot Learning}} (FSL) is a type of machine learning problems 
	(specified by $E$, $T$ and $P$), where $E$ contains only a limited number of examples with supervised information for the target $T$. 
\end{definition}

Existing FSL problems are mainly supervised learning problems. Concretely,
\textit{few-shot classification} 
learns classifiers given only a few labeled examples of each class. 
Example applications include image classification \cite{vinyals2016matching}, sentiment classification from short text \cite{yu2018diverse} and object recognition \cite{fei2006one}. 
Formally, using notations from Section~\ref{sec:overview_notation}, 
\textit{few-shot classification} learns a classifier $h$ which predicts label $\yii$ for each input $\xii$.
Usually, one considers the $N$-way-$K$-shot classification
\cite{vinyals2016matching,finn2017model}, in which
$\dtrain$ contains $I=KN$ examples 
from $N$ classes each with $K$ examples. 
\textit{Few-shot regression} 
\cite{finn2017model,yoon2018bayesian} estimates a regression function $h$ given
only a few input-output example pairs sampled from that function, where output
$y_i$ is the observed value of the dependent variable $y$, and 
$x_i$ is the input which records the observed value of the independent variable $x$. 
Apart from few-shot supervised learning, another instantiation of FSL is
\textit{few-shot reinforcement learning} \cite{duan2017one,al2018continuous}, which targets at 
finding a policy given only a few trajectories consisting of state-action pairs. 

We now show three typical scenarios of FSL (Table~\ref{tab:fsl_example}):
\begin{itemize}
	\item \textit{Acting as a test bed for learning like human}. 
	To move towards human intelligence, it is vital that 
	computer programs can solve the FSL problem. 
	A popular task ($T$) is to generate samples of a new character given only 
	a few examples \cite{lake2015human}. 
	Inspired by how humans learn, 
	the computer programs learn 
	with the $E$ consisting of both 
	the given examples 
	with supervised information
	and pre-trained concepts such as parts and relations as prior knowledge. 
	The generated characters are evaluated through the pass rate of visual Turing test ($P$), which discriminates whether the images are generated by humans or machines.
	With this prior knowledge, 
	computer programs can also learn to classify, parse and generate new 
	handwritten characters with a few examples like humans.

	\item 
	\textit{Learning for rare cases}. 
	When obtaining sufficient examples with supervised information is hard or impossible,
	FSL can learn models for the rare cases.
	For example, consider a drug discovery task ($T$) which tries to predict whether a
	new molecule has toxic effects \cite{altae2017low}.  The percentage of molecules
	correctly assigned as toxic or non-toxic ($P$) improves with $E$ obtained by both the new molecule's limited assay, 
	and many similar molecules' assays as prior knowledge. 
	
	\item 
	\textit{Reducing data gathering effort and computational cost}. 
	FSL can help relieve the burden of collecting large number of examples with supervised information. 
	Consider few-shot image classification task ($T$) \cite{fei2006one}.
	The image classification accuracy ($P$) improves with the $E$ obtained by a few labeled images for each class of the target $T$, and 
	prior knowledge extracted from the other classes (such as 
	raw images to co-training). 
	Methods succeed in this task usually have higher generality. Therefore, they can be easily applied for tasks of many samples.
\end{itemize}

\begin{table}[ht]
	\caption{Three FSL examples
		based on Definition~\ref{def:fsl}.}
	\footnotesize
	\begin{tabular}{C{100px} | C{90px} | C{90px} | C{65px} }
		\hline
		\multirow{2}{*}{task $T$}&      \multicolumn{2}{c|}{experience $E$}       & \multicolumn{1}{c}{\multirow{2}{*}{performance $P$}} \\ \cline{2-3}
		& supervised information & prior knowledge & \multicolumn{1}{c}{}                \\ \hline
		character generation \cite{lake2015human} &    a few examples of new character        &   pre-learned knowledge of parts and relations       &      pass rate of visual Turing test                                   \\ \hline
		drug toxicity discovery \cite{altae2017low}   &   new molecule's limited assay         &  similar molecules' assays        &    classification accuracy                                 \\ \hline
				image classification \cite{koch2015siamese} &    
		a few labeled images for each class of the target $T$
		&   raw images of other classes, 
		or pre-trained models         &    classification accuracy                                     \\ \hline
	\end{tabular}
	\label{tab:fsl_example}
\end{table}

In comparison to Table~\ref{tab:ml_example}, Table~\ref{tab:fsl_example} has one
extra column under ``experience $E$" which is marked as ``prior knowledge".  
As  $E$ only contains a few examples with supervised information directly related to $T$,
it is natural that common supervised learning approaches often fail on FSL problems. 
Therefore, FSL methods make the learning of target $T$ feasible
by combining 
the available supervised information in $E$ with some prior knowledge, which is ``any information the learner has about the unknown function before seeing the examples" \cite{mahadevan1994quantifying}. 
One typical type of FSL methods is
Bayesian learning \cite{lake2015human,fei2006one}. 
It combines the provided training set $\dtrain$
with some prior probability distribution which is available before $\dtrain$ is given \cite{bishop2006pattern}.

\begin{remark}
	When there is only one example with supervised information in $E$, 
	FSL is called \textbf{\textit{one-shot learning}} \cite{fei2006one,vinyals2016matching,bertinetto2016learning}. 
	When $E$ does not contain any example with supervised information for the target
	$T$, FSL becomes a \textbf{\textit{zero-shot learning}} problem (ZSL)
	\cite{lampert2009learning}. 
	As the target class does not contain examples with supervised information, ZSL
	requires $E$ to contain information from other modalities (such as attributes,
	WordNet, 
	and	word embeddings used in rare object recognition tasks), 
	so as to transfer some supervised information 
	and make learning possible. 
\end{remark}

\subsection{Relevant Learning Problems}
\label{sec:overview_relevant_topic}
In this section, we discuss some relevant machine learning problems. 
The relatedness and difference with respect to FSL are clarified.

\begin{itemize}
	\item \emph{Weakly supervised learning}	\cite{zhou2017brief} learns from
	experience $E$ containing only weak supervision (such as incomplete, inexact,
	inaccurate or noisy supervised information).
	The most relevant problem to FSL is \emph{weakly supervised learning with
		incomplete supervision} where only a small amount of samples have supervised information.  
	According to whether the oracle or human intervention is leveraged, this can be
	further classified into  the following:
	
	\begin{itemize}
\item  
\emph{Semi-supervised learning} \cite{zhu2005semi}, which learns from a small
number of labeled samples and (usually a large number of) unlabeled samples in $E$.
		Example applications are text and webpage classification.  
		\emph{Positive-unlabeled learning} \cite{li2009positive} is a special case of
		semi-supervised learning, in which only positive and unlabeled samples are given. 
		For example, to recommend friends in social networks, 
		we only know the users' current friends according to the friend list, while
		their relationships to other people are unknown.
		
		\item \emph{Active learning} \cite{settles2009active}, which 
		selects informative unlabeled data to query an oracle for output $y$. 
		This is usually used for applications where annotation labels are 
		costly, such as 
		pedestrian detection. 
	\end{itemize}
	
	By definition, \emph{weakly supervised learning with
		incomplete supervision} includes only classification and regression, while
	FSL also includes reinforcement learning problems. 
	Moreover, \emph{weakly supervised learning with
		incomplete supervision} mainly uses unlabeled data as additional information
	in $E$, while FSL leverages various kinds of prior knowledge such as
	pre-trained models, supervised data from other domains or modalities and does
	not restrict to using unlabeled data. 
	Therefore, FSL becomes weakly supervised learning problem only when prior knowledge is unlabeled data and the task is classification or regression.

	\item \emph{Imbalanced learning} \cite{he2008learning} learns from experience $E$ with a skewed distribution for $y$. 
	This happens when some values of $y$ are rarely taken, as in fraud detection and
	catastrophe anticipation applications.  
	It trains and tests to choose among all possible $y$'s.
	In contrast, FSL trains and tests for $y$ with a few examples, while possibly
	taking the other $y$'s as prior knowledge for learning.
	
	\item \emph{Transfer learning} \cite{pan2010survey} transfers knowledge 
	from the source domain/task, where  training data is abundant,
	to the target domain/task, where training data is scarce. 
	It can be used in applications such as
	cross-domain recommendation, WiFi localization 
	across time periods, space and mobile devices.
	\emph{Domain adaptation} \cite{ben2007analysis} is a type of transfer learning
	in which the source/target tasks are the same but the source/target domains are different. 
	For example, in
	sentiment analysis,
	the source domain data contains customer comments on movies, while the
	target domain data contains customer comments on daily goods. 
	Transfer learning methods are popularly used 
	in FSL \cite{luo2017label,azadi2018multi,liu2018feature}, where the prior knowledge is transferred from the source task to the few-shot task.

	\item 
	\emph{Meta-learning} \cite{hochreiter2001learning} 
	improves $P$ of the new task $\T$ by the provided data set and the meta-knowledge extracted across tasks by a meta-learner.
	Specifically, 
	the meta-learner gradually learns generic information (meta-knowledge) across tasks,
	and the
	learner generalizes the meta-learner for a new task $\T$ using task-specific information. 
	It has been 
	successfully applied in
	problems such as learning optimizers
	\cite{li2017learning,andrychowicz2016learning}, dealing with the cold-start problem
	in collaborative filtering \cite{vartak2017meta}, and guiding policies by natural language \cite{co-reyes2018metalearning}. 
	Meta-learning methods can be used to deal with the FSL problem. 
	As will be shown in Sections~\ref{sec:model} and ~\ref{sec:alg}, the meta-learner is taken as prior knowledge to guide each specific FSL task. 	
	A formal definition of meta-learning and how it is used for the FSL problem are provided in Appendix~\ref{app:meta}. 
	
\end{itemize}

\subsection{Core Issue}
\label{sec:overview_core_issue}

In any machine learning problem,
usually there are prediction errors and
one cannot obtain perfect predictions.
In this section, 
we illustrate the core issue of FSL based on error decomposition
in supervised machine learning \cite{bottou2008tradeoffs,bottou2018optimization}. 
This analysis applies to FSL supervised learning including classification and
regression, and can also provide insights for understanding FSL reinforcement learning.  

\subsubsection{Empirical Risk Minimization}
\label{sssec:errcomp}

Given a hypothesis $h$, 
we want to minimize its \emph{expected risk} $R$,
which is the loss measured 
with respect to $p(x,y)$. 
Specifically, 
\begin{align*}
R(h) = \int \ell(h(x),y)\ dp(x,y)=\E[\ell(h(x),y)].
\end{align*} 
As $p(x,y)$ is unknown,
the
\emph{empirical risk} 
(which is the average of sample losses over the training set $\dtrain$ of $I$ samples)
\begin{align*}
R_I(h) = \frac{1}{I} \sum_{i=1}^I \ell(h(\xii),\yii),
\end{align*}
is usually used as a proxy for $R(h)$,
leading to \textit{empirical risk minimization}
\cite{vapnik1992principles,mohri2018foundations} (with possibly some regularizers).
For illustration, let
\begin{itemize}
\item $\hat{h}
	=\arg\min_h R(h)$ 
	be the function that minimizes the expected risk;
	\item $h^*=\arg\min_{h\in\Hs} R(h)$
	be the function in $\Hs$ that minimizes the expected risk; 
	\item $h_I=\arg\min_{h\in\Hs} R_I(h)$
	be the function in $\Hs$ that minimizes the empirical risk.
\end{itemize}
As $\hat{h}$ is unknown, one has to approximate it by some $h\in \Hs$. 
$h^*$ is the best approximation for $\hat{h}$ in $\Hs$, 
while $h_I$ is the best hypothesis 
in $\Hs$
obtained by empirical risk minimization.
For simplicity,
we assume that 
$\hat{h}, 
h^*$ and $h_I$ are unique.
The \emph{total error} 
can be decomposed as
\cite{bottou2008tradeoffs,bottou2018optimization}:
\begin{align}
\label{eq:error_decomposition}
\E[R({h}_I)-R(\hat{h})] = \underbrace{\E[R(h^*) - R(\hat{h})]}_{\displaystyle\errapp(\Hs)} + ~ \underbrace{\E[R(h_I)-R(h^*)]}_{\displaystyle\errest(\Hs,I)},
\end{align}
where the expectation is with respect to the random choice of $\dtrain$.
The \emph{approximation error} $\errapp(\Hs)$ measures how close the
functions in $\Hs$ can approximate the optimal hypothesis $\hat{h}$,  and
the \emph{estimation error} $\errest(\Hs,I)$ measures the effect of minimizing the empirical risk $R_I(h)$ instead of the expected risk $R(h)$
within $\Hs$.

As shown, the total error is affected by $\Hs$ (hypothesis space) and $I$ (number of examples in $\dtrain$).
In other words, learning to reduce the total error can be attempted from the
perspectives of (i) data, which provides $\dtrain$; (ii) model, which determines
$\Hs$; and (iii) algorithm, which 
searches for the optimal $h_I\in\Hs$
that fits $\dtrain$.

\subsubsection{Unreliable Empirical Risk Minimizer}
\label{sssec:unrers}

In general, 
$\errest(\Hs,I)$ can be reduced by having a larger number of
examples 
\cite{friedman2001elements,bottou2008tradeoffs,bottou2018optimization}.  Thus,
when there is  
sufficient 
training data 
with supervised information
(i.e., $I$ is large),
the empirical risk minimizer $h_I$ can provide a good  
approximation $R(h_I)$ to the
best possible $R(h^*)$ for $h$'s in $\Hs$.

\begin{figure}[H]
	\centering				
	\subfigure[Large $I$.\label{fig:std}]{\includegraphics[width=0.47\textwidth]
		{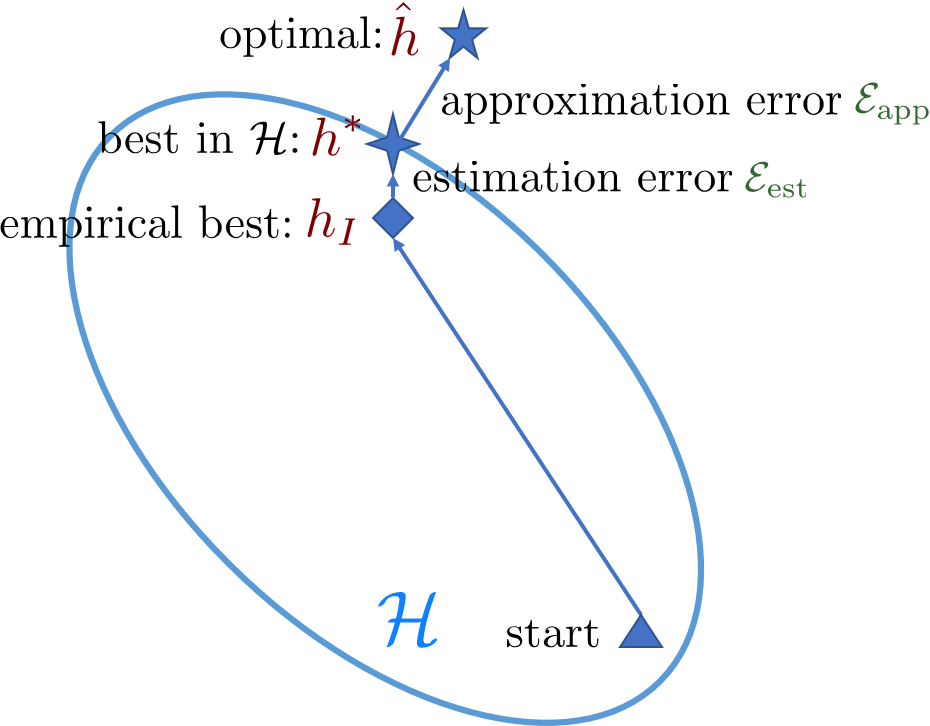}}	
	\hspace{10px}
	\subfigure[Small $I$.]{\includegraphics[width=0.32\textwidth]
		{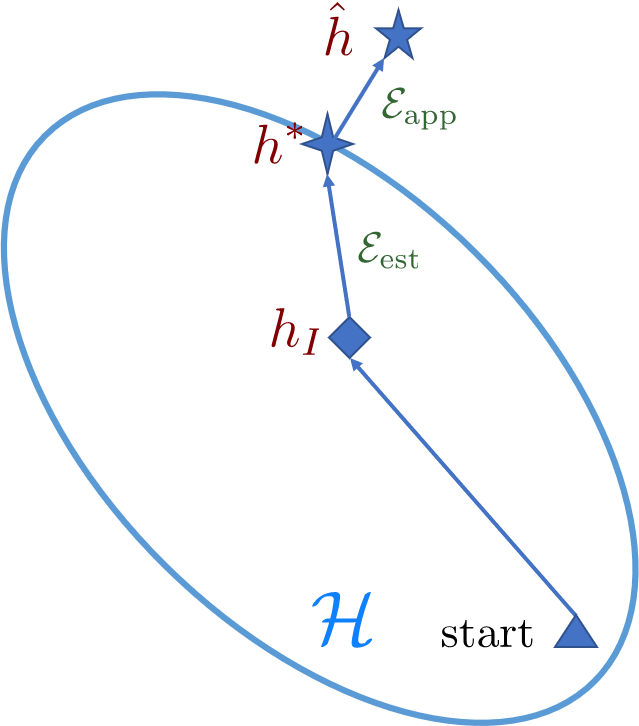}}
	\caption{Comparison of learning with sufficient and few training samples.
	}
	\label{fig:std_vs_fsl}
\end{figure}

However, 
in FSL,
the number of available examples $I$ is small. 
The empirical risk
$R_I(h)$ may then be far from being a good approximation of the expected risk $R(h)$,
and the resultant empirical risk minimizer $h_I$ overfits.
Indeed, 
this is the core issue of FSL supervised learning,
i.e., 
\textit{the empirical risk minimizer $h_I$ is no longer reliable}. 
Therefore, 
FSL is much harder. 
A comparison of learning with sufficient and few training samples is shown in 
Figure~\ref{fig:std_vs_fsl}.

\subsection{Taxonomy}
\label{sec:overview_taxonomy}

\begin{figure}[H]
	\centering				
	\subfigure[Data.\label{fig:fsl_err_data}]{\includegraphics[width=0.32\textwidth]
		{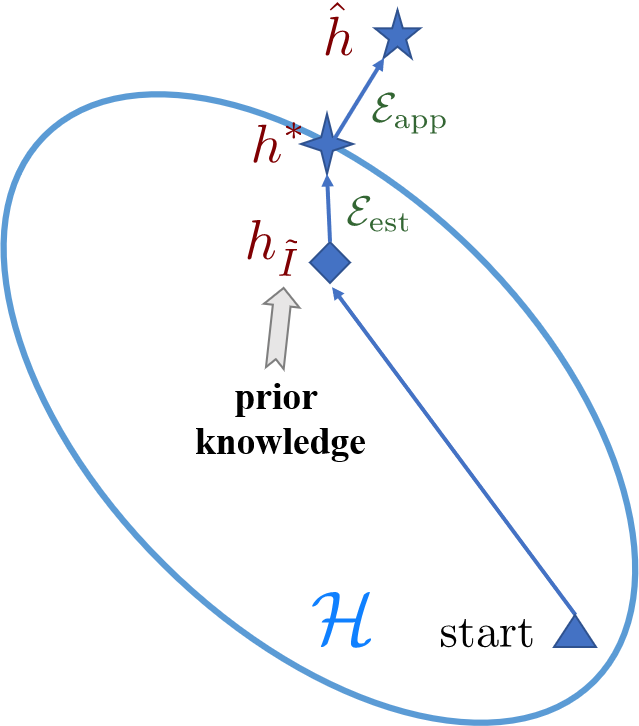}}	
	\subfigure[Model.\label{fig:fsl_err_model}]{\includegraphics[width=0.32\textwidth]
		{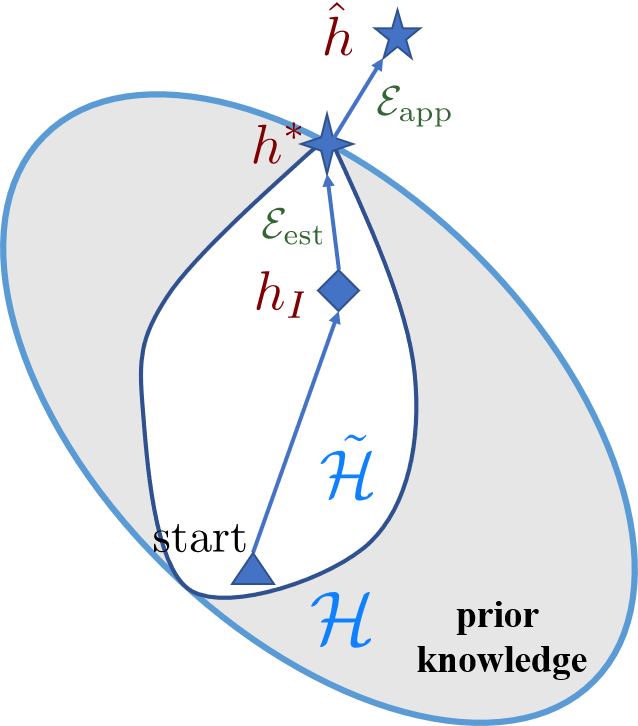}}
	\subfigure[Algorithm.\label{fig:fsl_err_alg}]{\includegraphics[width=0.32\textwidth]
		{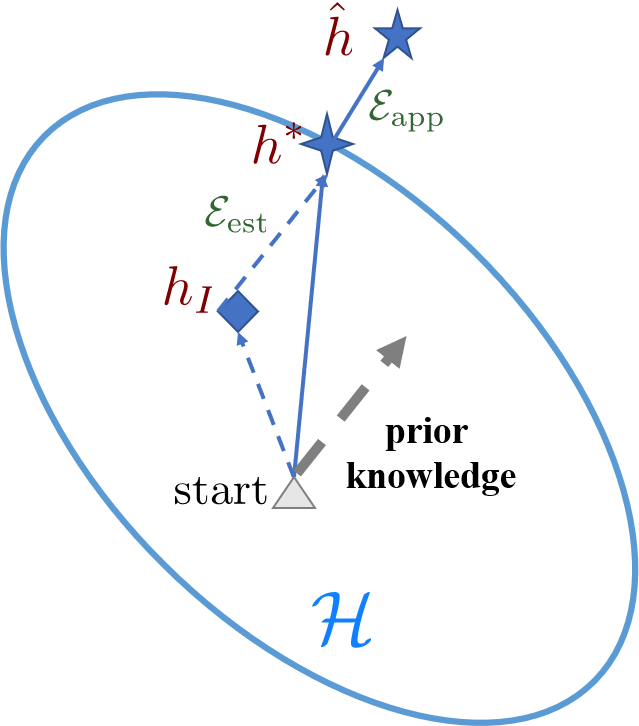}}	
	\caption{Different perspectives on how FSL methods solve the few-shot problem.  }
	\label{fig:fsl_err}
\end{figure}

To alleviate the problem of having an unreliable empirical risk minimizer $h_I$
in FSL supervised learning,
prior knowledge
must be used.
Based on which aspect is enhanced using prior knowledge, existing FSL works can be categorized
into the following perspectives  
(Figure~\ref{fig:fsl_err}).

\begin{itemize}
\item \textit{Data}. These methods use prior knowledge to augment $\dtrain$, and
increase the number of samples from $I$ to $\tilde{I}$, where $\tilde{I}\gg I$.
Standard machine learning models and algorithms can then be used on the augmented data,
and 
a more accurate empirical risk minimizer $h_{\tilde{I}}$ 
can be obtained
(Figure~\ref{fig:fsl_err_data}). 
	
\item \textit{Model}. These methods use prior knowledge 
to constrain the complexity of $\Hs$, which results in a much smaller hypothesis space
$\tilde{\Hs}$.  As shown in Figure ~\ref{fig:fsl_err_model}, the gray area is not
considered for optimization as they are known to be unlikely to contain the optimal
$h^*$ according to prior knowledge.  For this smaller $\tilde{\Hs}$, $\dtrain$ is
sufficient to learn a reliable $h_I$  \cite{mahadevan1994quantifying,nguyen2018improved,germain2016pac}.
	
\item \textit{Algorithm}. These methods use prior knowledge to
	search for the $\theta$ which parameterizes the best hypothesis $h^*$ in $\Hs$. 
	Prior knowledge alters the search strategy by providing a good
	initialization (gray triangle in Figure~\ref{fig:fsl_err_alg}), 
	or guiding the search steps (the gray dotted lines in Figure~\ref{fig:fsl_err_model}). 
	For the latter, 
	the resultant search steps are affected by both prior knowledge and empirical risk minimizer. 

\end{itemize}

\begin{figure*}[ht]
	\centering
	\includegraphics[width=1\linewidth]{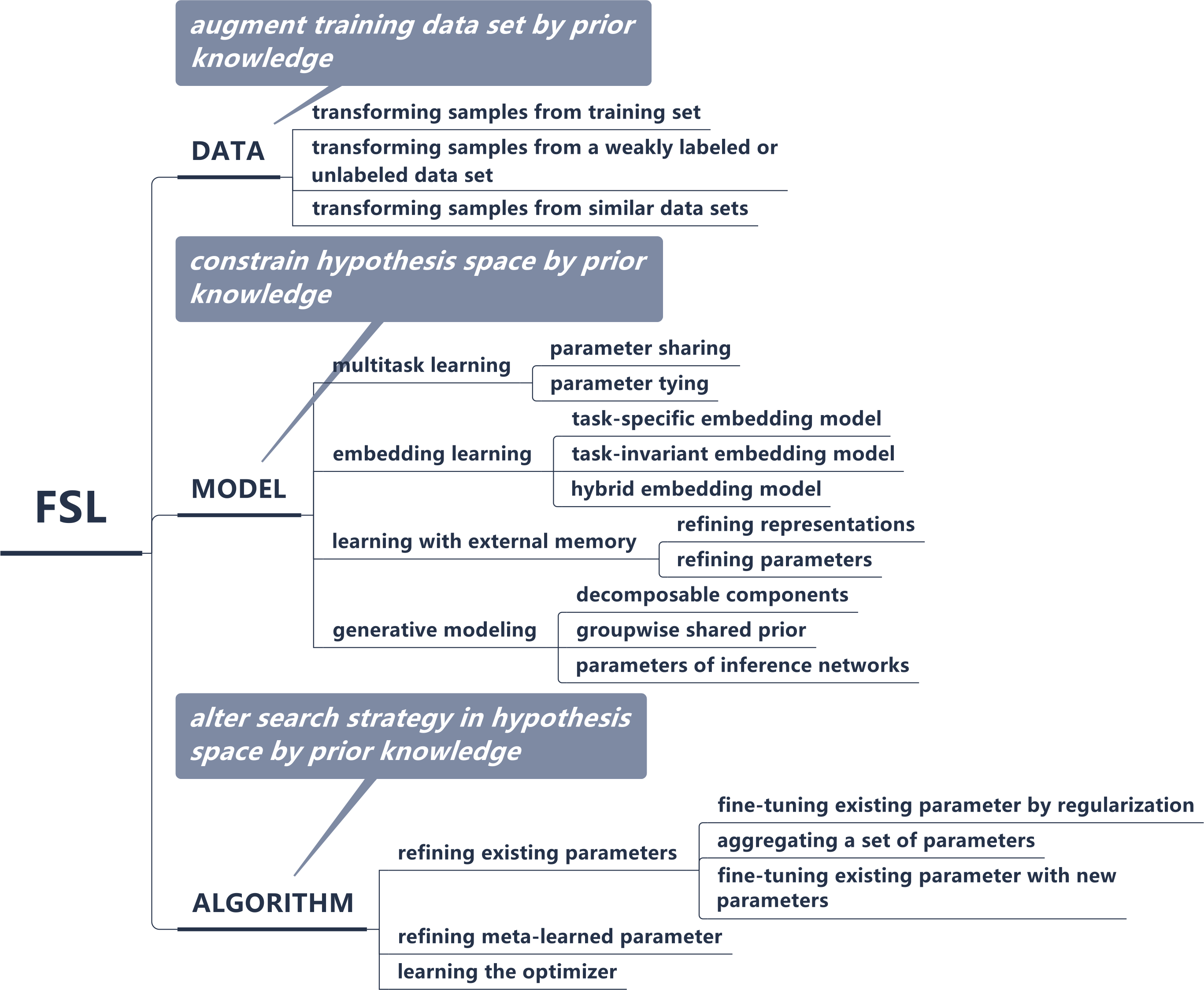}
	\caption{A taxonomy of FSL methods based on the focus of each method.}
	\label{fig:fsl_err_mind}
\end{figure*}

Accordingly, existing works can be categorized into a unified taxonomy as shown in 
Figure~\ref{fig:fsl_err_mind}. 
We will detail each category in the following sections.

\section{Data} 
\label{sec:data}
\begin{figure}[ht]
	\centering
	\includegraphics[width=0.6\linewidth]{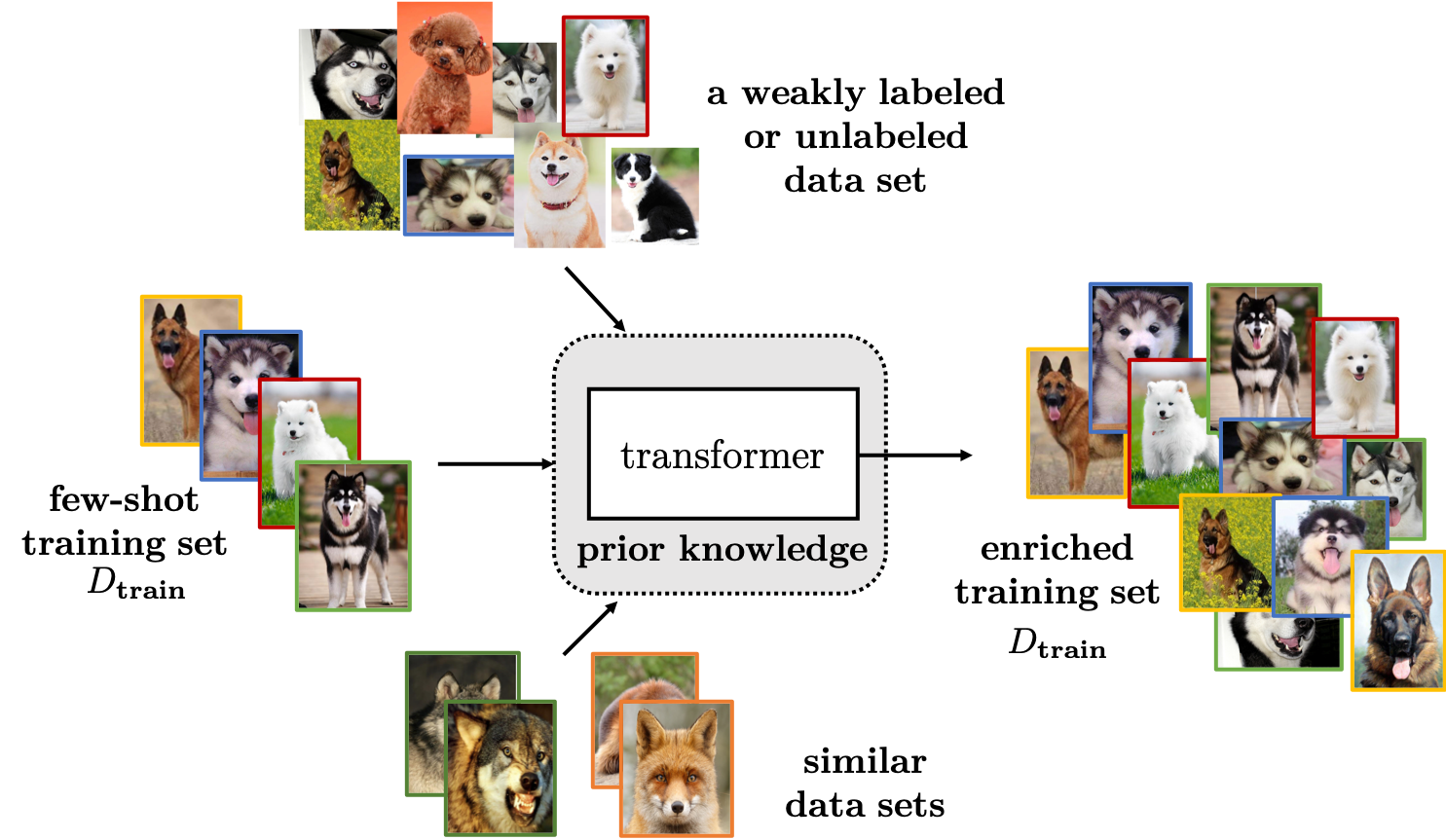}
	\caption{Solving the FSL problem by
	data augmentation.}
	\label{fig:transformer}
\end{figure}

FSL methods in this section use prior knowledge to augment data $\dtrain$, such that the supervised information in $E$ is enriched.
With the augmented sample set,  
the data is sufficient enough to 
obtain a reliable $h_I$ (Figure~\ref{fig:transformer}).

Data augmentation via hand-crafted rules is usually used as pre-processing in FSL methods.
They can introduce different kinds of invariance for the model to capture. 
For example, on images, one can use translation \cite{shyam17attentive,lake2015human,santoro2016meta,benaim2018one}, 
flipping \cite{shyam17attentive,qi2018low},
shearing \cite{shyam17attentive}, 
scaling \cite{lake2015human,zhang2018fine}, 
reflection \cite{edwards2017towards,kozerawski2018clear}, 
cropping \cite{qi2018low,zhang2018fine} and
rotation \cite{santoro2016meta,vinyals2016matching}.
However, designing these rules depends heavily on domain knowledge
and requires expensive labor cost. 
Moreover, the augmentation rules can be specific to the data set, making them hard to be
applied to other data sets.
Moreover, it is unlikely that human can enumerate all possible
invariance. 
Therefore, manual data augmentation cannot solve the FSL problem completely \cite{shyam17attentive,lake2015human,santoro2016meta,benaim2018one,edwards2017towards,kozerawski2018clear}. 

Besides these
hand-crafted rules,
we review 
in the following
more advanced data augmentation methods.
Depending on what samples are transformed and added to $\dtrain$, 
we categorize these methods 
as shown in Table~\ref{tab:data_aug_cha}.

\begin{table}[ht]
	\caption{Characteristics for FSL methods focusing on the data perspective.
	The transformer $t(\cdot)$ takes input $(x,y)$ and returns synthesized sample $(\tilde{x},\tilde{y})$ to augment the few-shot $\dtrain$.	
}
	\footnotesize
	\begin{tabular}
		{ C{100px}|C{90px}|C{90px}|C{70px}}
		\hline
		category&	input  $(x,y)$          & transformer $t$     & output $(\tilde{x},\tilde{y})$                    \\ \hline 
		transforming samples from $\dtrain$&
		original $\xyi$          & learned transformation function on $\xii$ & ($t({x}_i)$, $y_i$) \\ \hline
		transforming samples from a
		weakly labeled or unlabeled data set&
		weakly labeled or unlabeled $(\bar{x},-)$       & a predictor trained from $\dtrain$          & ($\bar{x}$, $t(\bar{x})$)   \\ \hline
		transforming samples from 
		similar data sets &
		samples $\{(\hat{x}_j,\hat{y}_j)\}$ from similar data sets & an aggregator to combine $\{(\hat{x}_j,\hat{y}_j)\}$  & 
		$(t(\{\hat{x}_j\}),t(\{\hat{y}_j\}))$
		\\ \hline
	\end{tabular}
	\label{tab:data_aug_cha}
\end{table}

\subsection{Transforming Samples from $\dtrain$}
This strategy augments $\dtrain$ by transforming 
each $\xyi\in\dtrain$ 
into several samples with variations. 
The transformation procedure is included in experience 
$E$ as prior knowledge so as to generate additional samples. 
An early
FSL paper \cite{miller2000learning} 
learns 
a set of geometric transformations from a similar class 
by iteratively aligning each sample with the other samples. 
The learned transformation is 
applied to each $\xyi$ to form a large data set, which can 
then 
be learned by standard
machine learning methods.
Similarly, 
a set of auto-encoders,
each representing one intra-class variability, 
are learned 
from similar classes
in \cite{eli2018delta}.
New samples 
are generated 
by adding the learned variations to $\xii$.
In \cite{hariharan2017low},
by assuming that all categories share some transformable variability across samples, a single transformation function is learned 
to transfer variation between sample pairs learned from the other classes to
$\xyi$.
In \cite{kwitt2016one}, instead of enumerating the variabilities within pairs, 
it transforms
each $\xii$ to several samples 
using a set of independent
attribute strength regressors learned 
from a large set of scene images, and assigns the label of the original $\xii$ to these new samples.
Improved upon \cite{kwitt2016one},  a continuous attribute subspace is used
to add attribute variations to $x$ in \cite{liu2018feature}.

\subsection{Transforming Samples from  a Weakly Labeled or Unlabeled Data Set}
\label{sec:data_unlabel}
This strategy augments $\dtrain$ by selecting samples with the target label from 
a large 
data set which is
weakly labeled or unlabeled. 
For example, in photos taken by a surveillance camera,
there are  people, cars and roads but none of them are labeled.
Another example is
a video
for a long presentation. This
contains a series of gestures of the speaker, but none of them are 
annotated
explicitly.
As such  
a data set contains large variations of samples, 
augmenting them to $\dtrain$ helps depict a clearer $p(x,y)$. 
Moreover, 
collecting such a data set is easier as human effort is not needed for labeling. 
However, though the
collection cost
is low, 
a major issue is how to select samples with the target label to be augmented to $\dtrain$. 
In \cite{pfister2014domain}, an exemplar SVM is learned for each target label in
$\dtrain$, which is then used to predict labels for samples from a weakly labeled
data set. Samples having the target labels are then added to $\dtrain$.
In \cite{douze2018low}, 
instead of learning a classifier,
label propagation is directly used to label an unlabeled data set.
In \cite{wu2018exploit},
a progressive strategy is used to select informative unlabeled samples. 
The selected samples are then assigned pseudo-labels and used to update the CNN.

\subsection{Transforming Samples from Similar Data Sets}
\label{sec:data_sim}
This strategy augments $\dtrain$ by aggregating and adapting input-output pairs
from a similar
but larger data sets. 
The aggregation weight  is usually 
based on some similarity measure between samples. 
In \cite{tsai2017improving}, it
extracts the aggregation weight from an auxiliary text corpus \cite{tsai2017improving}. 
As these samples may not come from the target FSL class,
directly augmenting the aggregated samples to $\dtrain$ may be misleading. 
Therefore, 
a generative adversarial network (GAN)
\cite{goodfellow2014generative} is designed to generate 
indistinguishable synthetic $\tilde{x}$
aggregated from a data set of many samples \cite{gao2018low}.
It has two generators, one maps samples of the few-shot class to the large-scale
class, and the other maps samples of the large-scale class to the few-shot class
(to compensate for the lack of samples in GAN training).

\subsection{Discussion and Summary}
\label{sec:data_sum}
  
The choice of which augmentation strategy  to use
depends on the application.
Sometimes,
a large number of weakly supervised or unlabeled samples 
exist
for the target task (or class), but
few-shot learning is preferred because
of the high cost of gathering
annotated data and/or
computational cost
(which corresponds to the third scenario introduced in Section~\ref{sec:overview_problem_define}).
In
this case, one can perform augmentation by transforming samples from a
weakly
labeled or unlabeled data set. 
When a large-scale unlabeled data set is hard to collect, 
but the few-shot class has some similar classes, 
one can transform samples from these similar classes. 
If only some learned transformers rather than raw samples are available, augmentation can be done by transforming the original samples from $\dtrain$.

In general, 
solving a FSL problem by augmenting $\dtrain$ is straightforward and easy to understand.
The data is augmented by taking advantage of the prior information for the target task. 
On the other hand, the weakness of solving the FSL problem by data augmentation is
that the augmentation policy is often
tailor-made for each data set in an adhoc manner,
and
cannot be used 
easily
on other data sets (especially for data sets from other domains). 
Recently, AutoAugment 
\cite{cubuk2019autoaugment}, which
automatically learns the augmentation policy for deep network training,
is proposed to address this issue. 
Apart from that, 
existing methods are mainly designed for images, as
the generated images can be 
visually 
evaluated 
by humans
easily.
In contrast, text and audio involve syntax and structures, and are harder to
generate. A recent attempt on using 
data augmentation for text is reported in \cite{wei2019eda}.

\section{Model}
\label{sec:model} 
In order to approximate the ground-truth
hypothesis $\hat{h}$,
the model has to determine
 a hypothesis space $\Hs$ containing a family of hypotheses $h$'s,
such that the distance between the optimal $h^*\in\Hs$ and $\hat{h}$ is small.

Given the few-shot $\dtrain$
with limited samples, 
one can
choose a small $\Hs$ 
with only simple models (such as linear classifiers)
\cite{tom1997machine,mohri2018foundations}. 
However, 
real-world 
problems 
are typically complicated, and cannot be well represented by an hypothesis $h$
from a small $\Hs$ (which can lead to a large $\errapp(\Hs)$ in \eqref{eq:error_decomposition}) \cite{goodfellow2016deep}.
Therefore, a large enough $\Hs$ is preferred in FSL, which makes standard machine learning models infeasible.
FSL methods in this section manage to learn 
by 
constraining 
$\Hs$ 
to a smaller hypothesis space $\tilde{\Hs}$ via prior knowledge  in $E$  
(Figure~\ref{fig:fsl_err_model}).
The empirical risk minimizer is then more reliable, and the risk of overfitting is reduced.

In terms of what prior knowledge is used, methods belonging to this category can be
further classified into four types
(Table~\ref{tab:model_aux}). 

\begin{table}[ht]
	\caption{Characteristics for FSL methods focusing on the model perspective.}
	\footnotesize
	\begin{tabular}{ C{100px}|C{130px}| C{130px}}
		\hline
		strategy   &           prior knowledge             & how to constrain $\Hs$   \\\hline
		multitask learning      &      other $T$'s with their data sets $D$'s      & share/tie parameter \\ \hline
		embedding learning      &   embedding learned from/together with other $T$'s
		& project samples to a smaller embedding space in which similar and dissimilar samples can be easily discriminated     \\ \hline
		learning with external memory & embedding learned from other $T$'s to interact with memory & refine samples using key-value pairs stored in memory\\ \hline
		generative modeling       &  prior model learned from other $T$'s   & restrict the form of distribution     \\ \hline
	\end{tabular}
	\label{tab:model_aux}
\end{table}

\subsection{Multitask Learning}
\label{sec:model_multitask}

In the presence of multiple related tasks,
\textit{multitask learning} \cite{caruana1997multitask,zhang2017survey}
learns these tasks simultaneously by exploiting both task-generic 
and task-specific information.  
Hence, they can be naturally used for FSL. 
Here, we present some instantiations of using multitask learning in FSL .  

We are given $C$ related tasks $\T_1,\dots,\T_C$, in which some of
them have
very few samples while some have a larger number of samples.
Each task $\T_c$ has a data set $\D_c =
\{\dtrain^c,\dtest^c\}$, in which
$\dtrain^c$
is the training set 
and 
$\dtest^c$ is the 
test set.
Among these $C$ tasks, 
we regard the few-shot tasks as \textit{target tasks}, and the rest as \textit{source tasks}.
Multitask learning learns from 
$\dtrain^c$'s to obtain $\theta_c$ for each $\T_c$. 
As these tasks are jointly learned, the parameter $\theta_c$ of $h_c$ learned for
task $T_c$ is constrained by the other tasks.
According to how the task parameters are constrained, we divide methods in this
strategy as (i) parameter sharing; and (ii) parameter tying \cite{goodfellow2016deep}.

\subsubsection{Parameter Sharing}

\begin{figure}[ht]
	\centering
	\includegraphics[width=0.7\linewidth]{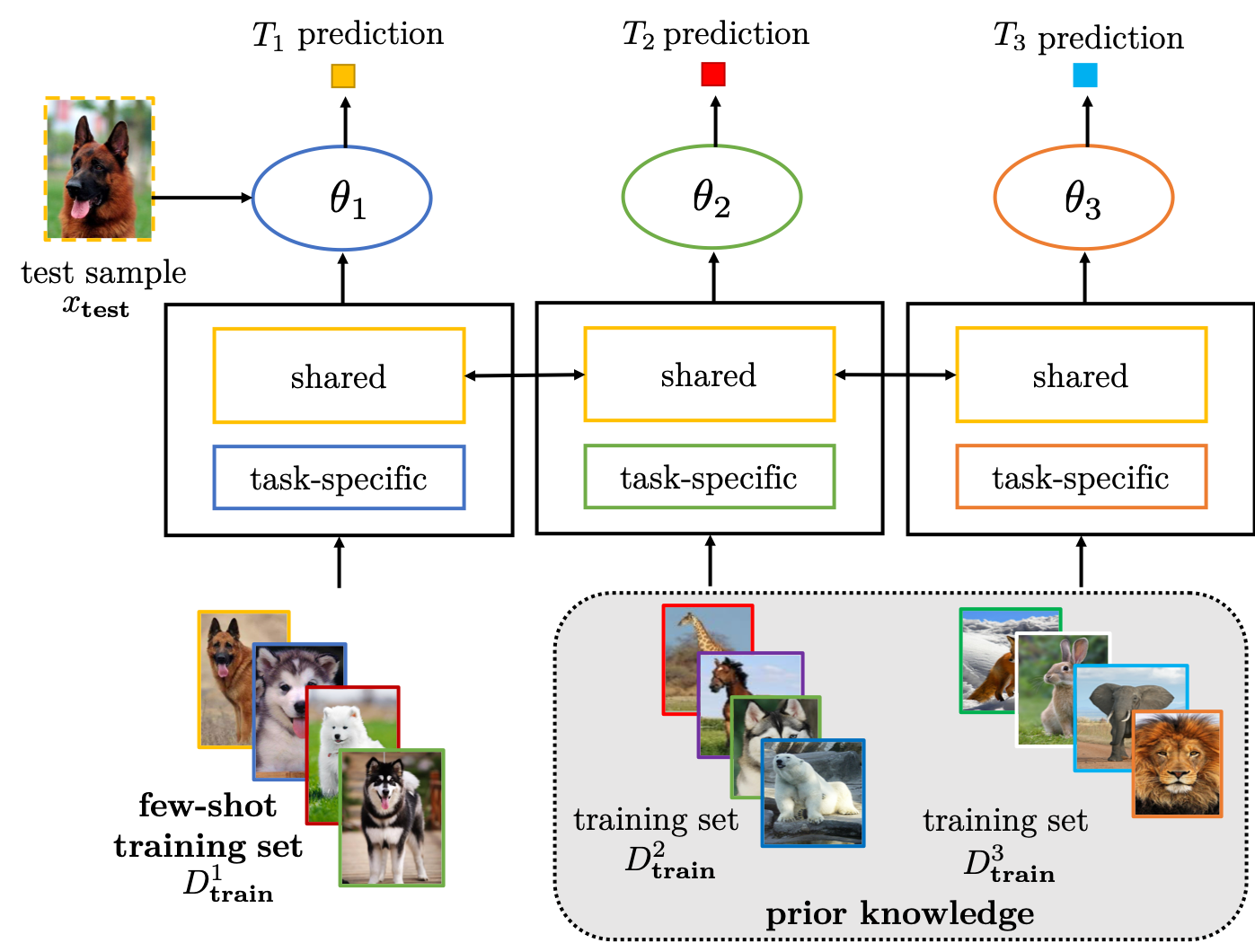}
	\caption{Solving the FSL problem by multitask learning with parameter sharing.}
	\label{fig:multitask_hard}
\end{figure}

This strategy directly shares some parameters among tasks (Figure \ref{fig:multitask_hard}). 
In \cite{zhang2018fine}, 
the two task networks
share the first few layers 
for the generic information, 
and learn different final layers to deal with different outputs. 
In \cite{hu2018few}, 
two
natural language processing 
tasks on 
legal texts 
are solved 
together:
charge prediction and legal attribute prediction.
A single embedding function 
is used
to encode the criminal case description, which is then fed to 
task-specific 
embedding functions and classifiers.
In \cite{motiian2017few},
a variational auto-encoder 
is first pre-trained
from the source tasks, and then cloned to the target task. 
Some layers in the two variational auto-encoders 
are shared 
in order to capture the generic information, 
while allowing both tasks to have some task-specific layers. 
The target task can only update its task-specific layers, while the source task
can update both the shared and task-specific layers. 
In \cite{benaim2018one}, 
both the original and generated samples 
are first mapped
to a task-specific space
by learning separate embedding functions for the source and target tasks, and are
then embedded by a shared variational auto-encoder.

\subsubsection{Parameter Tying}

\begin{figure}[ht]
	\centering
	\includegraphics[width=0.85\linewidth]{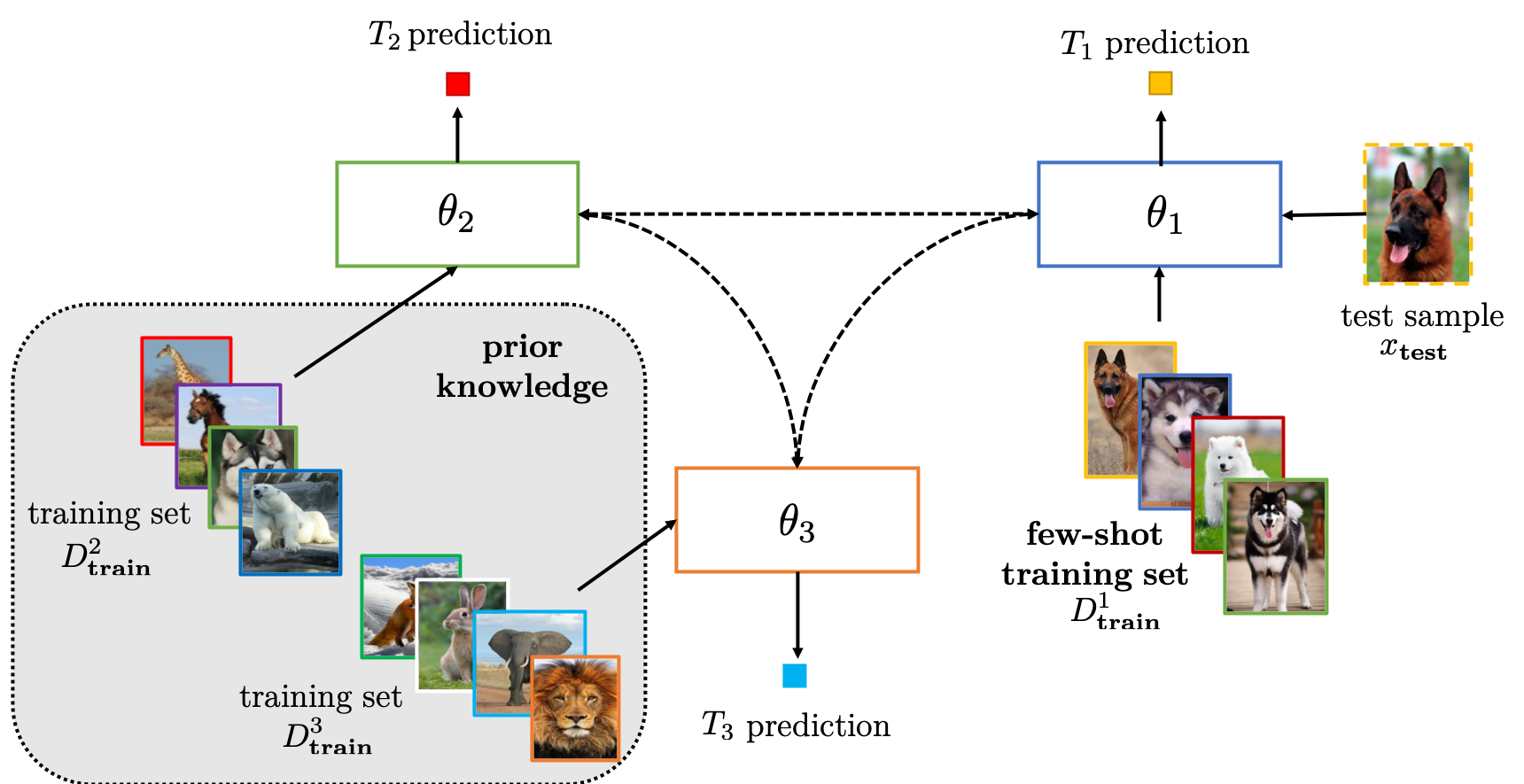}
	\caption{Solving the FSL problem by
	multitask learning with parameter tying.}
	\label{fig:multitask_soft}
\end{figure}

This strategy encourages parameters ($\theta_c$'s)
of different tasks to be similar (Figure~\ref{fig:multitask_soft}) \cite{goodfellow2016deep}. 
A popular approach is by
regularizing the $\theta_c$'s.
In \cite{yan2015multi}, all pairwise differences of $\theta_c$'s 
are penalized.
In \cite{luo2017label}, there is a 
CNN for the source task, and another one
for the
target task.
Layers of  these two CNNs 
are aligned using some specially designed regularization
terms.

\subsection{Embedding Learning}
\textit{Embedding learning} \cite{spivak1970comprehensive,jia2014caffe} 
embeds each sample $\xii\in \X \subseteq \R^{d}$ to  
a lower-dimensional
$z_i\in \Z \subseteq \R^{m}$,
such that similar samples are close together while dissimilar samples can be more easily differentiated.
In this lower-dimensional $\Z$, one can then construct a smaller hypothesis
space $\tilde{\Hs}$ which subsequently requires fewer training samples. 
The embedding function is mainly learned from prior knowledge, and can additionally use 
task-specific information
from $\dtrain$.

Embedding learning has the following key components: 
(i) a
function $f$ which embeds test sample $\xtest\in\dtest$ to $\Z$,  (ii) a 
function $g$ which embeds training sample $\xii\in\dtrain$ to $\Z$, 
and (iii) a similarity function $s(\cdot,\cdot)$ which measures the similarity
between $f(\xtest)$ and $g(\xii)$ in $\Z$.
The test sample $\xtest$ is 
assigned to the class of $\xii$, whose embedding $g(\xii)$ is most similar 
to $f(\xtest)$ in $\Z$
according to $s$.
Although one can use a common embedding function for both
$\xii$ and $\xtest$, 
using two separate embedding functions may obtain better accuracy \cite{bertinetto2016learning,vinyals2016matching}. 
A summary of existing embedding learning methods is shown in
Table~\ref{tab:model_embedding_cha}. 

According to whether the parameters of embedding functions $f$ and $g$ vary across tasks, 
we classify these FSL methods
as using a (i)
task-specific embedding model; (ii) task-invariant (i.e., general) embedding model; and (iii)
hybrid embedding model, which encodes both task-specific and task-invariant information.

\begin{table}[ht]
	\caption{Characteristics of embedding learning methods. 
	}
	\footnotesize
	\begin{tabular}
		{C{45px} |C{80px}|C{70px}|C{70px}|C{70px}}\hline 
		category&method	&embedding function $f$ for $\xtest$& embedding function $g$ for $\dtrain$&similarity measure $s$
		 \\
		\hline
		task-specific&mAP-DLM/SSVM\cite{triantafillou2017few}&CNN&the same as $f$&cosine similarity\\\hline
		\multirow{15}{*}{task-invariant}
		&class relevance pseudo-metric \cite{fink2005object}& kernel&the same as $f$&squared $\ell_2$ distance\\\cline{2-5}
		&convolutional siamese net \cite{koch2015siamese} & CNN&the same as $f$&weighted $\ell_1$ distance\\\cline{2-5}
		&Micro-Set\cite{tang2010optimizing}&logistic projection&the same as $f$&$\ell_2$ distance\\\cline{2-5}		
		&Matching Nets \cite{vinyals2016matching}& CNN, LSTM& CNN, biLSTM& cosine similarity\\\cline{2-5}
		&resLSTM	\cite{altae2017low}& GNN, LSTM& GNN, LSTM& cosine similarity \\\cline{2-5}
		&Active MN \cite{bachman17learning}&CNN&biLSTM&cosine similarity \\\cline{2-5}
		&SSMN \cite{choi2018structured}&CNN&another CNN&learned distance\\\cline{2-5}
		&ProtoNet	\cite{snell2017prototypical}&CNN&the same as $f$&
		squared $\ell_2$ distance\\\cline{2-5}
		&semi-supervised ProtoNet\cite{ren2018metalearning}&CNN&the same as $f$&
		squared $\ell_2$ distance\\\cline{2-5}
		&PMN	\cite{wang2018low}&CNN, LSTM& CNN, biLSTM& cosine similarity \\\cline{2-5}
		&ARC \cite{shyam17attentive}&LSTM, biLSTM&the same as $f$&-\\\cline{2-5}		
		&Relation Net \cite{sung2018learning}&CNN&the same as $f$&-\\\cline{2-5}
		&GNN \cite{garcia2018fewshot}&CNN, GNN&the same as $f$&learned distance\\\cline{2-5}		
		&TPN \cite{liu2019learning}&CNN&the same as $f$&Gaussian similarity\\\cline{2-5}
		&SNAIL \cite{mishra2018a}& CNN&the same as $f$&-\\\hline
		\multirow{4}{*}{hybrid}
		&Learnet	\cite{bertinetto2016learning}& adaptive CNN & CNN&weighted $\ell_1$ distance\\\cline{2-5} 
		&DCCN \cite{zhao2018dynamic}&adaptive CNN&CNN&-\\\cline{2-5}
		&R2-D2 \cite{bertinetto2018metalearning}& adaptive CNN&CNN&-\\\cline{2-5}
		&TADAM	\cite{oreshkin2018tadam}&adaptive CNN&the same as $f$&
		squared $\ell_2$ distance\\\hline
	\end{tabular}
	\label{tab:model_embedding_cha}
\end{table}

\subsubsection{Task-Specific Embedding Model}
Task-specific embedding methods learn an embedding function tailored for each task,
by using
only 
information from that
task.  
For example, 
using the few-shot data $\dtrain^c$
of task $T_c$,
all pairwise rankings among samples in 
$\dtrain^c$ 
are enumerated 
as sample pairs
in \cite{triantafillou2017few}. 
The number of training samples is thus increased, and
an embedding function  can be learned even though
only the task-specific information is used.

\subsubsection{Task-Invariant Embedding Model}
\begin{figure}[ht]
	\centering
	\includegraphics[width=0.6\linewidth]{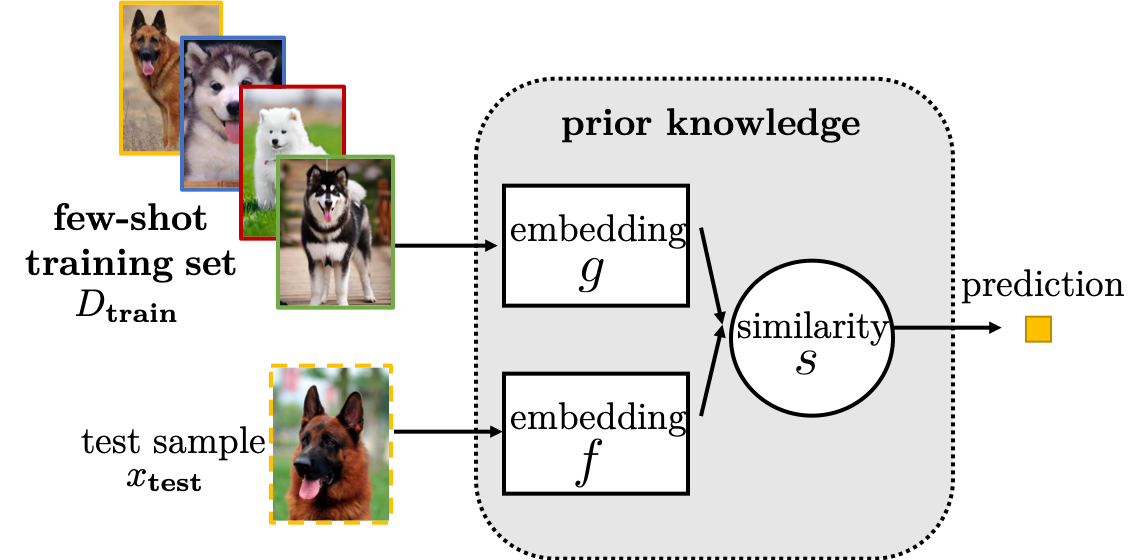}
	\caption{Solving 
		the FSL problem by
		task-invariant embedding model. 
	}
	\label{fig:embedding_invariant}
\end{figure}

Task-invariant embedding methods 
learn 
a general embedding function 
from a large-scale data set containing sufficient samples with various outputs, 
and
then directly use this
on the new few-shot $\dtrain$ without retraining (Figure~\ref{fig:embedding_invariant}). 
The first FSL embedding model \cite{fink2005object} embeds the samples using a
kernel.
Recently, more complicated embeddings are learned \cite{koch2015siamese,yan2018few}
by a
convolutional siamese net
\cite{bromley1994signature}.

Although task-invariant embedding does not update the 
embedding model 
parameter 
using 
the few-shot $\dtrain$, 
many methods in this category
\cite{vinyals2016matching,snell2017prototypical,sung2018learning} simulate the
few-shot scenario while training the embedding model.
Assume that we have training sets $\{\D_c\}$, each has 
$N$ classes.
In each $\D_c$, samples from only $U$ out of its $N$ classes 
are used for training.
The 
embedding model is optimized by maximizing the performance on the
remaining $N-U$ classes. 
Thus, the learned model will have good generalization for few-shot
tasks.
An early attempt \cite{tang2010optimizing} learns a linear embedding from $\{\D_c\}$.
Recently, more complicated task-invariant embedding models are learned via meta-learning\footnote{A brief introduction on meta-learning is provided in Appendix~\ref{app:meta}.  
The few-shot task is actually one of the meta-testing task $\T_t$ with $\D_t = \{\dtrain^t,\dtest^t\}$.
For illustration simplicity, we drop the subscript and superscript $t$.}
methods: 
\begin{enumerate}
	\item \textit{Matching Nets} 
	\cite{vinyals2016matching} 
	and its variants \cite{altae2017low,bachman17learning,choi2018structured}:
	Matching Nets
	\cite{vinyals2016matching} 
	meta-learns different embedding functions ($f$ and $g$) for the training sample $\xii$ and test sample $\xtest$.
	The residual LSTM (resLSTM)
	\cite{altae2017low} proposes better designs for $f$ and $g$. 
	An active learning variant of Matching Nets
	\cite{bachman17learning}
	adds a sample selection step, which labels the most beneficial unlabeled sample
	and uses it to augment $\dtrain$.
	The Matching Nets is also extended to set-to-set matching
	\cite{choi2018structured}, which is useful in labeling multiple parts of a sample.

	\item \textit{Prototypical Networks (ProtoNet)} \cite{snell2017prototypical} 
	and its variants \cite{wang2018low,oreshkin2018tadam,ren2018metalearning}:
	Instead of comparing $f(\xtest)$ with each $g(\xii)$ where $\xii\in\dtrain$, 
	ProtoNet \cite{snell2017prototypical} only compares
	$f(\xtest)$
	with the class prototypes in $\dtrain$.
	For class $n$,
	its prototype 
	is  simply  
	$c_n = \frac{1}{K}\sum_{i=1}^{K}g(\xii)$, where the
	$K$ 
	$\xii$'s
	are from class $n$.
	Empirically, this leads to more stable results and reduces the computation cost.  
	The idea of using prototypes is introduced to the Matching Nets in \cite{wang2018low}.  
	A semi-supervised variant of ProtoNet assigns unlabeled samples to augment $\dtrain$ via soft-assignment during learning \cite{ren2018metalearning}.
	\item \textit{Other methods}. Examples include Attentive Recurrent Comparators (ARC)
	\cite{shyam17attentive}, which uses a LSTM with
	attention \cite{bahdanau2015neural} to compare different regions of $\xtest$
	with
	prototype $c_n$,  and
	then embeds the comparison results
	as an intermediate embedding. 
	Additionally,
	it 
	uses a bidirectional LSTM (biLSTM) to embed all comparisons as the final embedding. 
	The Relation Net \cite{sung2018learning} uses a CNN to embed $\xtest$ and $\xii$ to
	$\Z$, then concatenates them as the embedding, which is fed to another CNN to
	output a similarity score. 
	The graph neural network (GNN) is used in \cite{garcia2018fewshot,liu2019learning} 
	to leverage information from local neighborhoods. 
	In few-shot reinforcement learning applications 
	(as in continuous control and visual navigation),
	temporal information
	is important.
	The Simple Neural AttentIve Learner (SNAIL) \cite{mishra2018a} is an embedding
	network with interleaved 
	temporal convolution layers and attention layers.
	The temporal convolution layer aggregates information from past time steps,
	while the attention layer selectively attends to specific time steps relevant to the current input. 
\end{enumerate}

\subsubsection{Hybrid Embedding Model}
\label{sec:model_embedding_meta}

\begin{figure}[ht]
	\centering
	\includegraphics[width=0.6\linewidth]{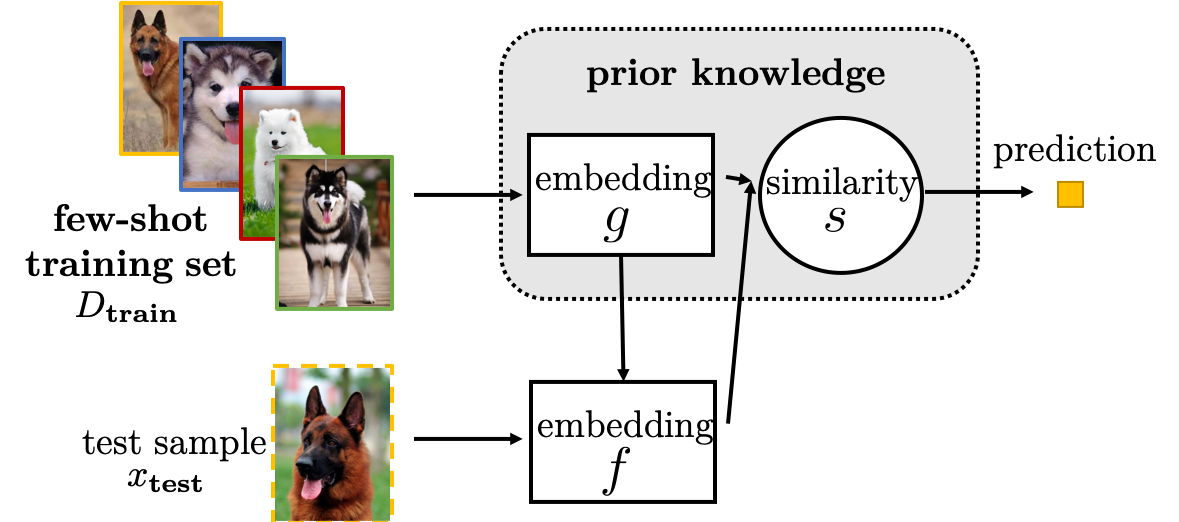}
	\caption{Solving 
		the FSL problem by
		hybrid embedding model. 
	}
	\label{fig:embedding_hybrid}
\end{figure}

Although task-invariant embedding methods can be applied to new tasks with 
a low 
computation cost, they do not leverage specific knowledge of the current
task.  
When task specialty is the reason that 
$\dtrain$ has only a few examples (such as learning for rare cases), 
simply applying a task-invariant embedding function may not be suitable. 
To alleviate this problem, hybrid embedding models 
adapt the generic task-invariant embedding model learned from prior knowledge by the task-specific information in $\dtrain$ 
This is done by learning a function 
which takes information extracted from $\dtrain$ as input and returns an embedding
which acts as the parameter for $f(\cdot)$ (Figure~\ref{fig:embedding_hybrid}).

Learnet \cite{bertinetto2016learning} improves the task-invariant convolutional siamese net
\cite{koch2015siamese} by incorporating the specific information of $\dtrain$.
It learns a meta-learner from multiple meta-training sets, and
maps each training example $\xii\in\dtrain$ to the parameter of 
the learner
(a convolutional siamese net). 
In this way, the parameter of $f(\cdot)$ changes with the given $\xii$, resulting in a hybrid embedding.
Improved upon Learnet, 
the classification layer of the learner is replaced
by ridge regression in \cite{bertinetto2018metalearning}, such that
parameters can be efficiently obtained in closed-form. 
The following two works \cite{zhao2018dynamic,oreshkin2018tadam} take $\dtrain$ as a whole to output the task-specific parameter for $f(\cdot)$.
Task dependent adaptive metric (TADAM) \cite{oreshkin2018tadam}  
averages class prototypes into the task embedding, and uses a meta-learned function to map it to the ProtoNet parameters. 
Dynamic Conditional Convolutional Network (DCCN) \cite{zhao2018dynamic} uses a fixed set of filters, and learns the
combination coefficients using $\dtrain$. 

\subsection{Learning with External Memory}

\begin{figure}[ht]
	\centering
	\includegraphics[width=0.8\linewidth]{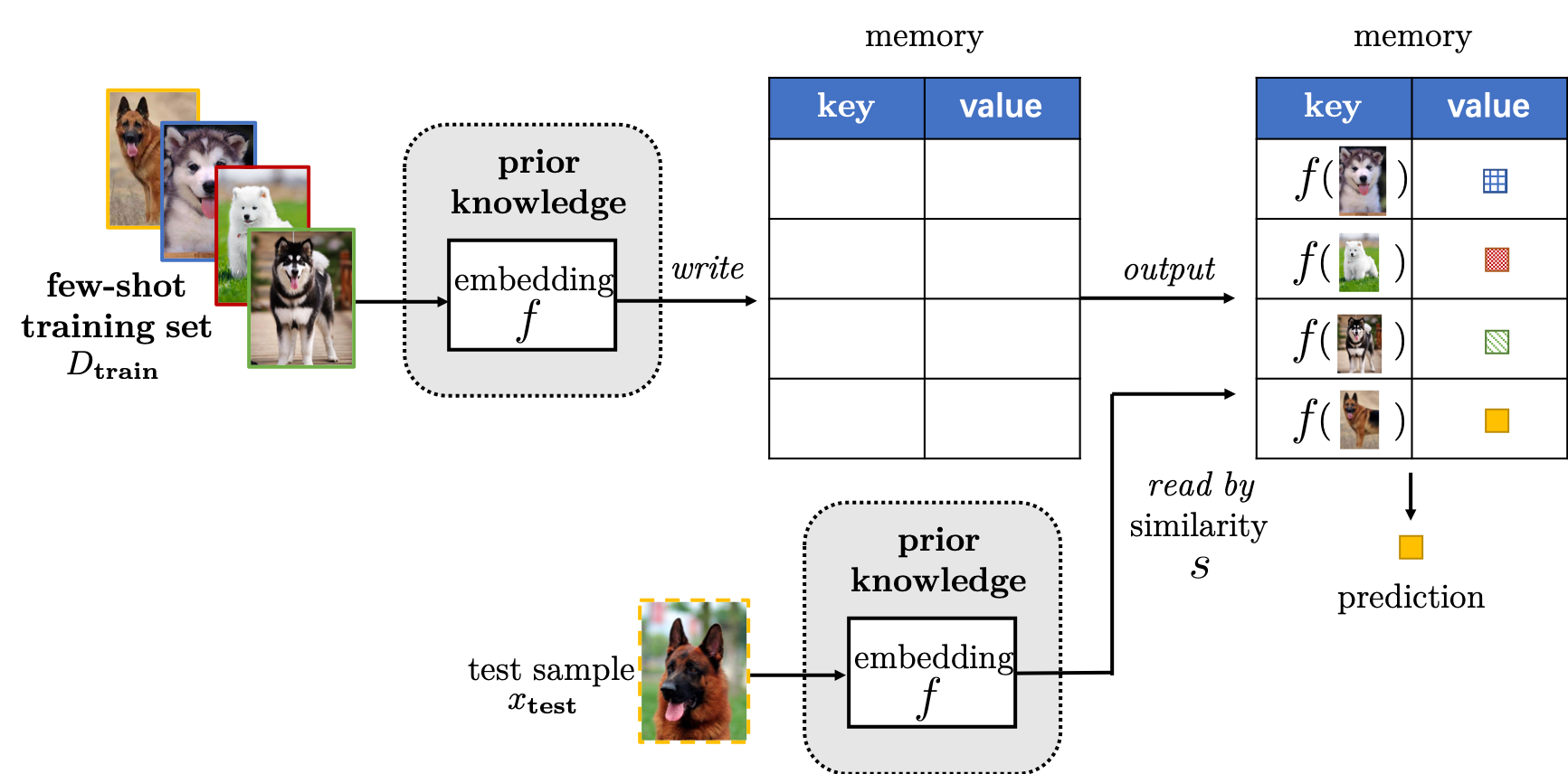}
	\caption{Solving the
	FSL problem
	by learning with external memory. 
	This figure illustrates a simplified example where the embedding function $f$ is used for representation learning 
	and the memory takes $f(\xii)$ as the key and output $y_i$ as the label.
	}
	\label{fig:memory}
\end{figure}

Learning with external memory \cite{graves2014neural,weston2014memory,sukhbaatar2015end,miller2016key}  
extracts
knowledge from $\dtrain$, and stores it in an external memory (Figure~\ref{fig:memory}). 
Each new sample $\xtest$ is then represented by a weighted average of contents extracted from the memory. 
This limits $\xtest$ to be represented by contents in the memory, and thus
essentially
reduces the size of $\Hs$.

A key-value memory \cite{miller2016key} is 
usually
used in FSL.  
Let the memory be $M\in\R^{b\times m}$, with each of its $b$ memory
slots $M(i)\in\R^{m}$ consisting of a key-value pair $M(i) = (M_{\text{key}}(i),M_{\text{value}}(i))$. 
A test sample
$\xtest$ is first embedded by an embedding function $f$. 
However, unlike embedding methods, $f(\xtest)$ 
is not used directly as the representation of $\xtest$. 
Instead, it is only used to query for the most similar memory slots, based on 
the similarity $s(f(\xtest),M_{\text{key}}(i))$
between 
$f(\xtest)$
and 
each key $M_{\text{key}}(i)$.
The values
of the most
 similar memory slots 
($M_{\text{value}}(i)$'s) 
are extracted and combined to form the representation of $\xtest$. This is  then
used as input to a simple classifier 
(such as a softmax function)
to make prediction.
 As manipulating $M$ is expensive, $M$ usually has a small size.
 When $M$ is not full, new samples can be written to vacant memory slots.
When $M$ is full, one has to decide which memory slots to be replaced. 
Table~\ref{tab:model_mem_cha}\
introduces the characteristics for methods with
external memory.
\begin{table}[ht]
	\centering
	\footnotesize
\caption{Characteristics of FSL methods based on learning with external memory.
Here, $f$ is an
embedding function usually 
pre-trained by
CNN or LSTM. 
	}
	\label{tab:model_mem_cha}
	\begin{tabular}
		{c|c|c|c|c} 
		\hline
		\multirow{2}{*}{category}&\multirow{2}{*}{method}& \multicolumn{2}{c|}{memory $M$} & \multirow{2}{*}{similarity $s$}\\ \cline{3-4}
		&&key $M_{\text{key}}$&value $M_{\text{value}}$&\\\hline
		&MANN \cite{santoro2016meta}     &$f(\xii,y_{i-1})$   &$f(\xii,y_{i-1})$ &cosine similarity   \\ \cline{2-5}
		&APL \cite{ramalho2018adaptive} &$f(\xii)$ &$\yii$ &squared $\ell_2$ distance \\\cline{2-5}
		refining&abstraction memory\cite{xu2017few}  &$f(\xii)$&word embedding of $\yii$   &dot product  \\ \cline{2-5}	
		representations&CMN \cite{zhu2018compound}    &$f(\xii)$&$\yii$, age   &dot product  \\ \cline{2-5}		
		&life-long memory \cite{kaiser2017learning}  &$f(\xii)$&$\yii$, age &cosine similarity\\\cline{2-5}	
	&Mem2Vec\cite{sun2018memory}   &$f(\xii)$&word embedding of $\yii$, age   &dot product  \\ \hline	
		&MetaNet \cite{munkhdalai2017meta}  &$f(\xii)$   &fast weight  &cosine similarity   \\ \cline{2-5}		
		refining&CSNs \cite{munkhdalai2018rapid} &$f(\xii)$   &fast weight  &cosine similarity   \\ \cline{2-5}		
		parameters&MN-Net \cite{cai2018memory} &$f(\xii)$ &$\yii$ &dot product\\\hline
	\end{tabular}
\end{table}

As each $\xtest$ is represented as a weighted average of values extracted from the memory, the quality of key-value pairs in the memory is important. 
According to the functionality of the memory, 
FSL methods in this category can be subdivided
into two types.

\subsubsection{Refining Representations}
The following methods carefully put $\dtrain$ into the memory, such that the stored key-value pairs can represent $\xtest$ more accurately.
Memory-Augmented Neural Networks (MANN) \cite{santoro2016meta} meta-learns the embedding $f$, and 
maps samples of
the same class to the same value.
Samples of the same
class then
refine their class representations in the memory together.
This class representation can be viewed as a refined class prototype in ProtoNet \cite{snell2017prototypical}. 
The surprise-based memory module \cite{ramalho2018adaptive} updates $M$ only when
it cannot represent an $\xii$ well. 
Hence, updating $M$ using this $\xii$ makes $M$ more expressive,
and also reduces the computation cost.
The abstract memory 
\cite{xu2017few} uses two memories. One extracts relevant
key-value pairs from a fixed memory containing  large-scale machine annotated data set,
and the other refines the extracted values and abstracts out the
most useful information for few-shot (image) classification. 
This idea is extended to few-shot video classification in \cite{zhu2018compound}.

Along this line, 
some methods pay special attention to protecting the few-shot classes in the memory.
Note that few-shot classes are small, and so
have a lower
chance of being
kept in $M$.
Each few-shot sample in $M$ can also be easily replaced by 
samples from the more abundant classes.
To alleviate this problem, lifelong memory \cite{kaiser2017learning} is proposed. 
Unlike previous memories
\cite{santoro2016meta,xu2017few,zhu2018compound,ramalho2018adaptive} which wipe out
the memory content across tasks, the lifelong memory 
erases the 
``oldest"
memory value 
when the memory is full.
The ages of all the memory slots are then reset to zero.
For a new sample, 
when the returned $M_{\text{value}}(i)$ value matches its ground-truth output, 
it is merged with the current $M_{\text{key}}(i)$ instead of being written to a new memory slot. 
Hence, it is more likely that
all classes occupy an equal number of memory slots, and rare classes are protected. 
Recently, this lifelong memory is adapted to learn word representations in \cite{sun2018memory}.

However,  even with the use of a lifelong memory,
rare samples can still be forgotten. 
After each update, the lifelong memory resets the age of the selected $M(i)$ to
zero, and increases the ages of the other non-empty memory slots by one.
When the memory is full and the returned value is wrong, 
the oldest memory slot is replaced. 
As the rare class samples seldom update their 
$M(i)$'s, they have a higher chance of being erased.
	
\subsubsection{Refining Parameters}

Recall 
that the Learnet
\cite{bertinetto2016learning} and its variants 
(Section ~\ref{sec:model_embedding_meta})
map information from 
$\dtrain$ to parameterize the embedding function $g(\cdot)$ for a new $\xtest$.  
This parameter can 
be refined 
using a memory. 
Meta Networks (MetaNet) \cite{munkhdalai2017meta} parameterizes a classification model using a
``slow" weight which is meta-learned from multiple data sets, and a ``fast" weight
which is a task-specific embedding of $\dtrain$. 
As shown in \cite{munkhdalai2018rapid}, 
the computation cost of MetaNet can be reduced by
learning to modify each neuron rather 
the complete parameter.
MN-Net \cite{cai2018memory} uses a memory to refine the embedding learned
in the Matching Nets, whose output is used to parameterize a CNN as in Learnet.

\subsection{Generative Modeling}
\begin{figure*}[htb]
	\centering
	\includegraphics[width=0.65\linewidth]{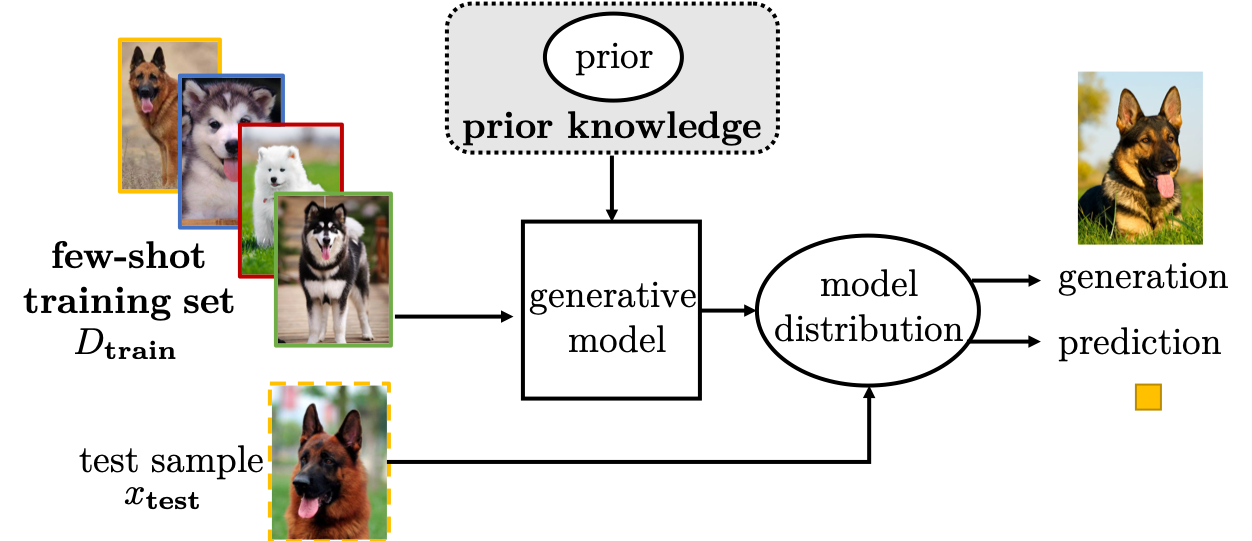}
	\caption{Solving the
		FSL problem by
		generative modeling. 
	}
	\label{fig:generative}
\end{figure*}

\textit{Generative modeling} methods estimate the probability distribution $p(x)$
from the observed $\xii$'s with the help of prior knowledge (Figure~\ref{fig:generative}).  Estimation of $p(x)$ usually involves estimations of
$p(x|y)$ and $p(y)$. 
Methods in this class can deal with many tasks, such as  generation \cite{reed2018fewshot,edwards2017towards,rezende2016one,lake2015human}, 
recognition
\cite{lake2015human,fei2006one,salakhutdinov2012one,torralba2011learning,edwards2017towards,gordon2018metalearning,zhang2018metagan},
reconstruction \cite{gordon2018metalearning}, and image flipping
\cite{reed2018fewshot}.

In generative modeling, the observed $x$ is assumed to be drawn from some distribution $p(x;\theta)$ parameterized by $\theta$. 
Usually, there exists a latent variable $z\sim p(z;\gamma)$, so that $x\sim \int p(x|z ;\theta)p(z;\gamma)dz$.  
The prior distribution $p(z;\gamma)$, which is 
learned from other data sets,
brings in prior knowledge that is vital to FSL. 
By combining the provided training set $\dtrain$
with this $p(z;\gamma)$, the resultant posterior probability distribution is constrained. 
In other words, $\Hs$ is constrained to a much smaller $\tilde{\Hs}$.

According to what the latent variable $z$ represents, we group these FSL generative
modeling methods into three  types.
\subsubsection{Decomposable Components}
Although samples with supervised information are scarce in a FSL problem, 
they
may share some
smaller decomposable components with samples from the other  tasks. 
For example, consider 
the recognition of
a person using only a few 
face 
photos provided.
Although similar faces may be hard to find, one can easily find photos with similar
eyes, noses or mouths. 
With a larger number of samples, models for these decomposable components can be easily learned. 
One then only needs to find the correct combination of these decomposable components, and decides which target class this combination belongs to. 
As the decomposable components are chosen by human, this strategy is more interpretable.
Bayesian One-Shot
\cite{fei2006one} uses a generative model to capture the interactions between decomposable components (i.e.,  shapes and appearances of objects) and target class (i.e., objects to be recognized).
Bayesian Program Learning (BPL)
\cite{lake2015human}  
models characters
by separating it into types, tokens and further templates, parts, primitives.
To generate a new character, one needs to search a large combination space containing theses components.  
In \cite{lake2015human}, this inference cost is reduced by only considering the top
possible combinations.
In natural language processing, a recent work \cite{joshi2018extending} models spans instead of
the complete parse tree, and 
adapts
parsers between syntactically distant domains by 
training individual classifiers for spans.

\subsubsection{Groupwise Shared
Prior}
Often, similar tasks have similar prior probabilities, and
this can be utilized in FSL.  For example,
consider the three-class classification of
``orange cat", ``leopard" and ``Bengal tiger". 
These three species  are similar, but
Bengal tiger is endangered, while orange cats and leopards are abundant.
Hence, one can learn a prior probability from 
``orange cats" and ``leopards",
and use this as the prior for the few-shot class
``Bengal tiger".

In \cite{salakhutdinov2012one},  a set of data sets $\{\D_c\}$ are grouped into a hierarchy via unsupervised learning.
Data sets in each group together learn the class prior probabilities.
For a new few-shot class, one first finds the group this new class belongs to, and
then models it by the class prior
drawn from the groupwise shared prior probability.
In \cite{torralba2011learning},
the feature learning step in
\cite{salakhutdinov2012one} 
is further improved 
with the use of deep Boltzmann machines \cite{salakhutdinov2009deep}.

\subsubsection{Parameters of Inference Networks}
To find the best $\theta$, 
one has to  maximize the posterior
\begin{align}\label{eq:bayes}
p(z|x;\theta,\gamma) = \frac{p(x,z;\theta,\gamma)}{p(x;\gamma)}
=\frac{p(x|z;\theta)p(z;\gamma)}{\int p(x|z;\theta)p(z;\gamma)dz}.
\end{align}
Due to the integral in the denominator, it is intractable to solve \eqref{eq:bayes}. 
A variational distribution $q(z;\delta)$, which  is learned from data, is often used to approximate $p(z|x;\theta,\gamma)$. 
Recently, this $q(z;\delta)$ is approximated via
amortized variational inference with the inference network \cite{zhang2018advances}. 
Although $z$ no longer has semantic meaning, the powerful representation learned by
these deep models can lead to better performance.
Once learned, the inference network can be applied to a new task directly, which is more efficient and requires less human knowledge.
As the inference network has a large number of parameters, it is usually trained using
some auxiliary large-scale data sets.  Many classic inference networks are adapted
to the FSL problem. 
For example, the variational auto-encoder (VAE) \cite{kingma2014auto} is used in \cite{rezende2016one,edwards2017towards,hewitt2018variational}, 
 autoregressive model \cite{van2016conditional} is used in \cite{reed2018fewshot}, 
 generative adversarial networks (GAN) \cite{goodfellow2014generative} is used in \cite{zhang2018metagan}, and a combination of VAE and GAN is proposed in \cite{gordon2018metalearning}.

\subsection{Discussion and Summary}
\label{sec:model_sum}

When there exist similar tasks or auxiliary tasks, 
multitask learning can be used to constrain the $\Hs$ of the few-shot task.
However, note that
joint training of all the tasks  together is required. Thus,
when a new few-shot task arrives,
the whole multitask model has to be trained again,
which can be costly and slow. Moreover,
the sizes of $D$ and $\D_c$ should not comparable, 
otherwise, the few-shot task may be overwhelmed by tasks with many samples. 

When there exist a large-scale data set containing sufficient samples of various
classes, 
one can use
embedding learning
methods.
These methods map samples 
to a good embedding space
in which samples from different classes can be 
well-separated,
and so a smaller $\tilde{\Hs}$ is needed. 
However,
they may not work well  
when the few-shot task 
is not closely related to 
the other tasks.
Moreover, more exploration on 
how to mix the invariant and task-specific information of tasks 
is helpful.

When 
a 
memory network is available, it can be readily used for FSL by training a simple
model (e.g., classifier) on top of the memory.
By using carefully-designed update rule, one can selectively protect memory slots.
The weakness of this strategy is that it incurs additional space and computational
costs, which increase with memory size.
Therefore, current external memory has a limited size.

Finally, when one wants to perform tasks such as generation and reconstruction
besides FSL,
generative models can be used.
They learn prior probability
$p(z;\gamma)$ 
from the other data sets, which reduces $\Hs$ to a smaller $\tilde{\Hs}$.
The learned generative models can also be used to generate samples for data
augmentation.
However, generative modeling methods have high inference cost, and are more difficult to
derive than deterministic models.

\section{Algorithm}
\label{sec:alg}
The algorithm is the strategy to search in the hypothesis space $\Hs$ for the parameter $\theta$ of the best hypothesis $h^*$ \cite{bottou2008tradeoffs,bottou2018optimization}. 
At the $t$th iteration, 
$\theta_{t} = \theta_{t-1}+\Delta \theta_{t-1}$, where $\Delta \theta_{t-1}$ is the update. 
For example, for the popular stochastic gradient descent (SGD) and its variants \cite{bottou2008tradeoffs,bottou2018optimization}, $\theta$ is updated as 
\begin{align}
\label{eq:ll_obj}
\theta_{t}=\theta_{t-1}-\alpha_t
\nabla_{\theta_{t-1}} 
\ell(h(x_t;\theta_{t-1}),y_t), 
\end{align}
where $\alpha_t$ is the stepsize. 
With 
$\theta$
initialized at
$\theta_0$, $\theta_t$ can be written as 
\begin{align}
\label{eq:alg}
\theta_{t}=\theta_0+\sum_{i=1}^{t}\Delta\theta_{i-1}.
\end{align}
When supervised information is rich, 
there are enough training samples to update $\theta$, and to find an appropriate
stepsize $\alpha$ by cross-validation.
However, in FSL, the provided few-shot $\dtrain$ is not large enough, and 
the obtained empirical risk minimizer
is unreliable.

Methods in this section 
use prior knowledge to influence how $\theta$ is obtained, either by   (i)
providing a good initialized parameter $\theta_0$, or (ii) directly learning an optimizer to output search steps.  
In terms of how the search strategy is affected by prior knowledge, we classify
methods in this section into three groups
(Table~\ref{tab:alg_aux}):
\begin{enumerate}
	\item \textit{Refining existing parameters}. An initial $\theta_0$ learned
	from other tasks, and is then refined using $\dtrain$.
	
	\item \textit{Refining meta-learned parameters}.  
	An initial $\theta_0$ is meta-learned from a set
	of tasks, which are drawn from the same task distribution as the few-shot task, 
	and then further refined by the learner using $\dtrain$.
	
	\item \textit{Learning the optimizer}.   
	This strategy learns a meta-learner as optimizer to output search steps for each learner directly, such as
	changing the search direction or stepsize. 
\end{enumerate}
\begin{table}[ht]
	\caption{Characteristics for FSL methods focusing on the algorithm perspective.}
	\footnotesize
	\begin{tabular}
		{c|c|c}
		\hline
		strategy  &     prior knowledge             & how to search $\theta$ of the $h^*$ in $\Hs$   \\\hline		
		refining existing parameters     &  learned $\theta_0$   & refine $\theta_0$ by $\dtrain$    \\ \hline
		refining meta-learned parameters &      meta-learner  & refine $\theta_0$ by $\dtrain$ \\ \hline
		learning the optimizer    &   meta-learner     & use search steps provided by the meta-learner \\ \hline
	\end{tabular}
	\label{tab:alg_aux}
\end{table}

\subsection{Refining Existing Parameters}
\label{sec:alg_fine}

This strategy takes $\theta_0$ of a pre-trained model learned from related tasks
as a good initialization, and adapts it to $\theta$ by $\dtrain$.
The assumption is that $\theta_0$ captures some general structures of the
large-scale data. Therefore, it can be 
adapted 
to $\D$
with a few iterations.

\subsubsection{Fine-Tuning Existing Parameter by Regularization}
\begin{figure}[ht]
	\centering
	\includegraphics[width=0.70\linewidth]{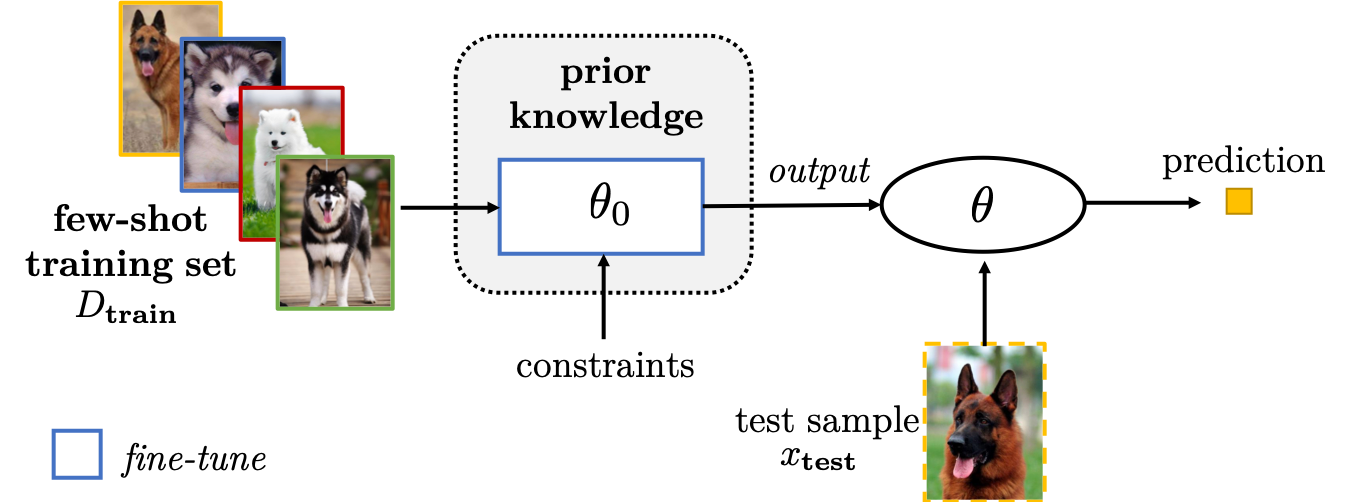}
	\caption{Solving the
	FSL problem by
	fine-tuning existing parameter $\theta_0$ by regularization.  
	}
	\label{fig:alg_finetune}
\end{figure}

This strategy fine-tunes the pre-trained $\theta_0$ for the few-shot task by
regularization
(Figure~\ref{fig:alg_finetune}),
and is popularly used
in practice. 
In \cite{caelles2017one}, a CNN pre-trained on the ImageNet for image classification is tuned using a large data set for foreground segmentation, then further fine-tuned using a single shot of segmented object for object segmentation.
Given the few-shot $\dtrain$, 
simply fine-tuning $\theta_0$ by gradient descent may lead to overfitting. 
Hence, how to adapt $\theta_0$ without overfitting to $\dtrain$ is a key design issue.

In this section, methods fine-tune $\theta_0$ by regularization to prevent overfitting.
They can be grouped as follows: 
\begin{enumerate}
	\item \textit{Early-stopping}.
	It 
	requires separating a validation set from $\dtrain$ to monitor the training
	procedure. Learning is stopped when there is no performance improvement on the validation set 
	\cite{arik2018neural}.  
	
	\item
	\textit{Selectively updating $\theta_0$}. Only a portion
	of $\theta_0$ is updated in order to avoid overfitting. 
	For example, in \cite{keshari2018learning}, given a set of pre-trained filters,  
	it only learns a 
	strength parameter that is multiplied with the filters.
	
	\item \textit{Updating related parts of $\theta_0$ together}. 
	One can group elements of $\theta_0$ (such as the
	neurons in a deep neural network),  
	and update each group
	jointly with the same update information. 
	In \cite{yoo2018efficient}, 
	the filters of a pre-trained CNN are clustered together according to some auxiliary information, and then fine-tuned by groupwise back-propagation using $\dtrain$.
	
	\item \textit{Using a model regression network}. 
	A model regression network \cite{wang2016learninga} 
captures the task-agnostic transformation which maps the parameter values obtained
by training on
a few examples to the parameter values that will be obtained by training on a lot
of samples.
	Similarly, in \cite{kozerawski2018clear}, the transformation function that maps the embedding of $\xii$ to a classification decision boundary is learned.
	
\end{enumerate}

\subsubsection{Aggregating a Set of Parameters}
\begin{figure}[ht]
	\centering
	\includegraphics[width=0.70\linewidth]{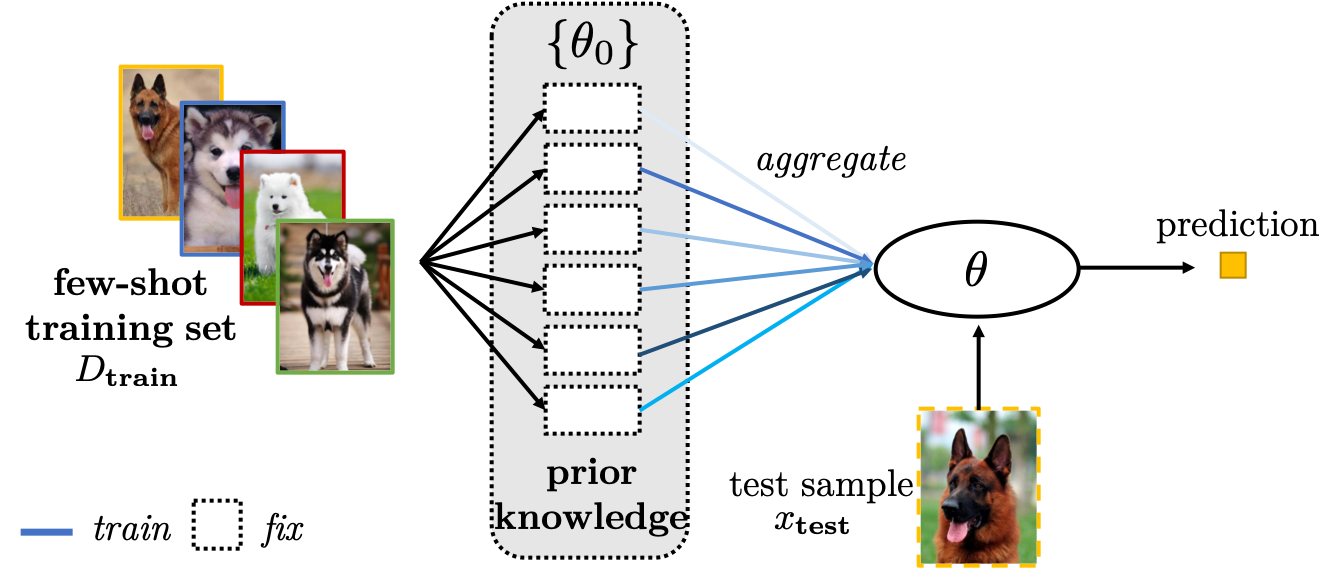}
	\caption{Solving the
	FSL problem by
	aggregating a set of parameters $\theta_0$'s into $\theta$.
	Provided with a set of pre-trained $\theta_0$'s, one only needs to learn the
	combination weights (blue lines).
	}
	\label{fig:alg_pick}
\end{figure}

Sometimes, we do not have a suitable $\theta_0$ to start with.
Instead, we have many models  
that are learned from related tasks.
For example, in 
face recognition,
we may already have recognition models for the eye, nose, and ear. 
Therefore, one can 
aggregate these model parameters to a
suitable model, which is then either directly used or refined by $\dtrain$
(Figure~\ref{fig:alg_pick}).

As discussed in Section~\ref{sec:data}, 
samples from unlabeled data sets (Section~\ref{sec:data_unlabel})
and similar
labeled data sets (Section~\ref{sec:data_sim})
can be used to augment the few-shot $\dtrain$. 
Instead of using the samples directly, the following methods use
models (with parameters $\theta_0$'s)
pre-trained 
from these data sets.
The problem is then how to adapt them efficiently to the new task using $\dtrain$.

\begin{enumerate}
	
\item \textit{Unlabeled
data set}.
Although there is no 
supervised information, similar samples can be grouped
together. 
Therefore, one can pre-train functions
from the unlabeled data to cluster and separate samples well. 
A neural network 
is then used 
to 
adapt
them to the new task with the few-shot $\dtrain$ \cite{wang2016learninga,wang2016learningc}.

\item \textit{Similar data sets}. 
In \cite{bart2005cross}, 
few-shot object classification 
is performed by leveraging samples and classifiers from similar classes. 
First,
it 
replaces the features of samples from these similar classes by features from the 
new class. 
The learned classifier is then reused,
and only 
the classification threshold 
is adjusted 
for the new class. 
In \cite{gidaris2018dynamic,yu2018diverse}, they learn to combine existing parameters learned from similar data sets using $\dtrain$. 

\end{enumerate}

\subsubsection{Fine-Tuning Existing Parameter with New Parameters}

\begin{figure}[ht]
	\centering
	\includegraphics[width=0.7\linewidth]{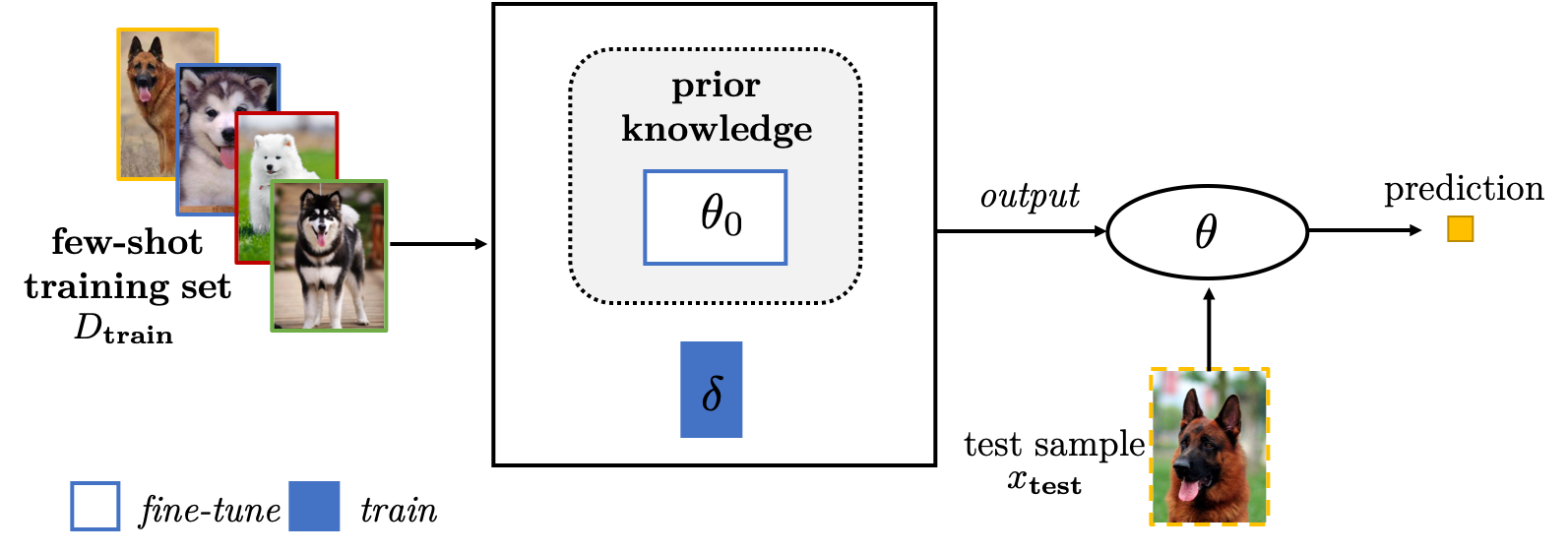}
	\caption{Solving the 
	FSL problem by
	fine-tuning existing parameter $\theta_0$ with new parameters. 
	}
	\label{fig:alg_joint}
\end{figure}

The pre-trained $\theta_0$ may not be enough to encode
the new FSL task completely. 
Hence, an additional parameter(s) $\delta$ 
is used to take the specialty of $\dtrain$ into account (Figure~\ref{fig:alg_joint}).
Specifically, this strategy expands 
the model parameter to become $\theta=\{\theta_0,\delta\}$,
and fine-tunes $\theta_0$ while learning $\delta$.
In \cite{hoffman2013one}, it uses the lower layers of a pre-trained
CNN for feature embedding, and learns a linear classifier on the embedded features using $\dtrain$. 
In font style transfer
\cite{azadi2018multi},
a network is first
pre-trained  to capture the fonts in gray images. 
To generate stylish colored fonts,
this is fine-tuned together with the training of an additional network.

\subsection{Refining Meta-Learned Parameter}
\label{sec:alg_refine_meta}
\begin{figure}[htb]
	\centering
	\includegraphics[width=0.9\linewidth]{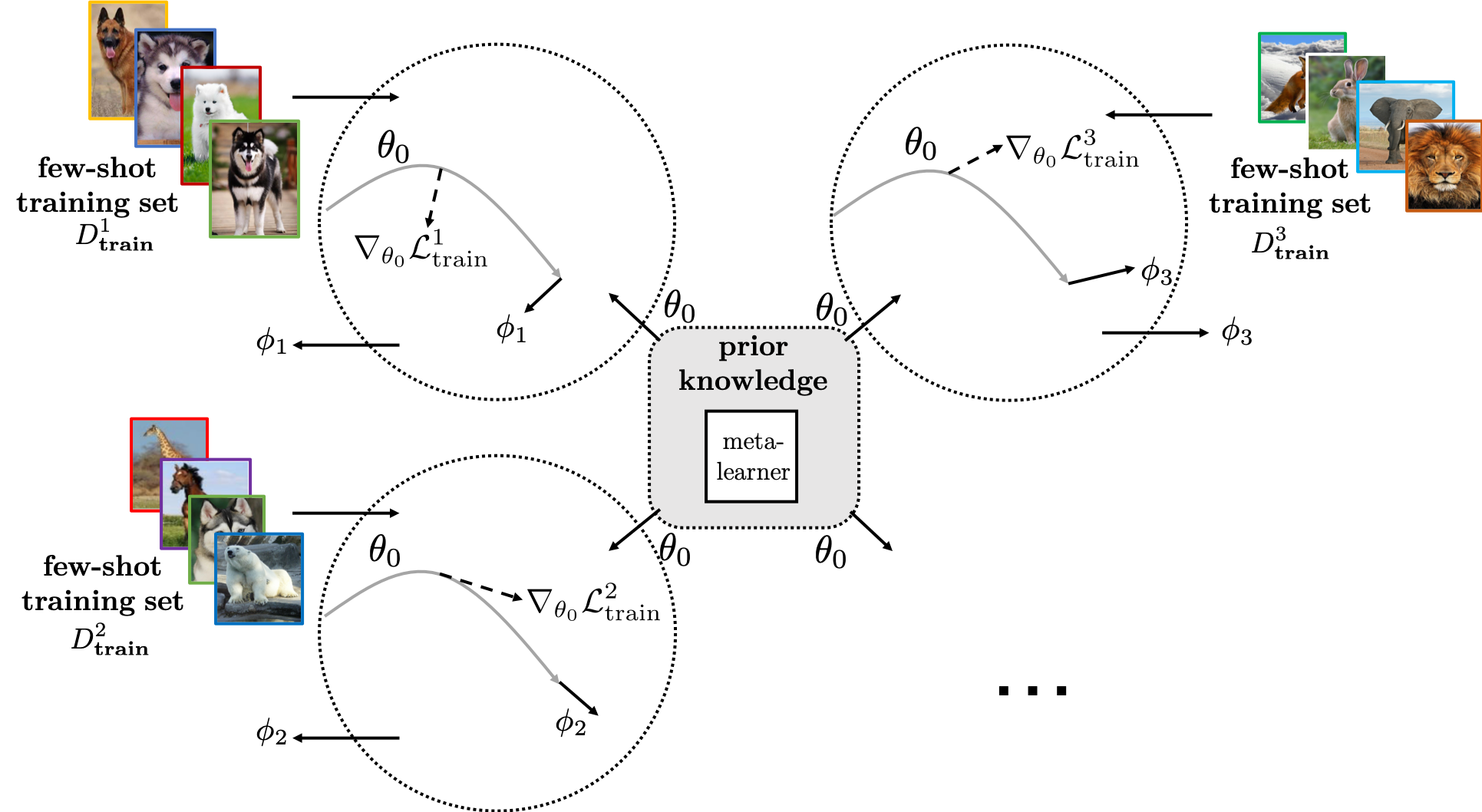}
	\caption{Solving the 
		FSL problem by
		refining the meta-learned parameter $\theta_0$.}
	\label{fig:alg_meta_para}
\end{figure}

Methods in this section use meta-learning to refine the meta-learned parameter
$\theta_0$
	(Figure~\ref{fig:alg_meta_para}). 
The $\theta_0$
is continuously optimized by the meta-learner according to performance of the learner.
This is different from Section~\ref{sec:alg_fine}
in which 
$\theta_0$ is 
fixed.

The meta-learned $\theta_0$ is often refined by gradient descent. A representative
method is the
Model-Agnostic Meta-Learning (MAML) \cite{finn2017model}.
It meta-learns $\theta_0$, 
which 
is then adjusted 
to obtain a good task-specific parameter $\phi_s$ for some $\T_s\sim P(T)$
via a few effective gradient descent
steps,
as:
$\phi_s=\theta_0-\alpha\nabla_{\theta_0} \mL_{\text{train}}^s(\theta_0)$.
Here,
$\mL_{\text{train}}^s(\theta_0)$ is the sum of losses over the training samples in $\dtrain$,
and
$\alpha$ is the stepsize.
Note that $\phi_s$ is
invariant 
to permutation of the samples.
The meta-learned parameter $\theta_0$ is updated by feedbacks from multiple meta-training tasks as
$\theta_0 \leftarrow \theta_0-\beta
\nabla_{\theta_0} \sum_{\T_s\sim P(\T)}\mL_{\text{test}}^s(\theta_0)$, 
where
$\mL_{\text{test}}^s(\theta_0)$ is the sum of losses over the test
samples in $\dtest$ and 
$\beta$ is another  stepsize. 
By continuously refining $\theta_0$ using the few-shot samples in
$\dtrain$, the meta-learner improves its $\theta_0$ to quickly adapt to the few-shot training set.

Recently, 
many improvements 
have been proposed for
MAML,
mainly along the following three aspects: 
\begin{enumerate}
\item \textit{Incorporating task-specific information}. MAML provides the same initialization for all tasks. 
However, this 
neglects task-specific information, and
is appropriate only when the set of tasks are all very similar. To address this problem, 
in \cite{lee2018gradient}, it learns to choose 
	$\{\theta_0\}$ 
	from a subset of 
	a good initialization parameter 
	for a new task.

	\item \textit{Modeling the uncertainty of using a meta-learned $\theta_0$}. 
	 Learning with a few examples inevitably results in a model with higher
	 uncertainty \cite{finn2018probabilistic}. Hence, 
	the learned model may not be able to perform prediction on the new task with high confidence. 
	The ability to measure this uncertainty
	provides hints for active learning and further data collection \cite{finn2018probabilistic}.
	There are works that consider uncertainty for the meta-learned $\theta_0$ \cite{yoon2018bayesian,finn2018probabilistic}, 
uncertainty for the task-specific $\phi_s$ \cite{ravi2018amortized,grant2018recasting}, and 
	uncertainty for class $n$'s class-specific parameter $\phi_{s,n}$ \cite{rusu2018metalearning}.
	
	\item \textit{Improving the refining procedure}. 
	Refinement by a few gradient descent steps may not be reliable. Regularization
	can be used to correct the descent direction. 
	In \cite{gui2018few},
	the model regression network \cite{wang2016learninga}
	is used to
	regularize
	task $\T_s$'s
	$\phi_s$ 
	to be close to the model trained with large-scale samples.
	
\end{enumerate}

\subsection{Learning the Optimizer}
\label{sec:alg_search_step}

\begin{figure}[ht]
	\centering
	\includegraphics[width=0.7\linewidth]{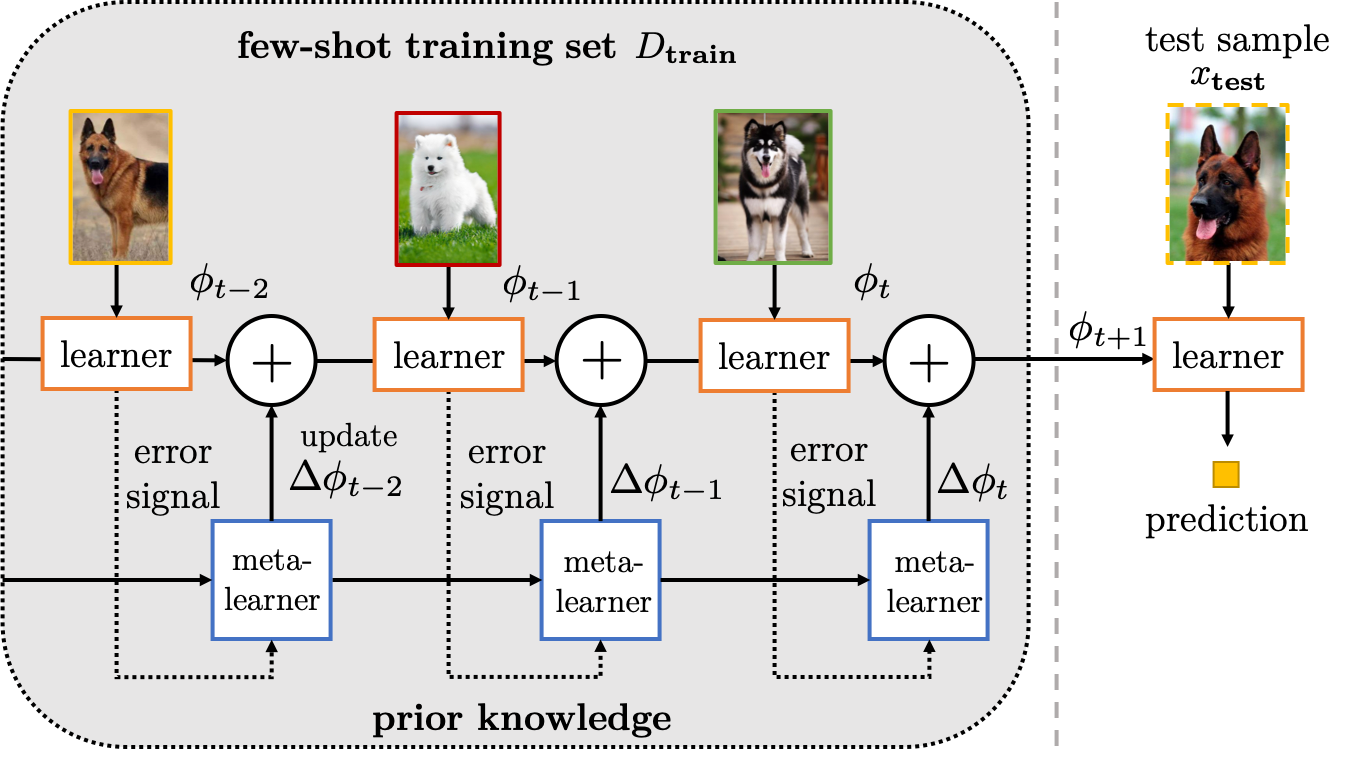}
	\caption{Solving the
	FSL problem
	by learning the optimizer.}
	\label{fig:alg_opt}
\end{figure}

In Section~\ref{sec:alg_refine_meta},
the meta-learner $\theta_0$ acts as 
a good
initialization  for $\T \sim P(T)$ with data $\D$, and it is adjusted to a task-specific parameter $\phi$ via 
a few effective gradient descent
steps.
In contrast, instead of using gradient descent, methods in this section
learns an optimizer which can directly output the update 
($\sum_{i=1}^{t}\Delta\theta^{i-1}$ in \eqref{eq:alg}) (Figure~\ref{fig:alg_opt}). 
There is then no need to tune the stepsize $\alpha$ or find the search
direction, as the learning algorithm does that automatically. 

At the $t$th iteration, this line of works \cite{andrychowicz2016learning,ravi2017optimization} learn an meta-learner 
which 
takes the error signal computed at the $(t-1)$th iteration, and directly outputs
update
$\Delta \phi_{t-1}$ to update the task-specific 
parameter $\phi_{t-1}$ of the learner as $\phi_t=\phi_{t-1}+\Delta \phi_{t-1}$. 
Therefore, in contrast to strategies mentioned in Sections~\ref{sec:alg_fine} and ~\ref{sec:alg_refine_meta}, this strategy provides an optimizer for the new task (which is in turn optimized by the learner). 
This $\phi_{t}$ is then used to compute the loss
$\ell_{t}(\phi_{t})=\ell(h(x_t;\phi_{t}),y_t)$ using the $t$th sample $(x_t,y_t)\in \dtrain$,
which acts as the error signal to be fed into the meta-learner at the next iteration. 
After learning a task, the meta-learner is improved by 
gradient descent
on the loss on the test set $\dtest$.
By learning from a set of $\T_s$'s drawn from $P(\T)$, the meta-learner improves on proposing efficient algorithms for FSL. 
Recently, \cite{ravi2017optimization} obtains $\phi_{t}$ by
instantiating
\eqref{eq:ll_obj} with the cell state update in the LSTM (where 
$\phi$ is set to the cell state of the LSTM).

\subsection{Discussion and Summary}
\label{sec:alg_sum}

Refining existing parameters can reduce the 
search effort in $\Hs$.
By using an existing $\theta_0$ as initialization, these methods usually need a
lower computation cost to obtain a good hypothesis $h\in\Hs$.
Learning focuses on refining these existing parameters. 
However, as $\theta_0$ is learned from tasks different from the current task, 
this strategy may sacrifice 
precision for speed. 

The other two strategies rely on meta-learning.
By learning from a set of related tasks, 
the meta-learned  $\theta_0$ can be closer to the task-specific parameter
$\phi_t$ for a new task $\T_t$.
Learning search steps by a meta-learner 
can directly guide the learning algorithm. 
In other words, the meta-learner acts as an optimizer. 
However, important issues such as how to meta-learn across
different granularities (such as coarse-grained classifications of animals versus fine-grained classification of dog species) 
or different data sources (such as images versus texts) \cite{triantafillou2019meta}
are still open. 
From this perspective, meta-learning and multi-tasks are similar, and so there is
also a concern on 
how to 
avoid negative transfer
\cite{deleu2018effects}

\section{Future Works}
\label{sec:future}

In this section, we discuss four key directions for the further development of FSL,
namely, (i) problem setups, (ii) techniques, (iii) applications and (iv) theories.

\subsection{Problem Setups}
Existing FSL methods often use prior knowledge from one single modality 
(such as images, texts, or videos).
However, though $\dtrain$ has a few examples for the modality currently used, there may exist
another modality in which 
supervised samples
are abundant.
An example is in the study of extinct animals. While this animal species may only
have a limited number of visual examples, there might be a lot of information
about it
in the textual domain (such as textbooks
or web pages), as people tend to pay special attention to the rare class. 
Therefore, 
prior knowledge 
from multiple modalities
can 
provide prior knowledge for complementary views. 
In zero-shot learning (ZSL),
multi-modality data has been 
frequently
used.
Example prior information are 
attributes
\cite{hwang2014unified,akata2013label}, 
WordNet 
\cite{hwang2014unified,akata2013label}, 
word embeddings 
\cite{tsai2017improving,wang2017multi}, 
co-occurrence statistics 
\cite{mensink2014costa}, 
and 
knowledge graphs 
\cite{wang2018zero}.

Recently, there have been efforts in borrowing techniques from ZSL methods
to FSL problems. 
For example, one can use the few-shot $\dtrain$ to fine-tune the parameters learned by ZSL methods \cite{hwang2014unified,akata2013label}. 
However, fine-tuning using a small number of samples may lead to overfitting.
Another possibility is to force the embedding learned by multiple modalities to match in a shared space \cite{tsai2017improving,wang2017multi}. 
A recent work \cite{rios2018few} exploits
the structured relationships among labels and utilizes a GNN  to align the embedding for FSL. 
As different modalities may contain different structures, this should be carefully handled.
For example, texts need to obey syntactic structures while images do not. 
In the future, 
a promising direction is to 
consider the use of multi-modality information in
designing FSL methods.

\subsection{Techniques}
\label{sec:future_tech}

In previous sections, 
according to how the prior knowledge in FSL is used,
we 
categorize
FSL methods from the perspectives of
data (Section~\ref{sec:data}), 
model (Section~\ref{sec:model}), and algorithm (Section~\ref{sec:alg}).
Each of these components can be 
improved.
For example, using 
state-of-the-art
ResNet \cite{he2016deep} as the embedding function can be better than using the VGG \cite{srivastava2015training}.

Meta-learning-based FSL methods, as reviewed in Sections~\ref{sec:model} and \ref{sec:alg},
are particularly interesting.
By learning across tasks, 
meta-learning 
can adapt to new tasks rapidly with a small inference cost. 
However, the tasks considered in meta-learning are often assumed to be 
drawn from a single task distribution $p(\T)$. 
In practice, we can have a large number of tasks 
whose task relatedness is unknown 
or 
expensive to determine.
In this case, directly learning from all these tasks 
can lead to negative transfer \cite{deleu2018effects}. 
Besides, current FSL methods often
consider a static and fixed $P(\T)$ \cite{finn2017model,ravi2017optimization}.
However, in streaming 
applications, $p(\T)$ is dynamic \cite{finn2018meta} and
new tasks are continually arriving.  Hence,
this 
should also be incorporated 
into $p(\T)$. 
An important issue is 
how to avoid catastrophic forgetting \cite{kirkpatrick2017overcoming} in a
dynamic setting, which means that information on the old tasks should not be forgotten. 

As discussed in previous sections, different FSL methods have pros and cons, and there is no absolute winner in all settings.
Moreover, both the hypothesis space $\Hs$ and search strategies in $\Hs$ often rely on human design. 
\textit{Automated machine learning} (AutoML) \cite{quanming2018taking},
by constructing task-aware machine learning models,  
has achieved state-of-the-art on many applications. 
Recently, AutoML has been used on data augmentation \cite{cubuk2019autoaugment}. 
Another direction is to extend 
the AutoML methods of
automated feature engineering \cite{kanter2015deep}, model selection \cite{kotthoff2017auto} and neural architecture search \cite{zoph2017neural} 
to FSL.
One can then obtain better algorithm designs whose components are learned by
AutoML in an economic, efficient and effective manner.

\subsection{Applications} 

Recall that FSL is needed due to rareness of samples, endeavor to reduce data
gathering effort and computational cost, or as a stepping stone to mimic human-like learning. 
Hence, 
many real-world applications involve FSL.
Computer vision is one of very first testbed for FSL algorithms.  
FSL 
has also attracted a lot of recent attention
in many other applications, such as
robotics, natural language processing, and acoustic signal processing.
In summary, there are many interesting fields and applications for FSL to explore.

\subsubsection{Computer Vision} 
Most
existing works target FSL problems in computer vision. 
The two most popular applications are 
character recognition    \cite{fink2005object,santoro2016meta,vinyals2016matching,munkhdalai2017meta,finn2017model,woodward2017active,kaiser2017learning,koch2015siamese,triantafillou2017few,bertinetto2016learning,shyam17attentive,salakhutdinov2012one,snell2017prototypical}
and 
image classification  \cite{ravi2017optimization,finn2017model,vinyals2016matching,xu2017few,munkhdalai2017meta,shyam17attentive,tang2010optimizing,triantafillou2017few,koch2015siamese,wang2016learningc,wang2016learninga,hubert2017learning,snell2017prototypical}.
Very high accuracies have already been obtained on  the
standard benchmark
data sets
(such as Ominiglot and miniImageNet),
leaving little space for further improvement
\cite{triantafillou2019meta}.  
Recently,
a large and diverse benchmark data set, constructed from multiple image data
sources, is  presented in \cite{triantafillou2019meta}.
Besides character recognition    and image classification,
other image applications have also been considered. These include
object recognition \cite{fink2005object,fei2006one,liu2018feature}, 
font style transfer \cite{azadi2018multi}, 
phrase grounding \cite{zhao2018dynamic},
image retrieval \cite{triantafillou2017few}, 
object tracking \cite{bertinetto2016learning}, 
specific object counting in images \cite{zhao2018dynamic}, 
scene location recognition \cite{kwitt2016one},
gesture recognition \cite{pfister2014domain}, 
part labeling \cite{choi2018structured}, 
image generation \cite{lake2015human,reed2018fewshot,rezende2016one,edwards2017towards}, 
image translation across domains \cite{benaim2018one}, 
shape view reconstruction for 3D objects \cite{gordon2018metalearning}, 
and 
image captioning and visual question answering \cite{dong2018fast}.

FSL has also been successfully used in video applications, including 
motion prediction \cite{gui2018few}, 
video classification \cite{zhu2018compound}, 
action localization \cite{yang2018one}, 
person re-identification \cite{wu2018exploit}, 
event detection \cite{yan2015multi}, 
and object segmentation \cite{caelles2017one}.

\subsubsection{Robotics}
In order for
robots 
to behave more like human, 
they should be able to generalize from a few demonstrations.  
Hence, FSL has played an important role in robotics. 
For example, 
learning of robot arm movement 
using imitating learning from
a single demonstration \cite{wu2010towards},  and
learning manipulation actions from a few demonstrations with the help of a teacher
who corrects the false actions \cite{abdo2013learning}. 

Apart from imitating users, robots can improve their behavior through interacting with users. 
Recently, 
assistive strategies 
are learned 
from a few interactions through FSL reinforcement learning \cite{hamaya2016learning}. 
Other examples of FSL in robotics include
multi-armed bandits \cite{duan2017one}, 
visual navigation \cite{duan2017one,finn2017model}, and continuous control \cite{finn2017model,yoon2018bayesian,mishra2018a}.
Recently, 
these applications are further extended to
dynamic environments \cite{al2018continuous,nagabandi2018deep}.

\subsubsection{Natural Language Processing}

Recently, the use of FSL has drawn attention in natural language processing.
Example applications include parsing \cite{joshi2018extending}, translation \cite{kaiser2017learning}, 
sentence completion (which fills in the blanks using a word chosen from a provided
set) \cite{vinyals2016matching,munkhdalai2018rapid}, 
sentiment classification from short reviews \cite{yu2018diverse,yan2018few}, user
intent classification for dialog systems \cite{yu2018diverse},  criminal charge prediction \cite{hu2018few}, word similarity tasks such as nonce definition \cite{herbelot2017high,sun2018memory}, and multi-label text classification \cite{rios2018few}. 
Recently, 
a new 
relation classification 
data set called FewRel \cite{han2018fewrel} 
is released. This
compensates for the lack of benchmark data set for FSL tasks in natural language processing.

\subsubsection{Acoustic Signal Processing} 
Apart from the early efforts on using FSL to recognize spoken words from
one example \cite{lake2014one}, recent endeavors are on voice synthesis.
A popular task is voice cloning from a few audio samples of the user \cite{arik2018neural}. 
This can be useful in generating personal voice navigation in map applications, or
mimicking the parents' voice in story-telling to kids in a smart home toolkit.
Recently, 
it is possible to perform 
voice conversion from one user to another using one-shot voice or text sample \cite{tjandra2018machine} or even across different languages \cite{mohammadi2018investigation}. 

\subsubsection{Others} 
For example,
a recent
attempt
in the context of medical applications
is  
few-shot drug discovery 
\cite{altae2017low}. 
For learning of deep networks,
one-shot architecture search (OAS) 
is studied in
\cite{brock2018smash,liu2018darts,yao2020efficient}. 
Unlike random search and grid search which require multiple runs to find the best architecture, OAS methods can find good architectures by training the supernet once. 
FSL has also been used in 
curve fitting \cite{yoon2018bayesian,finn2018probabilistic,grant2018recasting,santoro2016meta} and 
understanding
number analogy by logic reasoning to perform calculations \cite{ramalho2018adaptive}.

\subsection{Theories}
FSL uses prior knowledge to compensate for the lack of supervised information. 
This is related to 
the theoretical study of
sample complexity,
which is the number of training samples needed to obtain a model with
small empirical risk $R_I(h)$ 
having high probability \cite{tom1997machine,mohri2018foundations}.
$\Hs$ needs to be less complicated to make the provided $I$ samples enough.
Recall that 
FSL methods use prior knowledge to augment more samples (i.e., increase $I$),
constrain $\Hs$ (i.e., reduce the complexity of $\Hs$) and alter the search
strategy (i.e., increase the probability of finding a good $h$).
This suggests that 
FSL methods 
can reduce the required sample complexity using prior knowledge. 
A detailed analysis on this aspect will be useful.

Besides, 
recall that FSL is related to domain adaptation
\cite{pfister2014domain,motiian2017few,luo2017label}, and
existing theoretical bounds on 
domain adaptation 
may be inspiring \cite{ben2007analysis,blitzer2008learning}. 
For example, recent analysis shows that 
better risk bounds can be obtained by fine-tuning feedforward neural networks \cite{mcnamara2017risk}. 
By considering a specific meta-learning method,
the risk of transferring a model trained on one task to 
another task is examined in \cite{denevi2018learning}.
However, only a small number of methods 
have been studied so far. There are still a lot 
of theoretical issues
to explore.

Finally, convergence of the FSL algorithms is not fully understood. 
In particular, meta-learning methods optimize $\theta$ over a task distribution 
instead of over a single task.
Recent analysis in \cite{franceschi2018bilevel} provides sufficient conditions for
convergence of one meta-learning method. 
The meta-learner learns the lower layers of a deep network, while the learner 
learns the last layer, all using gradient descent. 
A more general analysis on the convergence of meta-learning methods will be highly 
useful.

\section{Conclusion}
\label{sec:conclusion}

Few-Shot Learning (FSL) targets at
bridging the gap between AI and human learning.
It can learn new tasks containing only a few examples with supervised information by incorporating prior knowledge.
FSL acts as a test-bed for AI, makes the learning of rare cases possible, or helps to relieve the burden of collecting 
large-scale supervised date in industrial applications.
In this survey, we provide a comprehensive and systematic review of FSL.
We first formally define FSL, and 
discuss the relatedness and differences of FSL with relevant learning problems such as weakly supervised learning, imbalanced learning, transfer learning and meta-learning. 
We 
then
point out the core issue of FSL 
is the unreliable empirical risk minimizer that makes FSL hard to learn. 
Understanding the core issue
helps categorize different works into data, model and algorithm according to how they solve the core issue using prior knowledge:
data augments the supervised experience of FSL, model constrains the hypothesis
space of FSL to be smaller, and algorithm alters the search strategy for the best
hypothesis in the given hypothesis space.
In each
category,
 the pros and cons are thoroughly discussed and 
some summary and insights are presented. 
To inspire future research in FSL,
we also provide possible directions on problem setups, techniques, applications and theories to explore.

\appendix
\section{Appendix: Meta-Learning}
\label{app:meta} 
\begin{figure}[ht]
	\centering
	\includegraphics[width=1\linewidth]{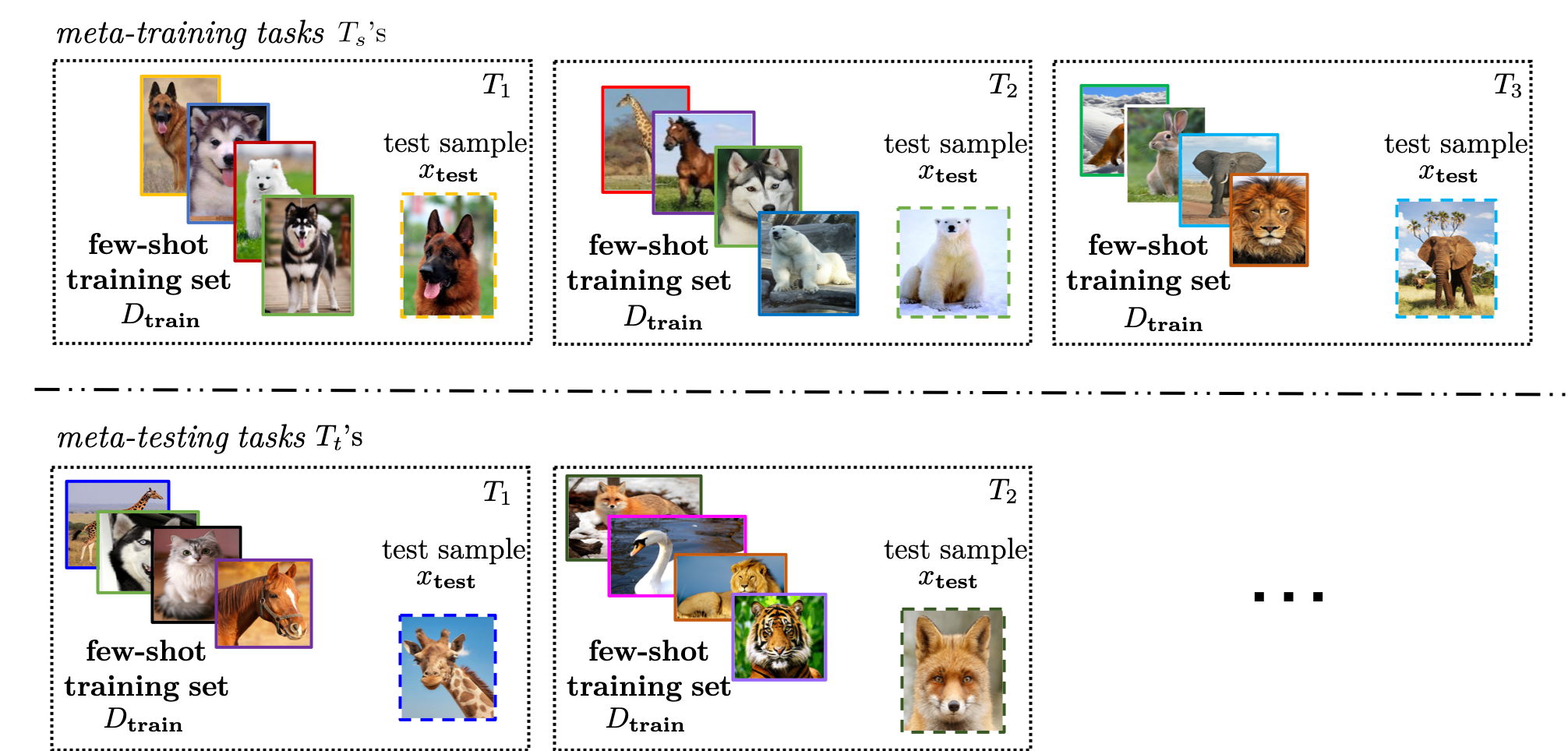}
	\caption{Solving the 
	FSL problem
	by meta-learning.}
	\label{fig:meta}
\end{figure}

Meta-learning \cite{hochreiter2001learning} 
improves $P$ of the new task $\T$ by the provided data set and the meta-knowledge extracted across tasks by a meta-learner (Figure~\ref{fig:meta}).
Let $p(\T)$ be the distribution of task $\T$.
In mete-training, it learns from a set of tasks $\T_s \sim p(\T)$. 
Each task $\T_s$ operates on data set $\D_s$ of $N$ classes, where $\D_s = \{\dtrain^s,\dtest^s\}$ 
consists of a training set $\dtrain^s$ and a test set $\dtest^s$. 
Each learner learns from 
$\dtrain^s$ and measures the test error on $\dtest^s$. 
The parameter $\theta_0$ of meta-learner is optimized to minimize the error across
all learners, as:
\begin{align*}
\theta_0 = \arg\min_{\theta_0}\mathbb{E}_{\T_s\sim p(\T)}
\sum\nolimits_{(\xyi)\in\D_s}
\ell(h(\xii;\theta_0),\yii).
\end{align*}

In meta-testing, another disjoint set of tasks $\T_t\sim p(\T)$ is used to test the generalization ability of the meta-learner.
Each $\T_t$ works on a data set $\D_t$ of $N'$ classes, where 
$\D_t = \{\dtrain^t,\dtest^t\}$. 
The learner learns from the training set
$\dtrain^t$ and evaluates on the test set $\dtest^t$.
The loss averaged across $\T_t$'s is taken as the meta-learning testing error. 

\begin{acks}
This research is partially done in 4Paradigm Inc. when Yaqing Wang took the internship.
\end{acks}

{
	\bibliographystyle{ACM-Reference-Format}
	
	\bibliography{fsl}}


\begin{thebibliography}{166}


\ifx \showCODEN    \undefined \def \showCODEN     #1{\unskip}     \fi
\ifx \showDOI      \undefined \def \showDOI       #1{#1}\fi
\ifx \showISBNx    \undefined \def \showISBNx     #1{\unskip}     \fi
\ifx \showISBNxiii \undefined \def \showISBNxiii  #1{\unskip}     \fi
\ifx \showISSN     \undefined \def \showISSN      #1{\unskip}     \fi
\ifx \showLCCN     \undefined \def \showLCCN      #1{\unskip}     \fi
\ifx \shownote     \undefined \def \shownote      #1{#1}          \fi
\ifx \showarticletitle \undefined \def \showarticletitle #1{#1}   \fi
\ifx \showURL      \undefined \def \showURL       {\relax}        \fi
\providecommand\bibfield[2]{#2}
\providecommand\bibinfo[2]{#2}
\providecommand\natexlab[1]{#1}
\providecommand\showeprint[2][]{arXiv:#2}

\bibitem[\protect\citeauthoryear{Abdo, Kretzschmar, Spinello, and
  Stachniss}{Abdo et~al\mbox{.}}{2013}]%
        {abdo2013learning}
\bibfield{author}{\bibinfo{person}{N. Abdo}, \bibinfo{person}{H. Kretzschmar},
  \bibinfo{person}{L. Spinello}, {and} \bibinfo{person}{C. Stachniss}.}
  \bibinfo{year}{2013}\natexlab{}.
\newblock \showarticletitle{Learning manipulation actions from a few
  demonstrations}. In \bibinfo{booktitle}{\emph{International Conference on
  Robotics and Automation}}. \bibinfo{pages}{1268--1275}.
\newblock


\bibitem[\protect\citeauthoryear{Akata, Perronnin, Harchaoui, and Schmid}{Akata
  et~al\mbox{.}}{2013}]%
        {akata2013label}
\bibfield{author}{\bibinfo{person}{Z. Akata}, \bibinfo{person}{F. Perronnin},
  \bibinfo{person}{Z. Harchaoui}, {and} \bibinfo{person}{C. Schmid}.}
  \bibinfo{year}{2013}\natexlab{}.
\newblock \showarticletitle{Label-embedding for attribute-based
  classification}. In \bibinfo{booktitle}{\emph{Conference on Computer Vision
  and Pattern Recognition}}. \bibinfo{pages}{819--826}.
\newblock


\bibitem[\protect\citeauthoryear{Al-Shedivat, Bansal, Burda, Sutskever,
  Mordatch, and Abbeel}{Al-Shedivat et~al\mbox{.}}{2018}]%
        {al2018continuous}
\bibfield{author}{\bibinfo{person}{M. Al-Shedivat}, \bibinfo{person}{T.
  Bansal}, \bibinfo{person}{Y. Burda}, \bibinfo{person}{I. Sutskever},
  \bibinfo{person}{I. Mordatch}, {and} \bibinfo{person}{P. Abbeel}.}
  \bibinfo{year}{2018}\natexlab{}.
\newblock \showarticletitle{Continuous adaptation via meta-learning in
  nonstationary and competitive environments}. In
  \bibinfo{booktitle}{\emph{International Conference on Learning
  Representations}}.
\newblock


\bibitem[\protect\citeauthoryear{Altae-Tran, Ramsundar, Pappu, and
  Pande}{Altae-Tran et~al\mbox{.}}{2017}]%
        {altae2017low}
\bibfield{author}{\bibinfo{person}{H. Altae-Tran}, \bibinfo{person}{B.
  Ramsundar}, \bibinfo{person}{A.~S. Pappu}, {and} \bibinfo{person}{V. Pande}.}
  \bibinfo{year}{2017}\natexlab{}.
\newblock \showarticletitle{Low data drug discovery with one-shot learning}.
\newblock \bibinfo{journal}{\emph{ACS Central Science}} \bibinfo{volume}{3},
  \bibinfo{number}{4} (\bibinfo{year}{2017}), \bibinfo{pages}{283--293}.
\newblock


\bibitem[\protect\citeauthoryear{Andrychowicz, Denil, Gomez, Hoffman, Pfau,
  Schaul, and de~Freitas}{Andrychowicz et~al\mbox{.}}{2016}]%
        {andrychowicz2016learning}
\bibfield{author}{\bibinfo{person}{M. Andrychowicz}, \bibinfo{person}{M.
  Denil}, \bibinfo{person}{S. Gomez}, \bibinfo{person}{M.~W. Hoffman},
  \bibinfo{person}{D. Pfau}, \bibinfo{person}{T. Schaul}, {and}
  \bibinfo{person}{N. de Freitas}.} \bibinfo{year}{2016}\natexlab{}.
\newblock \showarticletitle{Learning to learn by gradient descent by gradient
  descent}. In \bibinfo{booktitle}{\emph{Advances in Neural Information
  Processing Systems}}. \bibinfo{pages}{3981--3989}.
\newblock


\bibitem[\protect\citeauthoryear{Arik, Chen, Peng, Ping, and Zhou}{Arik
  et~al\mbox{.}}{2018}]%
        {arik2018neural}
\bibfield{author}{\bibinfo{person}{S. Arik}, \bibinfo{person}{J. Chen},
  \bibinfo{person}{K. Peng}, \bibinfo{person}{W. Ping}, {and}
  \bibinfo{person}{Y. Zhou}.} \bibinfo{year}{2018}\natexlab{}.
\newblock \showarticletitle{Neural voice cloning with a few samples}. In
  \bibinfo{booktitle}{\emph{Advances in Neural Information Processing
  Systems}}. \bibinfo{pages}{10019--10029}.
\newblock


\bibitem[\protect\citeauthoryear{Azadi, Fisher, Kim, Wang, Shechtman, and
  Darrell}{Azadi et~al\mbox{.}}{2018}]%
        {azadi2018multi}
\bibfield{author}{\bibinfo{person}{S. Azadi}, \bibinfo{person}{M. Fisher},
  \bibinfo{person}{V.~G. Kim}, \bibinfo{person}{Z. Wang}, \bibinfo{person}{E.
  Shechtman}, {and} \bibinfo{person}{T. Darrell}.}
  \bibinfo{year}{2018}\natexlab{}.
\newblock \showarticletitle{Multi-content GAN for few-shot font style
  transfer}. In \bibinfo{booktitle}{\emph{Conference on Computer Vision and
  Pattern Recognition}}. \bibinfo{pages}{7564--7573}.
\newblock


\bibitem[\protect\citeauthoryear{Bachman, Sordoni, and Trischler}{Bachman
  et~al\mbox{.}}{2017}]%
        {bachman17learning}
\bibfield{author}{\bibinfo{person}{P. Bachman}, \bibinfo{person}{A. Sordoni},
  {and} \bibinfo{person}{A. Trischler}.} \bibinfo{year}{2017}\natexlab{}.
\newblock \showarticletitle{Learning algorithms for active learning}. In
  \bibinfo{booktitle}{\emph{International Conference on Machine Learning}}.
  \bibinfo{pages}{301--310}.
\newblock


\bibitem[\protect\citeauthoryear{Bahdanau~D}{Bahdanau~D}{2015}]%
        {bahdanau2015neural}
\bibfield{author}{\bibinfo{person}{Bengio~Y. Bahdanau~D, Cho~K}.}
  \bibinfo{year}{2015}\natexlab{}.
\newblock \showarticletitle{Neural machine translation by jointly learning to
  align and translate}. In \bibinfo{booktitle}{\emph{International Conference
  on Learning Representations}}.
\newblock


\bibitem[\protect\citeauthoryear{Bart and Ullman}{Bart and Ullman}{2005}]%
        {bart2005cross}
\bibfield{author}{\bibinfo{person}{E. Bart} {and} \bibinfo{person}{S. Ullman}.}
  \bibinfo{year}{2005}\natexlab{}.
\newblock \showarticletitle{Cross-generalization: Learning novel classes from a
  single example by feature replacement}. In
  \bibinfo{booktitle}{\emph{Conference on Computer Vision and Pattern
  Recognition}}, Vol.~\bibinfo{volume}{1}. \bibinfo{pages}{672--679}.
\newblock


\bibitem[\protect\citeauthoryear{Ben-David, Blitzer, Crammer, and
  Pereira}{Ben-David et~al\mbox{.}}{2007}]%
        {ben2007analysis}
\bibfield{author}{\bibinfo{person}{S. Ben-David}, \bibinfo{person}{J. Blitzer},
  \bibinfo{person}{K. Crammer}, {and} \bibinfo{person}{F. Pereira}.}
  \bibinfo{year}{2007}\natexlab{}.
\newblock \showarticletitle{Analysis of representations for domain adaptation}.
  In \bibinfo{booktitle}{\emph{Advances in Neural Information Processing
  Systems}}. \bibinfo{pages}{137--144}.
\newblock


\bibitem[\protect\citeauthoryear{Benaim and Wolf}{Benaim and Wolf}{2018}]%
        {benaim2018one}
\bibfield{author}{\bibinfo{person}{S. Benaim} {and} \bibinfo{person}{L. Wolf}.}
  \bibinfo{year}{2018}\natexlab{}.
\newblock \showarticletitle{One-shot unsupervised cross domain translation}. In
  \bibinfo{booktitle}{\emph{Advances in Neural Information Processing
  Systems}}. \bibinfo{pages}{2104--2114}.
\newblock


\bibitem[\protect\citeauthoryear{Bertinetto, Henriques, Torr, and
  Vedaldi}{Bertinetto et~al\mbox{.}}{2019}]%
        {bertinetto2018metalearning}
\bibfield{author}{\bibinfo{person}{L. Bertinetto}, \bibinfo{person}{J.~F.
  Henriques}, \bibinfo{person}{P. Torr}, {and} \bibinfo{person}{A. Vedaldi}.}
  \bibinfo{year}{2019}\natexlab{}.
\newblock \showarticletitle{Meta-learning with differentiable closed-form
  solvers}. In \bibinfo{booktitle}{\emph{International Conference on Learning
  Representations}}.
\newblock


\bibitem[\protect\citeauthoryear{Bertinetto, Henriques, Valmadre, Torr, and
  Vedaldi}{Bertinetto et~al\mbox{.}}{2016}]%
        {bertinetto2016learning}
\bibfield{author}{\bibinfo{person}{L. Bertinetto}, \bibinfo{person}{J.~F.
  Henriques}, \bibinfo{person}{J. Valmadre}, \bibinfo{person}{P. Torr}, {and}
  \bibinfo{person}{A. Vedaldi}.} \bibinfo{year}{2016}\natexlab{}.
\newblock \showarticletitle{Learning feed-forward one-shot learners}. In
  \bibinfo{booktitle}{\emph{Advances in Neural Information Processing
  Systems}}. \bibinfo{pages}{523--531}.
\newblock


\bibitem[\protect\citeauthoryear{Bishop}{Bishop}{2006}]%
        {bishop2006pattern}
\bibfield{author}{\bibinfo{person}{C.~M. Bishop}.}
  \bibinfo{year}{2006}\natexlab{}.
\newblock \bibinfo{booktitle}{\emph{Pattern Recognition and Machine Learning}}.
\newblock \bibinfo{publisher}{Springer}.
\newblock


\bibitem[\protect\citeauthoryear{Blitzer, Crammer, Kulesza, Pereira, and
  Wortman}{Blitzer et~al\mbox{.}}{2008}]%
        {blitzer2008learning}
\bibfield{author}{\bibinfo{person}{J. Blitzer}, \bibinfo{person}{K. Crammer},
  \bibinfo{person}{A. Kulesza}, \bibinfo{person}{F. Pereira}, {and}
  \bibinfo{person}{J. Wortman}.} \bibinfo{year}{2008}\natexlab{}.
\newblock \showarticletitle{Learning bounds for domain adaptation}. In
  \bibinfo{booktitle}{\emph{Advances in Neural Information Processing
  Systems}}. \bibinfo{pages}{129--136}.
\newblock


\bibitem[\protect\citeauthoryear{Bottou and Bousquet}{Bottou and
  Bousquet}{2008}]%
        {bottou2008tradeoffs}
\bibfield{author}{\bibinfo{person}{L. Bottou} {and} \bibinfo{person}{O.
  Bousquet}.} \bibinfo{year}{2008}\natexlab{}.
\newblock \showarticletitle{The tradeoffs of large scale learning}. In
  \bibinfo{booktitle}{\emph{Advances in Neural Information Processing
  Systems}}. \bibinfo{pages}{161--168}.
\newblock


\bibitem[\protect\citeauthoryear{Bottou, Curtis, and Nocedal}{Bottou
  et~al\mbox{.}}{2018}]%
        {bottou2018optimization}
\bibfield{author}{\bibinfo{person}{L. Bottou}, \bibinfo{person}{F.~E. Curtis},
  {and} \bibinfo{person}{J. Nocedal}.} \bibinfo{year}{2018}\natexlab{}.
\newblock \showarticletitle{Optimization methods for large-scale machine
  learning}.
\newblock \bibinfo{journal}{\emph{SIAM Rev.}} \bibinfo{volume}{60},
  \bibinfo{number}{2} (\bibinfo{year}{2018}), \bibinfo{pages}{223--311}.
\newblock


\bibitem[\protect\citeauthoryear{Brock, Lim, Ritchie, and Weston}{Brock
  et~al\mbox{.}}{2018}]%
        {brock2018smash}
\bibfield{author}{\bibinfo{person}{A. Brock}, \bibinfo{person}{T. Lim},
  \bibinfo{person}{J.M. Ritchie}, {and} \bibinfo{person}{N. Weston}.}
  \bibinfo{year}{2018}\natexlab{}.
\newblock \showarticletitle{{SMASH}: One-shot model architecture search through
  hypernetworks}. In \bibinfo{booktitle}{\emph{International Conference on
  Learning Representations}}.
\newblock


\bibitem[\protect\citeauthoryear{Bromley, Guyon, LeCun, S{\"a}ckinger, and
  Shah}{Bromley et~al\mbox{.}}{1994}]%
        {bromley1994signature}
\bibfield{author}{\bibinfo{person}{J. Bromley}, \bibinfo{person}{I. Guyon},
  \bibinfo{person}{Y. LeCun}, \bibinfo{person}{E. S{\"a}ckinger}, {and}
  \bibinfo{person}{R. Shah}.} \bibinfo{year}{1994}\natexlab{}.
\newblock \showarticletitle{Signature verification using a "siamese" time delay
  neural network}. In \bibinfo{booktitle}{\emph{Advances in Neural Information
  Processing Systems}}. \bibinfo{pages}{737--744}.
\newblock


\bibitem[\protect\citeauthoryear{Caelles, Maninis, Pont-Tuset, Leal-Taix{\'e},
  Cremers, and Van~Gool}{Caelles et~al\mbox{.}}{2017}]%
        {caelles2017one}
\bibfield{author}{\bibinfo{person}{S. Caelles}, \bibinfo{person}{K.-K.
  Maninis}, \bibinfo{person}{J. Pont-Tuset}, \bibinfo{person}{L.
  Leal-Taix{\'e}}, \bibinfo{person}{D. Cremers}, {and} \bibinfo{person}{L.
  Van~Gool}.} \bibinfo{year}{2017}\natexlab{}.
\newblock \showarticletitle{One-shot video object segmentation}. In
  \bibinfo{booktitle}{\emph{Conference on Computer Vision and Pattern
  Recognition}}. \bibinfo{pages}{221--230}.
\newblock


\bibitem[\protect\citeauthoryear{Cai, Pan, Yao, Yan, and Mei}{Cai
  et~al\mbox{.}}{2018}]%
        {cai2018memory}
\bibfield{author}{\bibinfo{person}{Q. Cai}, \bibinfo{person}{Y. Pan},
  \bibinfo{person}{T. Yao}, \bibinfo{person}{C. Yan}, {and} \bibinfo{person}{T.
  Mei}.} \bibinfo{year}{2018}\natexlab{}.
\newblock \showarticletitle{Memory matching networks for one-shot image
  recognition}. In \bibinfo{booktitle}{\emph{Conference on Computer Vision and
  Pattern Recognition}}. \bibinfo{pages}{4080--4088}.
\newblock


\bibitem[\protect\citeauthoryear{Caruana}{Caruana}{1997}]%
        {caruana1997multitask}
\bibfield{author}{\bibinfo{person}{R. Caruana}.}
  \bibinfo{year}{1997}\natexlab{}.
\newblock \showarticletitle{Multitask learning}.
\newblock \bibinfo{journal}{\emph{Machine learning}} \bibinfo{volume}{28},
  \bibinfo{number}{1} (\bibinfo{year}{1997}), \bibinfo{pages}{41--75}.
\newblock


\bibitem[\protect\citeauthoryear{Choi, Krishnamurthy, Kembhavi, and
  Farhadi}{Choi et~al\mbox{.}}{2018}]%
        {choi2018structured}
\bibfield{author}{\bibinfo{person}{J. Choi}, \bibinfo{person}{J.
  Krishnamurthy}, \bibinfo{person}{A. Kembhavi}, {and} \bibinfo{person}{A.
  Farhadi}.} \bibinfo{year}{2018}\natexlab{}.
\newblock \showarticletitle{Structured set matching networks for one-shot part
  labeling}. In \bibinfo{booktitle}{\emph{Conference on Computer Vision and
  Pattern Recognition}}. \bibinfo{pages}{3627--3636}.
\newblock


\bibitem[\protect\citeauthoryear{Co-Reyes, Gupta, Sanjeev, Altieri, DeNero,
  Abbeel, and Levine}{Co-Reyes et~al\mbox{.}}{2019}]%
        {co-reyes2018metalearning}
\bibfield{author}{\bibinfo{person}{J.~D. Co-Reyes}, \bibinfo{person}{A. Gupta},
  \bibinfo{person}{S. Sanjeev}, \bibinfo{person}{N. Altieri},
  \bibinfo{person}{J. DeNero}, \bibinfo{person}{P. Abbeel}, {and}
  \bibinfo{person}{S. Levine}.} \bibinfo{year}{2019}\natexlab{}.
\newblock \showarticletitle{Meta-learning language-guided policy learning}. In
  \bibinfo{booktitle}{\emph{International Conference on Learning
  Representations}}.
\newblock


\bibitem[\protect\citeauthoryear{Craig}{Craig}{2009}]%
        {craig2009introduction}
\bibfield{author}{\bibinfo{person}{J.~J. Craig}.}
  \bibinfo{year}{2009}\natexlab{}.
\newblock \bibinfo{booktitle}{\emph{Introduction to Robotics: Mechanics and
  Control}}.
\newblock \bibinfo{publisher}{Pearson Education India}.
\newblock


\bibitem[\protect\citeauthoryear{Cubuk, Zoph, Mane, Vasudevan, and Le}{Cubuk
  et~al\mbox{.}}{2019}]%
        {cubuk2019autoaugment}
\bibfield{author}{\bibinfo{person}{E.~D. Cubuk}, \bibinfo{person}{B. Zoph},
  \bibinfo{person}{D. Mane}, \bibinfo{person}{V. Vasudevan}, {and}
  \bibinfo{person}{Q.~V. Le}.} \bibinfo{year}{2019}\natexlab{}.
\newblock \showarticletitle{AutoAugment: Learning augmentation policies from
  data}. In \bibinfo{booktitle}{\emph{Conference on Computer Vision and Pattern
  Recognition}}. \bibinfo{pages}{113--123}.
\newblock


\bibitem[\protect\citeauthoryear{Deleu and Bengio}{Deleu and Bengio}{2018}]%
        {deleu2018effects}
\bibfield{author}{\bibinfo{person}{T. Deleu} {and} \bibinfo{person}{Y.
  Bengio}.} \bibinfo{year}{2018}\natexlab{}.
\newblock \showarticletitle{The effects of negative adaptation in
  Model-Agnostic Meta-Learning}.
\newblock \bibinfo{journal}{\emph{arXiv preprint arXiv:1812.02159}}
  (\bibinfo{year}{2018}).
\newblock


\bibitem[\protect\citeauthoryear{Denevi, Ciliberto, Stamos, and Pontil}{Denevi
  et~al\mbox{.}}{2018}]%
        {denevi2018learning}
\bibfield{author}{\bibinfo{person}{G. Denevi}, \bibinfo{person}{C. Ciliberto},
  \bibinfo{person}{D. Stamos}, {and} \bibinfo{person}{M. Pontil}.}
  \bibinfo{year}{2018}\natexlab{}.
\newblock \showarticletitle{Learning to learn around a common mean}. In
  \bibinfo{booktitle}{\emph{Advances in Neural Information Processing
  Systems}}. \bibinfo{pages}{10190--10200}.
\newblock


\bibitem[\protect\citeauthoryear{Deng, Dong, Socher, Li, Li, and Fei-Fei}{Deng
  et~al\mbox{.}}{2009}]%
        {deng2009imagenet}
\bibfield{author}{\bibinfo{person}{J. Deng}, \bibinfo{person}{W. Dong},
  \bibinfo{person}{R. Socher}, \bibinfo{person}{L.-J. Li}, \bibinfo{person}{K.
  Li}, {and} \bibinfo{person}{L. Fei-Fei}.} \bibinfo{year}{2009}\natexlab{}.
\newblock \showarticletitle{ImageNet: A large-scale hierarchical image
  database}. In \bibinfo{booktitle}{\emph{Conference on Computer Vision and
  Pattern Recognition}}. \bibinfo{pages}{248--255}.
\newblock


\bibitem[\protect\citeauthoryear{Dong, Zhu, Zhang, Yang, and Wu}{Dong
  et~al\mbox{.}}{2018}]%
        {dong2018fast}
\bibfield{author}{\bibinfo{person}{X. Dong}, \bibinfo{person}{L. Zhu},
  \bibinfo{person}{D. Zhang}, \bibinfo{person}{Y. Yang}, {and}
  \bibinfo{person}{F. Wu}.} \bibinfo{year}{2018}\natexlab{}.
\newblock \showarticletitle{Fast parameter adaptation for few-shot image
  captioning and visual question answering}. In \bibinfo{booktitle}{\emph{ACM
  International Conference on Multimedia}}. \bibinfo{pages}{54--62}.
\newblock


\bibitem[\protect\citeauthoryear{Douze, Szlam, Hariharan, and J{\'e}gou}{Douze
  et~al\mbox{.}}{2018}]%
        {douze2018low}
\bibfield{author}{\bibinfo{person}{M. Douze}, \bibinfo{person}{A. Szlam},
  \bibinfo{person}{B. Hariharan}, {and} \bibinfo{person}{H. J{\'e}gou}.}
  \bibinfo{year}{2018}\natexlab{}.
\newblock \showarticletitle{Low-shot learning with large-scale diffusion}. In
  \bibinfo{booktitle}{\emph{Conference on Computer Vision and Pattern
  Recognition}}. \bibinfo{pages}{3349--3358}.
\newblock


\bibitem[\protect\citeauthoryear{Duan, Andrychowicz, Stadie, Ho, Schneider,
  Sutskever, Abbeel, and Zaremba}{Duan et~al\mbox{.}}{2017}]%
        {duan2017one}
\bibfield{author}{\bibinfo{person}{Y. Duan}, \bibinfo{person}{M. Andrychowicz},
  \bibinfo{person}{B. Stadie}, \bibinfo{person}{J. Ho}, \bibinfo{person}{J.
  Schneider}, \bibinfo{person}{I. Sutskever}, \bibinfo{person}{P. Abbeel},
  {and} \bibinfo{person}{W. Zaremba}.} \bibinfo{year}{2017}\natexlab{}.
\newblock \showarticletitle{One-shot imitation learning}. In
  \bibinfo{booktitle}{\emph{Advances in Neural Information Processing
  Systems}}. \bibinfo{pages}{1087--1098}.
\newblock


\bibitem[\protect\citeauthoryear{Edwards and Storkey}{Edwards and
  Storkey}{2017}]%
        {edwards2017towards}
\bibfield{author}{\bibinfo{person}{H. Edwards} {and} \bibinfo{person}{A.
  Storkey}.} \bibinfo{year}{2017}\natexlab{}.
\newblock \showarticletitle{Towards a neural statistician}. In
  \bibinfo{booktitle}{\emph{International Conference on Learning
  Representations}}.
\newblock


\bibitem[\protect\citeauthoryear{Fei-Fei, Fergus, and Perona}{Fei-Fei
  et~al\mbox{.}}{2006}]%
        {fei2006one}
\bibfield{author}{\bibinfo{person}{L. Fei-Fei}, \bibinfo{person}{R. Fergus},
  {and} \bibinfo{person}{P. Perona}.} \bibinfo{year}{2006}\natexlab{}.
\newblock \showarticletitle{One-shot learning of object categories}.
\newblock \bibinfo{journal}{\emph{IEEE Transactions on Pattern Analysis and
  Machine Intelligence}} \bibinfo{volume}{28}, \bibinfo{number}{4}
  (\bibinfo{year}{2006}), \bibinfo{pages}{594--611}.
\newblock


\bibitem[\protect\citeauthoryear{Fink}{Fink}{2005}]%
        {fink2005object}
\bibfield{author}{\bibinfo{person}{M. Fink}.} \bibinfo{year}{2005}\natexlab{}.
\newblock \showarticletitle{Object classification from a single example
  utilizing class relevance metrics}. In \bibinfo{booktitle}{\emph{Advances in
  Neural Information Processing Systems}}. \bibinfo{pages}{449--456}.
\newblock


\bibitem[\protect\citeauthoryear{Finn, Abbeel, and Levine}{Finn
  et~al\mbox{.}}{2017}]%
        {finn2017model}
\bibfield{author}{\bibinfo{person}{C. Finn}, \bibinfo{person}{P. Abbeel}, {and}
  \bibinfo{person}{S. Levine}.} \bibinfo{year}{2017}\natexlab{}.
\newblock \showarticletitle{Model-agnostic meta-learning for fast adaptation of
  deep networks}. In \bibinfo{booktitle}{\emph{International Conference on
  Machine Learning}}. \bibinfo{pages}{1126--1135}.
\newblock


\bibitem[\protect\citeauthoryear{Finn and Levine}{Finn and Levine}{2018}]%
        {finn2018meta}
\bibfield{author}{\bibinfo{person}{C. Finn} {and} \bibinfo{person}{S. Levine}.}
  \bibinfo{year}{2018}\natexlab{}.
\newblock \showarticletitle{Meta-learning and universality: Deep
  representations and gradient descent can approximate any learning algorithm}.
  In \bibinfo{booktitle}{\emph{International Conference on Learning
  Representations}}.
\newblock


\bibitem[\protect\citeauthoryear{Finn, Xu, and Levine}{Finn
  et~al\mbox{.}}{2018}]%
        {finn2018probabilistic}
\bibfield{author}{\bibinfo{person}{C. Finn}, \bibinfo{person}{K. Xu}, {and}
  \bibinfo{person}{S. Levine}.} \bibinfo{year}{2018}\natexlab{}.
\newblock \showarticletitle{Probabilistic model-agnostic meta-learning}. In
  \bibinfo{booktitle}{\emph{Advances in Neural Information Processing
  Systems}}. \bibinfo{pages}{9537--9548}.
\newblock


\bibitem[\protect\citeauthoryear{Franceschi, Frasconi, Salzo, Grazzi, and
  Pontil}{Franceschi et~al\mbox{.}}{2018}]%
        {franceschi2018bilevel}
\bibfield{author}{\bibinfo{person}{L. Franceschi}, \bibinfo{person}{P.
  Frasconi}, \bibinfo{person}{S. Salzo}, \bibinfo{person}{R. Grazzi}, {and}
  \bibinfo{person}{M. Pontil}.} \bibinfo{year}{2018}\natexlab{}.
\newblock \showarticletitle{Bilevel programming for hyperparameter optimization
  and meta-learning}. In \bibinfo{booktitle}{\emph{International Conference on
  Machine Learning}}. \bibinfo{pages}{1563--1572}.
\newblock


\bibitem[\protect\citeauthoryear{Friedman, Hastie, and Tibshirani}{Friedman
  et~al\mbox{.}}{2001}]%
        {friedman2001elements}
\bibfield{author}{\bibinfo{person}{J. Friedman}, \bibinfo{person}{T. Hastie},
  {and} \bibinfo{person}{R. Tibshirani}.} \bibinfo{year}{2001}\natexlab{}.
\newblock \bibinfo{booktitle}{\emph{The Elements of Statistical Learning}}.
  Vol.~\bibinfo{volume}{1}.
\newblock \bibinfo{publisher}{Springer series in statistics New York}.
\newblock


\bibitem[\protect\citeauthoryear{Gao, Shou, Zareian, Zhang, and Chang}{Gao
  et~al\mbox{.}}{2018}]%
        {gao2018low}
\bibfield{author}{\bibinfo{person}{H. Gao}, \bibinfo{person}{Z. Shou},
  \bibinfo{person}{A. Zareian}, \bibinfo{person}{H. Zhang}, {and}
  \bibinfo{person}{S. Chang}.} \bibinfo{year}{2018}\natexlab{}.
\newblock \showarticletitle{Low-shot learning via covariance-preserving
  adversarial augmentation networks}. In \bibinfo{booktitle}{\emph{Advances in
  Neural Information Processing Systems}}. \bibinfo{pages}{983--993}.
\newblock


\bibitem[\protect\citeauthoryear{Germain, Bach, Lacoste, and
  Lacoste-Julien}{Germain et~al\mbox{.}}{2016}]%
        {germain2016pac}
\bibfield{author}{\bibinfo{person}{P. Germain}, \bibinfo{person}{F. Bach},
  \bibinfo{person}{A. Lacoste}, {and} \bibinfo{person}{S. Lacoste-Julien}.}
  \bibinfo{year}{2016}\natexlab{}.
\newblock \showarticletitle{PAC-Bayesian theory meets Bayesian inference}. In
  \bibinfo{booktitle}{\emph{Advances in Neural Information Processing
  Systems}}. \bibinfo{pages}{1884--1892}.
\newblock


\bibitem[\protect\citeauthoryear{Gidaris and Komodakis}{Gidaris and
  Komodakis}{2018}]%
        {gidaris2018dynamic}
\bibfield{author}{\bibinfo{person}{S. Gidaris} {and} \bibinfo{person}{N.
  Komodakis}.} \bibinfo{year}{2018}\natexlab{}.
\newblock \showarticletitle{Dynamic few-shot visual learning without
  forgetting}. In \bibinfo{booktitle}{\emph{Conference on Computer Vision and
  Pattern Recognition}}. \bibinfo{pages}{4367--4375}.
\newblock


\bibitem[\protect\citeauthoryear{Goodfellow, Bengio, and Courville}{Goodfellow
  et~al\mbox{.}}{2016}]%
        {goodfellow2016deep}
\bibfield{author}{\bibinfo{person}{I. Goodfellow}, \bibinfo{person}{Y. Bengio},
  {and} \bibinfo{person}{A. Courville}.} \bibinfo{year}{2016}\natexlab{}.
\newblock \bibinfo{booktitle}{\emph{Deep Learning}}.
\newblock \bibinfo{publisher}{MIT Press}.
\newblock


\bibitem[\protect\citeauthoryear{Goodfellow, Pouget-Abadie, Mirza, Xu,
  Warde-Farley, Ozair, Courville, and Bengio}{Goodfellow et~al\mbox{.}}{2014}]%
        {goodfellow2014generative}
\bibfield{author}{\bibinfo{person}{I. Goodfellow}, \bibinfo{person}{J.
  Pouget-Abadie}, \bibinfo{person}{M. Mirza}, \bibinfo{person}{B. Xu},
  \bibinfo{person}{D. Warde-Farley}, \bibinfo{person}{S. Ozair},
  \bibinfo{person}{A. Courville}, {and} \bibinfo{person}{Y. Bengio}.}
  \bibinfo{year}{2014}\natexlab{}.
\newblock \showarticletitle{Generative adversarial nets}. In
  \bibinfo{booktitle}{\emph{Advances in Neural Information Processing
  Systems}}. \bibinfo{pages}{2672--2680}.
\newblock


\bibitem[\protect\citeauthoryear{Gordon, Bronskill, Bauer, Nowozin, and
  Turner}{Gordon et~al\mbox{.}}{2019}]%
        {gordon2018metalearning}
\bibfield{author}{\bibinfo{person}{J. Gordon}, \bibinfo{person}{J. Bronskill},
  \bibinfo{person}{M. Bauer}, \bibinfo{person}{S. Nowozin}, {and}
  \bibinfo{person}{R. Turner}.} \bibinfo{year}{2019}\natexlab{}.
\newblock \showarticletitle{Meta-learning probabilistic inference for
  prediction}. In \bibinfo{booktitle}{\emph{International Conference on
  Learning Representations}}.
\newblock


\bibitem[\protect\citeauthoryear{Grant, Finn, Levine, Darrell, and
  Griffiths}{Grant et~al\mbox{.}}{2018}]%
        {grant2018recasting}
\bibfield{author}{\bibinfo{person}{E. Grant}, \bibinfo{person}{C. Finn},
  \bibinfo{person}{S. Levine}, \bibinfo{person}{T. Darrell}, {and}
  \bibinfo{person}{T. Griffiths}.} \bibinfo{year}{2018}\natexlab{}.
\newblock \showarticletitle{Recasting gradient-based meta-learning as
  hierarchical Bayes}. In \bibinfo{booktitle}{\emph{International Conference on
  Learning Representations}}.
\newblock


\bibitem[\protect\citeauthoryear{Graves, Wayne, and Danihelka}{Graves
  et~al\mbox{.}}{2014}]%
        {graves2014neural}
\bibfield{author}{\bibinfo{person}{A. Graves}, \bibinfo{person}{G. Wayne},
  {and} \bibinfo{person}{I. Danihelka}.} \bibinfo{year}{2014}\natexlab{}.
\newblock \showarticletitle{Neural Turing machines}.
\newblock \bibinfo{journal}{\emph{arXiv preprint arXiv:1410.5401}}
  (\bibinfo{year}{2014}).
\newblock


\bibitem[\protect\citeauthoryear{Gui, Wang, Ramanan, and Moura}{Gui
  et~al\mbox{.}}{2018}]%
        {gui2018few}
\bibfield{author}{\bibinfo{person}{L.-Y. Gui}, \bibinfo{person}{Y.-X. Wang},
  \bibinfo{person}{D. Ramanan}, {and} \bibinfo{person}{J. Moura}.}
  \bibinfo{year}{2018}\natexlab{}.
\newblock \showarticletitle{Few-shot human motion prediction via
  meta-learning}. In \bibinfo{booktitle}{\emph{European Conference on Computer
  Vision}}. \bibinfo{pages}{432--450}.
\newblock


\bibitem[\protect\citeauthoryear{Hamaya, Matsubara, Noda, Teramae, and
  Morimoto}{Hamaya et~al\mbox{.}}{2016}]%
        {hamaya2016learning}
\bibfield{author}{\bibinfo{person}{M. Hamaya}, \bibinfo{person}{T. Matsubara},
  \bibinfo{person}{T. Noda}, \bibinfo{person}{T. Teramae}, {and}
  \bibinfo{person}{J. Morimoto}.} \bibinfo{year}{2016}\natexlab{}.
\newblock \showarticletitle{Learning assistive strategies from a few user-robot
  interactions: {M}odel-based reinforcement learning approach}. In
  \bibinfo{booktitle}{\emph{International Conference on Robotics and
  Automation}}. \bibinfo{pages}{3346--3351}.
\newblock


\bibitem[\protect\citeauthoryear{Han, Zhu, Yu, Wang, Yao, Liu, and Sun}{Han
  et~al\mbox{.}}{2018}]%
        {han2018fewrel}
\bibfield{author}{\bibinfo{person}{X. Han}, \bibinfo{person}{H. Zhu},
  \bibinfo{person}{P. Yu}, \bibinfo{person}{Z. Wang}, \bibinfo{person}{Y. Yao},
  \bibinfo{person}{Z. Liu}, {and} \bibinfo{person}{M. Sun}.}
  \bibinfo{year}{2018}\natexlab{}.
\newblock \showarticletitle{FewRel: A large-scale supervised few-shot relation
  classification dataset with state-of-the-art evaluation}. In
  \bibinfo{booktitle}{\emph{Conference on Empirical Methods in Natural Language
  Processing}}. \bibinfo{pages}{4803--4809}.
\newblock


\bibitem[\protect\citeauthoryear{Hariharan and Girshick}{Hariharan and
  Girshick}{2017}]%
        {hariharan2017low}
\bibfield{author}{\bibinfo{person}{B. Hariharan} {and} \bibinfo{person}{R.
  Girshick}.} \bibinfo{year}{2017}\natexlab{}.
\newblock \showarticletitle{Low-shot visual recognition by shrinking and
  hallucinating features}. In \bibinfo{booktitle}{\emph{International
  Conference on Computer Vision}}.
\newblock


\bibitem[\protect\citeauthoryear{He and Garcia}{He and Garcia}{2008}]%
        {he2008learning}
\bibfield{author}{\bibinfo{person}{H. He} {and} \bibinfo{person}{E.~A.
  Garcia}.} \bibinfo{year}{2008}\natexlab{}.
\newblock \showarticletitle{Learning from imbalanced data}.
\newblock \bibinfo{journal}{\emph{IEEE Transactions on Knowledge and Data
  Engineering}} \bibinfo{number}{9} (\bibinfo{year}{2008}),
  \bibinfo{pages}{1263--1284}.
\newblock


\bibitem[\protect\citeauthoryear{He, Zhang, Ren, and Sun}{He
  et~al\mbox{.}}{2016}]%
        {he2016deep}
\bibfield{author}{\bibinfo{person}{K. He}, \bibinfo{person}{X. Zhang},
  \bibinfo{person}{S. Ren}, {and} \bibinfo{person}{J. Sun}.}
  \bibinfo{year}{2016}\natexlab{}.
\newblock \showarticletitle{Deep residual learning for image recognition}. In
  \bibinfo{booktitle}{\emph{Conference on Computer Vision and Pattern
  Recognition}}. \bibinfo{pages}{770--778}.
\newblock


\bibitem[\protect\citeauthoryear{Herbelot and Baroni}{Herbelot and
  Baroni}{2017}]%
        {herbelot2017high}
\bibfield{author}{\bibinfo{person}{A. Herbelot} {and} \bibinfo{person}{M.
  Baroni}.} \bibinfo{year}{2017}\natexlab{}.
\newblock \showarticletitle{High-risk learning: Acquiring new word vectors from
  tiny data}. In \bibinfo{booktitle}{\emph{Conference on Empirical Methods in
  Natural Language Processing}}. \bibinfo{pages}{304--309}.
\newblock


\bibitem[\protect\citeauthoryear{Hewitt, Nye, Gane, Jaakkola, and
  Tenenbaum}{Hewitt et~al\mbox{.}}{2018}]%
        {hewitt2018variational}
\bibfield{author}{\bibinfo{person}{L.~B. Hewitt}, \bibinfo{person}{M.~I. Nye},
  \bibinfo{person}{A. Gane}, \bibinfo{person}{T. Jaakkola}, {and}
  \bibinfo{person}{J.~B. Tenenbaum}.} \bibinfo{year}{2018}\natexlab{}.
\newblock \showarticletitle{The variational homoencoder: Learning to learn high
  capacity generative models from few examples}. In
  \bibinfo{booktitle}{\emph{Uncertainty in Artificial Intelligence}}.
  \bibinfo{pages}{988--997}.
\newblock


\bibitem[\protect\citeauthoryear{Hochreiter and Schmidhuber}{Hochreiter and
  Schmidhuber}{1997}]%
        {hochreiter1997long}
\bibfield{author}{\bibinfo{person}{S. Hochreiter} {and} \bibinfo{person}{J.
  Schmidhuber}.} \bibinfo{year}{1997}\natexlab{}.
\newblock \showarticletitle{Long short-term memory}.
\newblock \bibinfo{journal}{\emph{Neural Computation}} \bibinfo{volume}{9},
  \bibinfo{number}{8} (\bibinfo{year}{1997}), \bibinfo{pages}{1735--1780}.
\newblock


\bibitem[\protect\citeauthoryear{Hochreiter, Younger, and Conwell}{Hochreiter
  et~al\mbox{.}}{2001}]%
        {hochreiter2001learning}
\bibfield{author}{\bibinfo{person}{S. Hochreiter}, \bibinfo{person}{A.~S.
  Younger}, {and} \bibinfo{person}{P.~R. Conwell}.}
  \bibinfo{year}{2001}\natexlab{}.
\newblock \showarticletitle{Learning to learn using gradient descent}. In
  \bibinfo{booktitle}{\emph{International Conference on Artificial Neural
  Networks}}. \bibinfo{pages}{87--94}.
\newblock


\bibitem[\protect\citeauthoryear{Hoffman, Tzeng, Donahue, Jia, Saenko, and
  Darrell}{Hoffman et~al\mbox{.}}{2013}]%
        {hoffman2013one}
\bibfield{author}{\bibinfo{person}{J. Hoffman}, \bibinfo{person}{E. Tzeng},
  \bibinfo{person}{J. Donahue}, \bibinfo{person}{Y. Jia}, \bibinfo{person}{K.
  Saenko}, {and} \bibinfo{person}{T. Darrell}.}
  \bibinfo{year}{2013}\natexlab{}.
\newblock \showarticletitle{One-shot adaptation of supervised deep
  convolutional models}. In \bibinfo{booktitle}{\emph{International Conference
  on Learning Representations}}.
\newblock


\bibitem[\protect\citeauthoryear{Hu, Li, Tu, Liu, and Sun}{Hu
  et~al\mbox{.}}{2018}]%
        {hu2018few}
\bibfield{author}{\bibinfo{person}{Z. Hu}, \bibinfo{person}{X. Li},
  \bibinfo{person}{C. Tu}, \bibinfo{person}{Z. Liu}, {and} \bibinfo{person}{M.
  Sun}.} \bibinfo{year}{2018}\natexlab{}.
\newblock \showarticletitle{Few-shot charge prediction with discriminative
  legal attributes}. In \bibinfo{booktitle}{\emph{International Conference on
  Computational Linguistics}}. \bibinfo{pages}{487--498}.
\newblock


\bibitem[\protect\citeauthoryear{Hwang and Sigal}{Hwang and Sigal}{2014}]%
        {hwang2014unified}
\bibfield{author}{\bibinfo{person}{S.~J. Hwang} {and} \bibinfo{person}{L.
  Sigal}.} \bibinfo{year}{2014}\natexlab{}.
\newblock \showarticletitle{A unified semantic embedding: Relating taxonomies
  and attributes}. In \bibinfo{booktitle}{\emph{Advances in Neural Information
  Processing Systems}}. \bibinfo{pages}{271--279}.
\newblock


\bibitem[\protect\citeauthoryear{Jia, Shelhamer, Donahue, Karayev, Long,
  Girshick, Guadarrama, and Darrell}{Jia et~al\mbox{.}}{2014}]%
        {jia2014caffe}
\bibfield{author}{\bibinfo{person}{Y. Jia}, \bibinfo{person}{E. Shelhamer},
  \bibinfo{person}{J. Donahue}, \bibinfo{person}{S. Karayev},
  \bibinfo{person}{J. Long}, \bibinfo{person}{R. Girshick}, \bibinfo{person}{S.
  Guadarrama}, {and} \bibinfo{person}{T. Darrell}.}
  \bibinfo{year}{2014}\natexlab{}.
\newblock \showarticletitle{Caffe: Convolutional architecture for fast feature
  embedding}. In \bibinfo{booktitle}{\emph{ACM International Conference on
  Multimedia}}. \bibinfo{pages}{675--678}.
\newblock


\bibitem[\protect\citeauthoryear{Joshi, Peters, and Hopkins}{Joshi
  et~al\mbox{.}}{2018}]%
        {joshi2018extending}
\bibfield{author}{\bibinfo{person}{V. Joshi}, \bibinfo{person}{M. Peters},
  {and} \bibinfo{person}{M. Hopkins}.} \bibinfo{year}{2018}\natexlab{}.
\newblock \showarticletitle{Extending a parser to distant domains using a few
  dozen partially annotated examples}. In \bibinfo{booktitle}{\emph{Annual
  Meeting of the Association for Computational Linguistics}}.
  \bibinfo{pages}{1190--1199}.
\newblock


\bibitem[\protect\citeauthoryear{Kaiser, Nachum, Roy, and Bengio}{Kaiser
  et~al\mbox{.}}{2017}]%
        {kaiser2017learning}
\bibfield{author}{\bibinfo{person}{{\L}. Kaiser}, \bibinfo{person}{O. Nachum},
  \bibinfo{person}{A. Roy}, {and} \bibinfo{person}{S. Bengio}.}
  \bibinfo{year}{2017}\natexlab{}.
\newblock \showarticletitle{Learning to remember rare events}. In
  \bibinfo{booktitle}{\emph{International Conference on Learning
  Representations}}.
\newblock


\bibitem[\protect\citeauthoryear{Kanter and Veeramachaneni}{Kanter and
  Veeramachaneni}{2015}]%
        {kanter2015deep}
\bibfield{author}{\bibinfo{person}{J.~M. Kanter} {and} \bibinfo{person}{K.
  Veeramachaneni}.} \bibinfo{year}{2015}\natexlab{}.
\newblock \showarticletitle{Deep feature synthesis: Towards automating data
  science endeavors}. In \bibinfo{booktitle}{\emph{International Conference on
  Data Science and Advanced Analytics}}. \bibinfo{pages}{1--10}.
\newblock


\bibitem[\protect\citeauthoryear{Keshari, Vatsa, Singh, and Noore}{Keshari
  et~al\mbox{.}}{2018}]%
        {keshari2018learning}
\bibfield{author}{\bibinfo{person}{R. Keshari}, \bibinfo{person}{M. Vatsa},
  \bibinfo{person}{R. Singh}, {and} \bibinfo{person}{A. Noore}.}
  \bibinfo{year}{2018}\natexlab{}.
\newblock \showarticletitle{Learning structure and strength of CNN filters for
  small sample size training}. In \bibinfo{booktitle}{\emph{Conference on
  Computer Vision and Pattern Recognition}}. \bibinfo{pages}{9349--9358}.
\newblock


\bibitem[\protect\citeauthoryear{Kingma and Welling}{Kingma and
  Welling}{2014}]%
        {kingma2014auto}
\bibfield{author}{\bibinfo{person}{D.~P. Kingma} {and} \bibinfo{person}{M.
  Welling}.} \bibinfo{year}{2014}\natexlab{}.
\newblock \showarticletitle{Auto-encoding variational Bayes}. In
  \bibinfo{booktitle}{\emph{International Conference on Learning
  Representations}}.
\newblock


\bibitem[\protect\citeauthoryear{Kirkpatrick, Pascanu, Rabinowitz, Veness,
  Desjardins, Rusu, Milan, Quan, Ramalho, Grabska-Barwinska,
  et~al\mbox{.}}{Kirkpatrick et~al\mbox{.}}{2017}]%
        {kirkpatrick2017overcoming}
\bibfield{author}{\bibinfo{person}{J. Kirkpatrick}, \bibinfo{person}{R.
  Pascanu}, \bibinfo{person}{N. Rabinowitz}, \bibinfo{person}{J. Veness},
  \bibinfo{person}{G. Desjardins}, \bibinfo{person}{A.~A. Rusu},
  \bibinfo{person}{K. Milan}, \bibinfo{person}{J. Quan}, \bibinfo{person}{T.
  Ramalho}, \bibinfo{person}{A. Grabska-Barwinska}, {et~al\mbox{.}}}
  \bibinfo{year}{2017}\natexlab{}.
\newblock \showarticletitle{Overcoming catastrophic forgetting in neural
  networks}.
\newblock \bibinfo{journal}{\emph{National Academy of Sciences}}
  \bibinfo{volume}{114}, \bibinfo{number}{13} (\bibinfo{year}{2017}),
  \bibinfo{pages}{3521--3526}.
\newblock


\bibitem[\protect\citeauthoryear{Koch}{Koch}{2015}]%
        {koch2015siamese}
\bibfield{author}{\bibinfo{person}{G. Koch}.} \bibinfo{year}{2015}\natexlab{}.
\newblock \emph{\bibinfo{title}{Siamese neural networks for one-shot image
  recognition}}.
\newblock \bibinfo{thesistype}{Ph.D. Dissertation}. \bibinfo{school}{University
  of Toronto}.
\newblock


\bibitem[\protect\citeauthoryear{Kotthoff, Thornton, Hoos, Hutter, and
  Leyton-Brown}{Kotthoff et~al\mbox{.}}{2017}]%
        {kotthoff2017auto}
\bibfield{author}{\bibinfo{person}{L. Kotthoff}, \bibinfo{person}{C. Thornton},
  \bibinfo{person}{H.~H. Hoos}, \bibinfo{person}{F. Hutter}, {and}
  \bibinfo{person}{K. Leyton-Brown}.} \bibinfo{year}{2017}\natexlab{}.
\newblock \showarticletitle{Auto-WEKA 2.0: Automatic model selection and
  hyperparameter optimization in WEKA}.
\newblock \bibinfo{journal}{\emph{Journal of Machine Learning Research}}
  \bibinfo{volume}{18}, \bibinfo{number}{1} (\bibinfo{year}{2017}),
  \bibinfo{pages}{826--830}.
\newblock


\bibitem[\protect\citeauthoryear{Kozerawski and Turk}{Kozerawski and
  Turk}{2018}]%
        {kozerawski2018clear}
\bibfield{author}{\bibinfo{person}{J. Kozerawski} {and} \bibinfo{person}{M.
  Turk}.} \bibinfo{year}{2018}\natexlab{}.
\newblock \showarticletitle{CLEAR: Cumulative learning for one-shot one-class
  image recognition}. In \bibinfo{booktitle}{\emph{Conference on Computer
  Vision and Pattern Recognition}}. \bibinfo{pages}{3446--3455}.
\newblock


\bibitem[\protect\citeauthoryear{Krizhevsky, Sutskever, and Hinton}{Krizhevsky
  et~al\mbox{.}}{2012}]%
        {krizhevsky2012imagenet}
\bibfield{author}{\bibinfo{person}{A. Krizhevsky}, \bibinfo{person}{I.
  Sutskever}, {and} \bibinfo{person}{G.~E. Hinton}.}
  \bibinfo{year}{2012}\natexlab{}.
\newblock \showarticletitle{ImageNet classification with deep convolutional
  neural networks}. In \bibinfo{booktitle}{\emph{Advances in Neural Information
  Processing Systems}}. \bibinfo{pages}{1097--1105}.
\newblock


\bibitem[\protect\citeauthoryear{Kwitt, Hegenbart, and Niethammer}{Kwitt
  et~al\mbox{.}}{2016}]%
        {kwitt2016one}
\bibfield{author}{\bibinfo{person}{R. Kwitt}, \bibinfo{person}{S. Hegenbart},
  {and} \bibinfo{person}{M. Niethammer}.} \bibinfo{year}{2016}\natexlab{}.
\newblock \showarticletitle{One-shot learning of scene locations via feature
  trajectory transfer}. In \bibinfo{booktitle}{\emph{Conference on Computer
  Vision and Pattern Recognition}}. \bibinfo{pages}{78--86}.
\newblock


\bibitem[\protect\citeauthoryear{Lake, Lee, Glass, and Tenenbaum}{Lake
  et~al\mbox{.}}{2014}]%
        {lake2014one}
\bibfield{author}{\bibinfo{person}{B. Lake}, \bibinfo{person}{C.-Y. Lee},
  \bibinfo{person}{J. Glass}, {and} \bibinfo{person}{J. Tenenbaum}.}
  \bibinfo{year}{2014}\natexlab{}.
\newblock \showarticletitle{One-shot learning of generative speech concepts}.
  In \bibinfo{booktitle}{\emph{Annual Meeting of the Cognitive Science
  Society}}, Vol.~\bibinfo{volume}{36}.
\newblock


\bibitem[\protect\citeauthoryear{Lake, Salakhutdinov, and Tenenbaum}{Lake
  et~al\mbox{.}}{2015}]%
        {lake2015human}
\bibfield{author}{\bibinfo{person}{B.~M. Lake}, \bibinfo{person}{R.
  Salakhutdinov}, {and} \bibinfo{person}{J.~B. Tenenbaum}.}
  \bibinfo{year}{2015}\natexlab{}.
\newblock \showarticletitle{Human-level concept learning through probabilistic
  program induction}.
\newblock \bibinfo{journal}{\emph{Science}} \bibinfo{volume}{350},
  \bibinfo{number}{6266} (\bibinfo{year}{2015}), \bibinfo{pages}{1332--1338}.
\newblock


\bibitem[\protect\citeauthoryear{Lake, Ullman, Tenenbaum, and Gershman}{Lake
  et~al\mbox{.}}{2017}]%
        {lake2017building}
\bibfield{author}{\bibinfo{person}{B.~M. Lake}, \bibinfo{person}{T.~D. Ullman},
  \bibinfo{person}{J.~B. Tenenbaum}, {and} \bibinfo{person}{S.~J. Gershman}.}
  \bibinfo{year}{2017}\natexlab{}.
\newblock \showarticletitle{Building machines that learn and think like
  people}.
\newblock \bibinfo{journal}{\emph{Behavioral and Brain Sciences}}
  \bibinfo{volume}{40} (\bibinfo{year}{2017}).
\newblock


\bibitem[\protect\citeauthoryear{Lampert, Nickisch, and Harmeling}{Lampert
  et~al\mbox{.}}{2009}]%
        {lampert2009learning}
\bibfield{author}{\bibinfo{person}{C.~H. Lampert}, \bibinfo{person}{H.
  Nickisch}, {and} \bibinfo{person}{S. Harmeling}.}
  \bibinfo{year}{2009}\natexlab{}.
\newblock \showarticletitle{Learning to detect unseen object classes by
  between-class attribute transfer}. In \bibinfo{booktitle}{\emph{Conference on
  Computer Vision and Pattern Recognition}}. \bibinfo{pages}{951--958}.
\newblock


\bibitem[\protect\citeauthoryear{Lee and Choi}{Lee and Choi}{2018}]%
        {lee2018gradient}
\bibfield{author}{\bibinfo{person}{Y. Lee} {and} \bibinfo{person}{S. Choi}.}
  \bibinfo{year}{2018}\natexlab{}.
\newblock \showarticletitle{Gradient-based meta-learning with learned layerwise
  metric and subspace}. In \bibinfo{booktitle}{\emph{International Conference
  on Machine Learning}}. \bibinfo{pages}{2933--2942}.
\newblock


\bibitem[\protect\citeauthoryear{Li and Malik}{Li and Malik}{2017}]%
        {li2017learning}
\bibfield{author}{\bibinfo{person}{K. Li} {and} \bibinfo{person}{J. Malik}.}
  \bibinfo{year}{2017}\natexlab{}.
\newblock \showarticletitle{Learning to optimize}. In
  \bibinfo{booktitle}{\emph{International Conference on Learning
  Representations}}.
\newblock


\bibitem[\protect\citeauthoryear{Li, Yu, Liu, and Ng}{Li et~al\mbox{.}}{2009}]%
        {li2009positive}
\bibfield{author}{\bibinfo{person}{X.-L. Li}, \bibinfo{person}{P.~S. Yu},
  \bibinfo{person}{B. Liu}, {and} \bibinfo{person}{S.-K. Ng}.}
  \bibinfo{year}{2009}\natexlab{}.
\newblock \showarticletitle{Positive unlabeled learning for data stream
  classification}. In \bibinfo{booktitle}{\emph{SIAM International Conference
  on Data Mining}}. \bibinfo{pages}{259--270}.
\newblock


\bibitem[\protect\citeauthoryear{Liu, Wang, Dixit, Kwitt, and Vasconcelos}{Liu
  et~al\mbox{.}}{2018}]%
        {liu2018feature}
\bibfield{author}{\bibinfo{person}{B. Liu}, \bibinfo{person}{X. Wang},
  \bibinfo{person}{M. Dixit}, \bibinfo{person}{R. Kwitt}, {and}
  \bibinfo{person}{N. Vasconcelos}.} \bibinfo{year}{2018}\natexlab{}.
\newblock \showarticletitle{Feature space transfer for data augmentation}. In
  \bibinfo{booktitle}{\emph{Conference on Computer Vision and Pattern
  Recognition}}. \bibinfo{pages}{9090--9098}.
\newblock


\bibitem[\protect\citeauthoryear{Liu, Simonyan, and Yang}{Liu
  et~al\mbox{.}}{2019b}]%
        {liu2018darts}
\bibfield{author}{\bibinfo{person}{H. Liu}, \bibinfo{person}{K. Simonyan},
  {and} \bibinfo{person}{Y. Yang}.} \bibinfo{year}{2019}\natexlab{b}.
\newblock \showarticletitle{{DARTS}: Differentiable architecture search}. In
  \bibinfo{booktitle}{\emph{International Conference on Learning
  Representations}}.
\newblock


\bibitem[\protect\citeauthoryear{Liu, Lee, Park, Kim, Yang, Hwang, and
  Yang}{Liu et~al\mbox{.}}{2019a}]%
        {liu2019learning}
\bibfield{author}{\bibinfo{person}{Y. Liu}, \bibinfo{person}{J. Lee},
  \bibinfo{person}{M. Park}, \bibinfo{person}{S. Kim}, \bibinfo{person}{E.
  Yang}, \bibinfo{person}{S. Hwang}, {and} \bibinfo{person}{Y Yang}.}
  \bibinfo{year}{2019}\natexlab{a}.
\newblock \showarticletitle{Learning to propopagate labels: Transductive
  propagation network for few-shot learning}. In
  \bibinfo{booktitle}{\emph{International Conference on Learning
  Representations}}.
\newblock


\bibitem[\protect\citeauthoryear{Luo, Zou, Hoffman, and Fei-Fei}{Luo
  et~al\mbox{.}}{2017}]%
        {luo2017label}
\bibfield{author}{\bibinfo{person}{Z. Luo}, \bibinfo{person}{Y. Zou},
  \bibinfo{person}{J. Hoffman}, {and} \bibinfo{person}{L. Fei-Fei}.}
  \bibinfo{year}{2017}\natexlab{}.
\newblock \showarticletitle{Label efficient learning of transferable
  representations acrosss domains and tasks}. In
  \bibinfo{booktitle}{\emph{Advances in Neural Information Processing
  Systems}}. \bibinfo{pages}{165--177}.
\newblock


\bibitem[\protect\citeauthoryear{Mahadevan and Tadepalli}{Mahadevan and
  Tadepalli}{1994}]%
        {mahadevan1994quantifying}
\bibfield{author}{\bibinfo{person}{S. Mahadevan} {and} \bibinfo{person}{P.
  Tadepalli}.} \bibinfo{year}{1994}\natexlab{}.
\newblock \showarticletitle{Quantifying prior determination knowledge using the
  PAC learning model}.
\newblock \bibinfo{journal}{\emph{Machine Learning}} \bibinfo{volume}{17},
  \bibinfo{number}{1} (\bibinfo{year}{1994}), \bibinfo{pages}{69--105}.
\newblock


\bibitem[\protect\citeauthoryear{McNamara and Balcan}{McNamara and
  Balcan}{2017}]%
        {mcnamara2017risk}
\bibfield{author}{\bibinfo{person}{D. McNamara} {and} \bibinfo{person}{M.-F.
  Balcan}.} \bibinfo{year}{2017}\natexlab{}.
\newblock \showarticletitle{Risk bounds for transferring representations with
  and without fine-tuning}. In \bibinfo{booktitle}{\emph{International
  Conference on Machine Learning}}. \bibinfo{pages}{2373--2381}.
\newblock


\bibitem[\protect\citeauthoryear{Mensink, Gavves, and Snoek}{Mensink
  et~al\mbox{.}}{2014}]%
        {mensink2014costa}
\bibfield{author}{\bibinfo{person}{T. Mensink}, \bibinfo{person}{E. Gavves},
  {and} \bibinfo{person}{C. Snoek}.} \bibinfo{year}{2014}\natexlab{}.
\newblock \showarticletitle{Costa: Co-occurrence statistics for zero-shot
  classification}. In \bibinfo{booktitle}{\emph{Conference on Computer Vision
  and Pattern Recognition}}. \bibinfo{pages}{2441--2448}.
\newblock


\bibitem[\protect\citeauthoryear{Miller, Fisch, Dodge, Karimi, Bordes, and
  Weston}{Miller et~al\mbox{.}}{2016}]%
        {miller2016key}
\bibfield{author}{\bibinfo{person}{A. Miller}, \bibinfo{person}{A. Fisch},
  \bibinfo{person}{J. Dodge}, \bibinfo{person}{A.-H. Karimi},
  \bibinfo{person}{A. Bordes}, {and} \bibinfo{person}{J. Weston}.}
  \bibinfo{year}{2016}\natexlab{}.
\newblock \showarticletitle{Key-value memory networks for directly reading
  documents}. In \bibinfo{booktitle}{\emph{Conference on Empirical Methods in
  Natural Language Processing}}. \bibinfo{pages}{1400--1409}.
\newblock


\bibitem[\protect\citeauthoryear{Miller, Matsakis, and Viola}{Miller
  et~al\mbox{.}}{2000}]%
        {miller2000learning}
\bibfield{author}{\bibinfo{person}{E.~G. Miller}, \bibinfo{person}{N.~E.
  Matsakis}, {and} \bibinfo{person}{P.~A. Viola}.}
  \bibinfo{year}{2000}\natexlab{}.
\newblock \showarticletitle{Learning from one example through shared densities
  on transforms}. In \bibinfo{booktitle}{\emph{Conference on Computer Vision
  and Pattern Recognition}}, Vol.~\bibinfo{volume}{1}.
  \bibinfo{pages}{464--471}.
\newblock


\bibitem[\protect\citeauthoryear{Mishra, Rohaninejad, Chen, and Abbeel}{Mishra
  et~al\mbox{.}}{2018}]%
        {mishra2018a}
\bibfield{author}{\bibinfo{person}{N. Mishra}, \bibinfo{person}{M.
  Rohaninejad}, \bibinfo{person}{X. Chen}, {and} \bibinfo{person}{P. Abbeel}.}
  \bibinfo{year}{2018}\natexlab{}.
\newblock \showarticletitle{A simple neural attentive meta-learner}. In
  \bibinfo{booktitle}{\emph{International Conference on Learning
  Representations}}.
\newblock


\bibitem[\protect\citeauthoryear{Mitchell}{Mitchell}{1997}]%
        {tom1997machine}
\bibfield{author}{\bibinfo{person}{M.~T. Mitchell}.}
  \bibinfo{year}{1997}\natexlab{}.
\newblock \bibinfo{booktitle}{\emph{Machine Learning}}.
\newblock \bibinfo{publisher}{McGraw-Hill}.
\newblock


\bibitem[\protect\citeauthoryear{Mohammadi and Kim}{Mohammadi and Kim}{2018}]%
        {mohammadi2018investigation}
\bibfield{author}{\bibinfo{person}{S.~H. Mohammadi} {and} \bibinfo{person}{T.
  Kim}.} \bibinfo{year}{2018}\natexlab{}.
\newblock \showarticletitle{Investigation of using disentangled and
  interpretable representations for one-shot cross-lingual voice conversion}.
  In \bibinfo{booktitle}{\emph{INTERSPEECH}}. \bibinfo{pages}{2833--2837}.
\newblock


\bibitem[\protect\citeauthoryear{Mohri, Rostamizadeh, and Talwalkar}{Mohri
  et~al\mbox{.}}{2018}]%
        {mohri2018foundations}
\bibfield{author}{\bibinfo{person}{M. Mohri}, \bibinfo{person}{A.
  Rostamizadeh}, {and} \bibinfo{person}{A. Talwalkar}.}
  \bibinfo{year}{2018}\natexlab{}.
\newblock \bibinfo{booktitle}{\emph{Foundations of machine learning}}.
\newblock \bibinfo{publisher}{MIT Press}.
\newblock


\bibitem[\protect\citeauthoryear{Motiian, Jones, Iranmanesh, and
  Doretto}{Motiian et~al\mbox{.}}{2017}]%
        {motiian2017few}
\bibfield{author}{\bibinfo{person}{S. Motiian}, \bibinfo{person}{Q. Jones},
  \bibinfo{person}{S. Iranmanesh}, {and} \bibinfo{person}{G. Doretto}.}
  \bibinfo{year}{2017}\natexlab{}.
\newblock \showarticletitle{Few-shot adversarial domain adaptation}. In
  \bibinfo{booktitle}{\emph{Advances in Neural Information Processing
  Systems}}. \bibinfo{pages}{6670--6680}.
\newblock


\bibitem[\protect\citeauthoryear{Munkhdalai and Yu}{Munkhdalai and Yu}{2017}]%
        {munkhdalai2017meta}
\bibfield{author}{\bibinfo{person}{T. Munkhdalai} {and} \bibinfo{person}{H.
  Yu}.} \bibinfo{year}{2017}\natexlab{}.
\newblock \showarticletitle{Meta networks}. In
  \bibinfo{booktitle}{\emph{International Conference on Machine Learning}}.
  \bibinfo{pages}{2554--2563}.
\newblock


\bibitem[\protect\citeauthoryear{Munkhdalai, Yuan, Mehri, and
  Trischler}{Munkhdalai et~al\mbox{.}}{2018}]%
        {munkhdalai2018rapid}
\bibfield{author}{\bibinfo{person}{T. Munkhdalai}, \bibinfo{person}{X. Yuan},
  \bibinfo{person}{S. Mehri}, {and} \bibinfo{person}{A. Trischler}.}
  \bibinfo{year}{2018}\natexlab{}.
\newblock \showarticletitle{Rapid adaptation with conditionally shifted
  neurons}. In \bibinfo{booktitle}{\emph{International Conference on Machine
  Learning}}. \bibinfo{pages}{3661--3670}.
\newblock


\bibitem[\protect\citeauthoryear{Nagabandi, Finn, and Levine}{Nagabandi
  et~al\mbox{.}}{2018}]%
        {nagabandi2018deep}
\bibfield{author}{\bibinfo{person}{A. Nagabandi}, \bibinfo{person}{C. Finn},
  {and} \bibinfo{person}{S. Levine}.} \bibinfo{year}{2018}\natexlab{}.
\newblock \showarticletitle{Deep online learning via meta-learning: Continual
  adaptation for model-based RL}. In \bibinfo{booktitle}{\emph{International
  Conference on Learning Representations}}.
\newblock


\bibitem[\protect\citeauthoryear{Nguyen and Zakynthinou}{Nguyen and
  Zakynthinou}{2018}]%
        {nguyen2018improved}
\bibfield{author}{\bibinfo{person}{H. Nguyen} {and} \bibinfo{person}{L.
  Zakynthinou}.} \bibinfo{year}{2018}\natexlab{}.
\newblock \showarticletitle{Improved algorithms for collaborative PAC
  learning}. In \bibinfo{booktitle}{\emph{Advances in Neural Information
  Processing Systems}}. \bibinfo{pages}{7631--7639}.
\newblock


\bibitem[\protect\citeauthoryear{Oreshkin, L{\'o}pez, and Lacoste}{Oreshkin
  et~al\mbox{.}}{2018}]%
        {oreshkin2018tadam}
\bibfield{author}{\bibinfo{person}{B. Oreshkin}, \bibinfo{person}{P.~R.
  L{\'o}pez}, {and} \bibinfo{person}{A. Lacoste}.}
  \bibinfo{year}{2018}\natexlab{}.
\newblock \showarticletitle{TADAM: Task dependent adaptive metric for improved
  few-shot learning}. In \bibinfo{booktitle}{\emph{Advances in Neural
  Information Processing Systems}}. \bibinfo{pages}{719--729}.
\newblock


\bibitem[\protect\citeauthoryear{Pan and Yang}{Pan and Yang}{2010}]%
        {pan2010survey}
\bibfield{author}{\bibinfo{person}{S.~J. Pan} {and} \bibinfo{person}{Q. Yang}.}
  \bibinfo{year}{2010}\natexlab{}.
\newblock \showarticletitle{A survey on transfer learning}.
\newblock \bibinfo{journal}{\emph{IEEE Transactions on Knowledge and Data
  Engineering}} \bibinfo{volume}{10}, \bibinfo{number}{22}
  (\bibinfo{year}{2010}), \bibinfo{pages}{1345--1359}.
\newblock


\bibitem[\protect\citeauthoryear{Pfister, Charles, and Zisserman}{Pfister
  et~al\mbox{.}}{2014}]%
        {pfister2014domain}
\bibfield{author}{\bibinfo{person}{T. Pfister}, \bibinfo{person}{J. Charles},
  {and} \bibinfo{person}{A. Zisserman}.} \bibinfo{year}{2014}\natexlab{}.
\newblock \showarticletitle{Domain-adaptive discriminative one-shot learning of
  gestures}. In \bibinfo{booktitle}{\emph{European Conference on Computer
  Vision}}. \bibinfo{pages}{814--829}.
\newblock


\bibitem[\protect\citeauthoryear{Qi, Brown, and Lowe}{Qi et~al\mbox{.}}{2018}]%
        {qi2018low}
\bibfield{author}{\bibinfo{person}{H. Qi}, \bibinfo{person}{M. Brown}, {and}
  \bibinfo{person}{D.~G. Lowe}.} \bibinfo{year}{2018}\natexlab{}.
\newblock \showarticletitle{Low-shot learning with imprinted weights}. In
  \bibinfo{booktitle}{\emph{Conference on Computer Vision and Pattern
  Recognition}}. \bibinfo{pages}{5822--5830}.
\newblock


\bibitem[\protect\citeauthoryear{Ramalho and Garnelo}{Ramalho and
  Garnelo}{2019}]%
        {ramalho2018adaptive}
\bibfield{author}{\bibinfo{person}{T. Ramalho} {and} \bibinfo{person}{M.
  Garnelo}.} \bibinfo{year}{2019}\natexlab{}.
\newblock \showarticletitle{Adaptive posterior learning: Few-shot learning with
  a surprise-based memory module}. In \bibinfo{booktitle}{\emph{International
  Conference on Learning Representations}}.
\newblock


\bibitem[\protect\citeauthoryear{Ravi and Beatson}{Ravi and Beatson}{2019}]%
        {ravi2018amortized}
\bibfield{author}{\bibinfo{person}{S. Ravi} {and} \bibinfo{person}{A.
  Beatson}.} \bibinfo{year}{2019}\natexlab{}.
\newblock \showarticletitle{Amortized Bayesian meta-learning}. In
  \bibinfo{booktitle}{\emph{International Conference on Learning
  Representations}}.
\newblock


\bibitem[\protect\citeauthoryear{Ravi and Larochelle}{Ravi and
  Larochelle}{2017}]%
        {ravi2017optimization}
\bibfield{author}{\bibinfo{person}{S. Ravi} {and} \bibinfo{person}{H.
  Larochelle}.} \bibinfo{year}{2017}\natexlab{}.
\newblock \showarticletitle{Optimization as a model for few-shot learning}. In
  \bibinfo{booktitle}{\emph{International Conference on Learning
  Representations}}.
\newblock


\bibitem[\protect\citeauthoryear{Reed, Chen, Paine, van~den Oord, Eslami,
  Rezende, Vinyals, and de~Freitas}{Reed et~al\mbox{.}}{2018}]%
        {reed2018fewshot}
\bibfield{author}{\bibinfo{person}{S. Reed}, \bibinfo{person}{Y. Chen},
  \bibinfo{person}{T. Paine}, \bibinfo{person}{A. van~den Oord},
  \bibinfo{person}{S.~M.~A. Eslami}, \bibinfo{person}{D. Rezende},
  \bibinfo{person}{O. Vinyals}, {and} \bibinfo{person}{N. de Freitas}.}
  \bibinfo{year}{2018}\natexlab{}.
\newblock \showarticletitle{Few-shot autoregressive density estimation: Towards
  learning to learn distributions}. In \bibinfo{booktitle}{\emph{International
  Conference on Learning Representations}}.
\newblock


\bibitem[\protect\citeauthoryear{Ren, Ravi, Triantafillou, Snell, Swersky,
  Tenenbaum, Larochelle, and Zemel}{Ren et~al\mbox{.}}{2018}]%
        {ren2018metalearning}
\bibfield{author}{\bibinfo{person}{M. Ren}, \bibinfo{person}{S. Ravi},
  \bibinfo{person}{E. Triantafillou}, \bibinfo{person}{J. Snell},
  \bibinfo{person}{K. Swersky}, \bibinfo{person}{J.~B. Tenenbaum},
  \bibinfo{person}{H. Larochelle}, {and} \bibinfo{person}{R.~S. Zemel}.}
  \bibinfo{year}{2018}\natexlab{}.
\newblock \showarticletitle{Meta-learning for semi-supervised few-shot
  classification}. In \bibinfo{booktitle}{\emph{International Conference on
  Learning Representations}}.
\newblock


\bibitem[\protect\citeauthoryear{Rezende, Danihelka, Gregor, and
  Wierstra}{Rezende et~al\mbox{.}}{2016}]%
        {rezende2016one}
\bibfield{author}{\bibinfo{person}{D. Rezende}, \bibinfo{person}{I. Danihelka},
  \bibinfo{person}{K. Gregor}, {and} \bibinfo{person}{D. Wierstra}.}
  \bibinfo{year}{2016}\natexlab{}.
\newblock \showarticletitle{One-shot generalization in deep generative models}.
  In \bibinfo{booktitle}{\emph{International Conference on Machine Learning}}.
  \bibinfo{pages}{1521--1529}.
\newblock


\bibitem[\protect\citeauthoryear{Rios and Kavuluru}{Rios and Kavuluru}{2018}]%
        {rios2018few}
\bibfield{author}{\bibinfo{person}{A. Rios} {and} \bibinfo{person}{R.
  Kavuluru}.} \bibinfo{year}{2018}\natexlab{}.
\newblock \showarticletitle{Few-shot and zero-shot multi-label learning for
  structured label spaces}. In \bibinfo{booktitle}{\emph{Conference on
  Empirical Methods in Natural Language Processing}}. \bibinfo{pages}{3132}.
\newblock


\bibitem[\protect\citeauthoryear{Rusu, Rao, Sygnowski, Vinyals, Pascanu,
  Osindero, and Hadsell}{Rusu et~al\mbox{.}}{2019}]%
        {rusu2018metalearning}
\bibfield{author}{\bibinfo{person}{A.~A. Rusu}, \bibinfo{person}{D. Rao},
  \bibinfo{person}{J. Sygnowski}, \bibinfo{person}{O. Vinyals},
  \bibinfo{person}{R. Pascanu}, \bibinfo{person}{S. Osindero}, {and}
  \bibinfo{person}{R. Hadsell}.} \bibinfo{year}{2019}\natexlab{}.
\newblock \showarticletitle{Meta-learning with latent embedding optimization}.
  In \bibinfo{booktitle}{\emph{International Conference on Learning
  Representations}}.
\newblock


\bibitem[\protect\citeauthoryear{Salakhutdinov and Hinton}{Salakhutdinov and
  Hinton}{2009}]%
        {salakhutdinov2009deep}
\bibfield{author}{\bibinfo{person}{R. Salakhutdinov} {and} \bibinfo{person}{G.
  Hinton}.} \bibinfo{year}{2009}\natexlab{}.
\newblock \showarticletitle{Deep boltzmann machines}. In
  \bibinfo{booktitle}{\emph{International Conference on Artificial Intelligence
  and Statistics}}. \bibinfo{pages}{448--455}.
\newblock


\bibitem[\protect\citeauthoryear{Salakhutdinov, Tenenbaum, and
  Torralba}{Salakhutdinov et~al\mbox{.}}{2012}]%
        {salakhutdinov2012one}
\bibfield{author}{\bibinfo{person}{R. Salakhutdinov}, \bibinfo{person}{J.
  Tenenbaum}, {and} \bibinfo{person}{A. Torralba}.}
  \bibinfo{year}{2012}\natexlab{}.
\newblock \showarticletitle{One-shot learning with a hierarchical nonparametric
  Bayesian model}. In \bibinfo{booktitle}{\emph{ICML Workshop on Unsupervised
  and Transfer Learning}}. \bibinfo{pages}{195--206}.
\newblock


\bibitem[\protect\citeauthoryear{Santoro, Bartunov, Botvinick, Wierstra, and
  Lillicrap}{Santoro et~al\mbox{.}}{2016}]%
        {santoro2016meta}
\bibfield{author}{\bibinfo{person}{A. Santoro}, \bibinfo{person}{S. Bartunov},
  \bibinfo{person}{M. Botvinick}, \bibinfo{person}{D. Wierstra}, {and}
  \bibinfo{person}{T. Lillicrap}.} \bibinfo{year}{2016}\natexlab{}.
\newblock \showarticletitle{Meta-learning with memory-augmented neural
  networks}. In \bibinfo{booktitle}{\emph{International Conference on Machine
  Learning}}. \bibinfo{pages}{1842--1850}.
\newblock


\bibitem[\protect\citeauthoryear{Satorras and Estrach}{Satorras and
  Estrach}{2018}]%
        {garcia2018fewshot}
\bibfield{author}{\bibinfo{person}{V.~G. Satorras} {and} \bibinfo{person}{J.~B.
  Estrach}.} \bibinfo{year}{2018}\natexlab{}.
\newblock \showarticletitle{Few-shot learning with graph neural networks}. In
  \bibinfo{booktitle}{\emph{International Conference on Learning
  Representations}}.
\newblock


\bibitem[\protect\citeauthoryear{Schwartz, Karlinsky, Shtok, Harary, Marder,
  Kumar, Feris, Giryes, and Bronstein}{Schwartz et~al\mbox{.}}{2018}]%
        {eli2018delta}
\bibfield{author}{\bibinfo{person}{E. Schwartz}, \bibinfo{person}{L.
  Karlinsky}, \bibinfo{person}{J. Shtok}, \bibinfo{person}{S. Harary},
  \bibinfo{person}{M. Marder}, \bibinfo{person}{A. Kumar}, \bibinfo{person}{R.
  Feris}, \bibinfo{person}{R. Giryes}, {and} \bibinfo{person}{A. Bronstein}.}
  \bibinfo{year}{2018}\natexlab{}.
\newblock \showarticletitle{Delta-encoder: An effective sample synthesis method
  for few-shot object recognition}. In \bibinfo{booktitle}{\emph{Advances in
  Neural Information Processing Systems}}. \bibinfo{pages}{2850--2860}.
\newblock


\bibitem[\protect\citeauthoryear{Settles}{Settles}{2009}]%
        {settles2009active}
\bibfield{author}{\bibinfo{person}{B. Settles}.}
  \bibinfo{year}{2009}\natexlab{}.
\newblock \bibinfo{booktitle}{\emph{Active learning literature survey}}.
\newblock \bibinfo{type}{{T}echnical {R}eport}.
  \bibinfo{institution}{University of Wisconsin-Madison Department of Computer
  Sciences}.
\newblock


\bibitem[\protect\citeauthoryear{Shu, Xu, and Meng}{Shu et~al\mbox{.}}{2018}]%
        {shu2018small}
\bibfield{author}{\bibinfo{person}{J. Shu}, \bibinfo{person}{Z. Xu}, {and}
  \bibinfo{person}{D Meng}.} \bibinfo{year}{2018}\natexlab{}.
\newblock \showarticletitle{Small sample learning in big data era}.
\newblock \bibinfo{journal}{\emph{arXiv preprint arXiv:1808.04572}}
  (\bibinfo{year}{2018}).
\newblock


\bibitem[\protect\citeauthoryear{Shyam, Gupta, and Dukkipati}{Shyam
  et~al\mbox{.}}{2017}]%
        {shyam17attentive}
\bibfield{author}{\bibinfo{person}{P. Shyam}, \bibinfo{person}{S. Gupta}, {and}
  \bibinfo{person}{A. Dukkipati}.} \bibinfo{year}{2017}\natexlab{}.
\newblock \showarticletitle{Attentive recurrent comparators}. In
  \bibinfo{booktitle}{\emph{International Conference on Machine Learning}}.
  \bibinfo{pages}{3173--3181}.
\newblock


\bibitem[\protect\citeauthoryear{Silver, Huang, Maddison, Guez, Sifre, Van
  Den~Driessche, Schrittwieser, Antonoglou, Panneershelvam, Lanctot,
  et~al\mbox{.}}{Silver et~al\mbox{.}}{2016}]%
        {silver2016mastering}
\bibfield{author}{\bibinfo{person}{D. Silver}, \bibinfo{person}{A. Huang},
  \bibinfo{person}{C.~J. Maddison}, \bibinfo{person}{A. Guez},
  \bibinfo{person}{L. Sifre}, \bibinfo{person}{G. Van Den~Driessche},
  \bibinfo{person}{J. Schrittwieser}, \bibinfo{person}{I. Antonoglou},
  \bibinfo{person}{V. Panneershelvam}, \bibinfo{person}{M. Lanctot},
  {et~al\mbox{.}}} \bibinfo{year}{2016}\natexlab{}.
\newblock \showarticletitle{Mastering the game of Go with deep neural networks
  and tree search}.
\newblock \bibinfo{journal}{\emph{Nature}} \bibinfo{volume}{529},
  \bibinfo{number}{7587} (\bibinfo{year}{2016}), \bibinfo{pages}{484--489}.
\newblock


\bibitem[\protect\citeauthoryear{Snell, Swersky, and Zemel}{Snell
  et~al\mbox{.}}{2017}]%
        {snell2017prototypical}
\bibfield{author}{\bibinfo{person}{J. Snell}, \bibinfo{person}{K. Swersky},
  {and} \bibinfo{person}{R.~S. Zemel}.} \bibinfo{year}{2017}\natexlab{}.
\newblock \showarticletitle{Prototypical networks for few-shot learning}. In
  \bibinfo{booktitle}{\emph{Advances in Neural Information Processing
  Systems}}. \bibinfo{pages}{4077--4087}.
\newblock


\bibitem[\protect\citeauthoryear{Spivak}{Spivak}{1970}]%
        {spivak1970comprehensive}
\bibfield{author}{\bibinfo{person}{M.~D. Spivak}.}
  \bibinfo{year}{1970}\natexlab{}.
\newblock \bibinfo{booktitle}{\emph{A Comprehensive Introduction to
  Differential Geometry}}.
\newblock \bibinfo{publisher}{Publish or Perish}.
\newblock


\bibitem[\protect\citeauthoryear{Srivastava, Greff, and Schmidhuber}{Srivastava
  et~al\mbox{.}}{2015}]%
        {srivastava2015training}
\bibfield{author}{\bibinfo{person}{R.~K. Srivastava}, \bibinfo{person}{K.
  Greff}, {and} \bibinfo{person}{J. Schmidhuber}.}
  \bibinfo{year}{2015}\natexlab{}.
\newblock \showarticletitle{Training very deep networks}. In
  \bibinfo{booktitle}{\emph{Advances in Neural Information Processing
  Systems}}. \bibinfo{pages}{2377--2385}.
\newblock


\bibitem[\protect\citeauthoryear{Sukhbaatar, Weston, Fergus,
  et~al\mbox{.}}{Sukhbaatar et~al\mbox{.}}{2015}]%
        {sukhbaatar2015end}
\bibfield{author}{\bibinfo{person}{S. Sukhbaatar}, \bibinfo{person}{J. Weston},
  \bibinfo{person}{R. Fergus}, {et~al\mbox{.}}}
  \bibinfo{year}{2015}\natexlab{}.
\newblock \showarticletitle{End-to-end memory networks}. In
  \bibinfo{booktitle}{\emph{Advances in Neural Information Processing
  Systems}}. \bibinfo{pages}{2440--2448}.
\newblock


\bibitem[\protect\citeauthoryear{Sun, Wang, and Zong}{Sun
  et~al\mbox{.}}{2018}]%
        {sun2018memory}
\bibfield{author}{\bibinfo{person}{J. Sun}, \bibinfo{person}{S. Wang}, {and}
  \bibinfo{person}{C. Zong}.} \bibinfo{year}{2018}\natexlab{}.
\newblock \showarticletitle{Memory, show the way: Memory based few shot word
  representation learning}. In \bibinfo{booktitle}{\emph{Conference on
  Empirical Methods in Natural Language Processing}}.
  \bibinfo{pages}{1435--1444}.
\newblock


\bibitem[\protect\citeauthoryear{Sung, Yang, Zhang, Xiang, Torr, and
  Hospedales}{Sung et~al\mbox{.}}{2018}]%
        {sung2018learning}
\bibfield{author}{\bibinfo{person}{F. Sung}, \bibinfo{person}{Y. Yang},
  \bibinfo{person}{L. Zhang}, \bibinfo{person}{T. Xiang},
  \bibinfo{person}{P.~H. Torr}, {and} \bibinfo{person}{T.~M. Hospedales}.}
  \bibinfo{year}{2018}\natexlab{}.
\newblock \showarticletitle{Learning to compare: Relation network for few-shot
  learning}. In \bibinfo{booktitle}{\emph{Conference on Computer Vision and
  Pattern Recognition}}. \bibinfo{pages}{1199--1208}.
\newblock


\bibitem[\protect\citeauthoryear{Tang, Tappen, Sukthankar, and Lampert}{Tang
  et~al\mbox{.}}{2010}]%
        {tang2010optimizing}
\bibfield{author}{\bibinfo{person}{K.~D. Tang}, \bibinfo{person}{M.~F. Tappen},
  \bibinfo{person}{R. Sukthankar}, {and} \bibinfo{person}{C.~H. Lampert}.}
  \bibinfo{year}{2010}\natexlab{}.
\newblock \showarticletitle{Optimizing one-shot recognition with micro-set
  learning}. In \bibinfo{booktitle}{\emph{Conference on Computer Vision and
  Pattern Recognition}}. \bibinfo{pages}{3027--3034}.
\newblock


\bibitem[\protect\citeauthoryear{Tjandra, Sakti, and Nakamura}{Tjandra
  et~al\mbox{.}}{2018}]%
        {tjandra2018machine}
\bibfield{author}{\bibinfo{person}{A. Tjandra}, \bibinfo{person}{S. Sakti},
  {and} \bibinfo{person}{S. Nakamura}.} \bibinfo{year}{2018}\natexlab{}.
\newblock \showarticletitle{Machine speech chain with one-shot speaker
  adaptation}. In \bibinfo{booktitle}{\emph{INTERSPEECH}}.
  \bibinfo{pages}{887--891}.
\newblock


\bibitem[\protect\citeauthoryear{Torralba, Tenenbaum, and
  Salakhutdinov}{Torralba et~al\mbox{.}}{2011}]%
        {torralba2011learning}
\bibfield{author}{\bibinfo{person}{A. Torralba}, \bibinfo{person}{J.~B.
  Tenenbaum}, {and} \bibinfo{person}{R.~R. Salakhutdinov}.}
  \bibinfo{year}{2011}\natexlab{}.
\newblock \showarticletitle{Learning to learn with compound HD models}. In
  \bibinfo{booktitle}{\emph{Advances in Neural Information Processing
  Systems}}. \bibinfo{pages}{2061--2069}.
\newblock


\bibitem[\protect\citeauthoryear{Triantafillou, Zemel, and
  Urtasun}{Triantafillou et~al\mbox{.}}{2017}]%
        {triantafillou2017few}
\bibfield{author}{\bibinfo{person}{E. Triantafillou}, \bibinfo{person}{R.
  Zemel}, {and} \bibinfo{person}{R. Urtasun}.} \bibinfo{year}{2017}\natexlab{}.
\newblock \showarticletitle{Few-shot learning through an information retrieval
  lens}. In \bibinfo{booktitle}{\emph{Advances in Neural Information Processing
  Systems}}. \bibinfo{pages}{2255--2265}.
\newblock


\bibitem[\protect\citeauthoryear{Triantafillou, Zhu, Dumoulin, Lamblin, Xu,
  Goroshin, Gelada, Swersky, Manzagol, et~al\mbox{.}}{Triantafillou
  et~al\mbox{.}}{2019}]%
        {triantafillou2019meta}
\bibfield{author}{\bibinfo{person}{E. Triantafillou}, \bibinfo{person}{T. Zhu},
  \bibinfo{person}{V. Dumoulin}, \bibinfo{person}{P. Lamblin},
  \bibinfo{person}{K. Xu}, \bibinfo{person}{R. Goroshin}, \bibinfo{person}{C.
  Gelada}, \bibinfo{person}{K. Swersky}, \bibinfo{person}{P.-A. Manzagol},
  {et~al\mbox{.}}} \bibinfo{year}{2019}\natexlab{}.
\newblock \showarticletitle{Meta-dataset: A dataset of datasets for learning to
  learn from few examples}.
\newblock \bibinfo{journal}{\emph{arXiv preprint arXiv:1903.03096}}
  (\bibinfo{year}{2019}).
\newblock


\bibitem[\protect\citeauthoryear{Tsai, Huang, and Salakhutdinov}{Tsai
  et~al\mbox{.}}{2017}]%
        {hubert2017learning}
\bibfield{author}{\bibinfo{person}{Y.-H. Tsai}, \bibinfo{person}{L.-K. Huang},
  {and} \bibinfo{person}{R. Salakhutdinov}.} \bibinfo{year}{2017}\natexlab{}.
\newblock \showarticletitle{Learning robust visual-semantic embeddings}. In
  \bibinfo{booktitle}{\emph{Conference on Computer Vision and Pattern
  Recognition}}. \bibinfo{pages}{3571--3580}.
\newblock


\bibitem[\protect\citeauthoryear{Tsai and Salakhutdinov}{Tsai and
  Salakhutdinov}{2017}]%
        {tsai2017improving}
\bibfield{author}{\bibinfo{person}{Y.~H. Tsai} {and} \bibinfo{person}{R.
  Salakhutdinov}.} \bibinfo{year}{2017}\natexlab{}.
\newblock \showarticletitle{Improving one-shot learning through fusing side
  information}.
\newblock \bibinfo{journal}{\emph{arXiv preprint arXiv:1710.08347}}
  (\bibinfo{year}{2017}).
\newblock


\bibitem[\protect\citeauthoryear{Turing}{Turing}{1950}]%
        {turing1950computing}
\bibfield{author}{\bibinfo{person}{M.~A. Turing}.}
  \bibinfo{year}{1950}\natexlab{}.
\newblock \showarticletitle{Computing machinery and intelligence}.
\newblock \bibinfo{journal}{\emph{Mind}} \bibinfo{volume}{59},
  \bibinfo{number}{236} (\bibinfo{year}{1950}), \bibinfo{pages}{433--433}.
\newblock


\bibitem[\protect\citeauthoryear{Van~den Oord, Kalchbrenner, Espeholt, Vinyals,
  Graves, et~al\mbox{.}}{Van~den Oord et~al\mbox{.}}{2016}]%
        {van2016conditional}
\bibfield{author}{\bibinfo{person}{A. Van~den Oord}, \bibinfo{person}{N.
  Kalchbrenner}, \bibinfo{person}{L. Espeholt}, \bibinfo{person}{O. Vinyals},
  \bibinfo{person}{A. Graves}, {et~al\mbox{.}}}
  \bibinfo{year}{2016}\natexlab{}.
\newblock \showarticletitle{Conditional image generation with {P}ixel{CNN}
  decoders}. In \bibinfo{booktitle}{\emph{Advances in Neural Information
  Processing Systems}}. \bibinfo{pages}{4790--4798}.
\newblock


\bibitem[\protect\citeauthoryear{Vapnik}{Vapnik}{1992}]%
        {vapnik1992principles}
\bibfield{author}{\bibinfo{person}{V.~N. Vapnik}.}
  \bibinfo{year}{1992}\natexlab{}.
\newblock \showarticletitle{Principles of risk minimization for learning
  theory}. In \bibinfo{booktitle}{\emph{Advances in Neural Information
  Processing Systems}}. \bibinfo{pages}{831--838}.
\newblock


\bibitem[\protect\citeauthoryear{Vartak, Thiagarajan, Miranda, Bratman, and
  Larochelle}{Vartak et~al\mbox{.}}{2017}]%
        {vartak2017meta}
\bibfield{author}{\bibinfo{person}{M. Vartak}, \bibinfo{person}{A.
  Thiagarajan}, \bibinfo{person}{C. Miranda}, \bibinfo{person}{J. Bratman},
  {and} \bibinfo{person}{H. Larochelle}.} \bibinfo{year}{2017}\natexlab{}.
\newblock \showarticletitle{A meta-learning perspective on cold-start
  recommendations for items}. In \bibinfo{booktitle}{\emph{Advances in Neural
  Information Processing Systems}}. \bibinfo{pages}{6904--6914}.
\newblock


\bibitem[\protect\citeauthoryear{Vinyals, Blundell, Lillicrap, Wierstra,
  et~al\mbox{.}}{Vinyals et~al\mbox{.}}{2016}]%
        {vinyals2016matching}
\bibfield{author}{\bibinfo{person}{O. Vinyals}, \bibinfo{person}{C. Blundell},
  \bibinfo{person}{T. Lillicrap}, \bibinfo{person}{D. Wierstra},
  {et~al\mbox{.}}} \bibinfo{year}{2016}\natexlab{}.
\newblock \showarticletitle{Matching networks for one shot learning}. In
  \bibinfo{booktitle}{\emph{Advances in Neural Information Processing
  Systems}}. \bibinfo{pages}{3630--3638}.
\newblock


\bibitem[\protect\citeauthoryear{Wang, Liu, Shen, Huang, van~den Hengel, and
  Tao~Shen}{Wang et~al\mbox{.}}{2017}]%
        {wang2017multi}
\bibfield{author}{\bibinfo{person}{P. Wang}, \bibinfo{person}{L. Liu},
  \bibinfo{person}{C. Shen}, \bibinfo{person}{Z. Huang}, \bibinfo{person}{A.
  van~den Hengel}, {and} \bibinfo{person}{H. Tao~Shen}.}
  \bibinfo{year}{2017}\natexlab{}.
\newblock \showarticletitle{Multi-attention network for one shot learning}. In
  \bibinfo{booktitle}{\emph{Conference on Computer Vision and Pattern
  Recognition}}. \bibinfo{pages}{2721--2729}.
\newblock


\bibitem[\protect\citeauthoryear{Wang, Ye, and Gupta}{Wang
  et~al\mbox{.}}{2018b}]%
        {wang2018zero}
\bibfield{author}{\bibinfo{person}{X. Wang}, \bibinfo{person}{Y. Ye}, {and}
  \bibinfo{person}{A. Gupta}.} \bibinfo{year}{2018}\natexlab{b}.
\newblock \showarticletitle{Zero-shot recognition via semantic embeddings and
  knowledge graphs}. In \bibinfo{booktitle}{\emph{Conference on Computer Vision
  and Pattern Recognition}}. \bibinfo{pages}{6857--6866}.
\newblock


\bibitem[\protect\citeauthoryear{Wang, Girshick, Hebert, and Hariharan}{Wang
  et~al\mbox{.}}{2018a}]%
        {wang2018low}
\bibfield{author}{\bibinfo{person}{Y.-X. Wang}, \bibinfo{person}{R. Girshick},
  \bibinfo{person}{M. Hebert}, {and} \bibinfo{person}{B. Hariharan}.}
  \bibinfo{year}{2018}\natexlab{a}.
\newblock \showarticletitle{Low-shot learning from imaginary data}. In
  \bibinfo{booktitle}{\emph{Conference on Computer Vision and Pattern
  Recognition}}. \bibinfo{pages}{7278--7286}.
\newblock


\bibitem[\protect\citeauthoryear{Wang and Hebert}{Wang and Hebert}{2016a}]%
        {wang2016learningc}
\bibfield{author}{\bibinfo{person}{Y.-X. Wang} {and} \bibinfo{person}{M.
  Hebert}.} \bibinfo{year}{2016}\natexlab{a}.
\newblock \showarticletitle{Learning from small sample sets by combining
  unsupervised meta-training with CNNs}. In \bibinfo{booktitle}{\emph{Advances
  in Neural Information Processing Systems}}. \bibinfo{pages}{244--252}.
\newblock


\bibitem[\protect\citeauthoryear{Wang and Hebert}{Wang and Hebert}{2016b}]%
        {wang2016learninga}
\bibfield{author}{\bibinfo{person}{Y.-X. Wang} {and} \bibinfo{person}{M.
  Hebert}.} \bibinfo{year}{2016}\natexlab{b}.
\newblock \showarticletitle{Learning to learn: Model regression networks for
  easy small sample learning}. In \bibinfo{booktitle}{\emph{European Conference
  on Computer Vision}}. \bibinfo{pages}{616--634}.
\newblock


\bibitem[\protect\citeauthoryear{Wei and Zou}{Wei and Zou}{2019}]%
        {wei2019eda}
\bibfield{author}{\bibinfo{person}{J. Wei} {and} \bibinfo{person}{K. Zou}.}
  \bibinfo{year}{2019}\natexlab{}.
\newblock \showarticletitle{EDA: Easy data augmentation techniques for boosting
  performance on text classification tasks}. In
  \bibinfo{booktitle}{\emph{Conference on Empirical Methods in Natural Language
  Processing and International Joint Conference on Natural Language
  Processing}}. \bibinfo{pages}{6383--6389}.
\newblock


\bibitem[\protect\citeauthoryear{Weston, Chopra, and Bordes}{Weston
  et~al\mbox{.}}{2014}]%
        {weston2014memory}
\bibfield{author}{\bibinfo{person}{J. Weston}, \bibinfo{person}{S. Chopra},
  {and} \bibinfo{person}{A. Bordes}.} \bibinfo{year}{2014}\natexlab{}.
\newblock \showarticletitle{Memory networks}.
\newblock \bibinfo{journal}{\emph{arXiv preprint arXiv:1410.3916}}
  (\bibinfo{year}{2014}).
\newblock


\bibitem[\protect\citeauthoryear{Woodward and Finn}{Woodward and Finn}{2017}]%
        {woodward2017active}
\bibfield{author}{\bibinfo{person}{M. Woodward} {and} \bibinfo{person}{C.
  Finn}.} \bibinfo{year}{2017}\natexlab{}.
\newblock \showarticletitle{Active one-shot learning}.
\newblock \bibinfo{journal}{\emph{arXiv preprint arXiv:1702.06559}}
  (\bibinfo{year}{2017}).
\newblock


\bibitem[\protect\citeauthoryear{Wu and Demiris}{Wu and Demiris}{2010}]%
        {wu2010towards}
\bibfield{author}{\bibinfo{person}{Y. Wu} {and} \bibinfo{person}{Y. Demiris}.}
  \bibinfo{year}{2010}\natexlab{}.
\newblock \showarticletitle{Towards one shot learning by imitation for humanoid
  robots}. In \bibinfo{booktitle}{\emph{International Conference on Robotics
  and Automation}}. \bibinfo{pages}{2889--2894}.
\newblock


\bibitem[\protect\citeauthoryear{Wu, Lin, Dong, Yan, Ouyang, and Yang}{Wu
  et~al\mbox{.}}{2018}]%
        {wu2018exploit}
\bibfield{author}{\bibinfo{person}{Y. Wu}, \bibinfo{person}{Y. Lin},
  \bibinfo{person}{X. Dong}, \bibinfo{person}{Y. Yan}, \bibinfo{person}{W.
  Ouyang}, {and} \bibinfo{person}{Y. Yang}.} \bibinfo{year}{2018}\natexlab{}.
\newblock \showarticletitle{Exploit the unknown gradually: One-shot video-based
  person re-identification by stepwise learning}. In
  \bibinfo{booktitle}{\emph{Conference on Computer Vision and Pattern
  Recognition}}. \bibinfo{pages}{5177--5186}.
\newblock


\bibitem[\protect\citeauthoryear{Xu, Zhu, and Yang}{Xu et~al\mbox{.}}{2017}]%
        {xu2017few}
\bibfield{author}{\bibinfo{person}{Z. Xu}, \bibinfo{person}{L. Zhu}, {and}
  \bibinfo{person}{Y. Yang}.} \bibinfo{year}{2017}\natexlab{}.
\newblock \showarticletitle{Few-shot object recognition from machine-labeled
  web images}. In \bibinfo{booktitle}{\emph{Conference on Computer Vision and
  Pattern Recognition}}. \bibinfo{pages}{1164--1172}.
\newblock


\bibitem[\protect\citeauthoryear{Yan, Zheng, and Cao}{Yan
  et~al\mbox{.}}{2018}]%
        {yan2018few}
\bibfield{author}{\bibinfo{person}{L. Yan}, \bibinfo{person}{Y. Zheng}, {and}
  \bibinfo{person}{J. Cao}.} \bibinfo{year}{2018}\natexlab{}.
\newblock \showarticletitle{Few-shot learning for short text classification}.
\newblock \bibinfo{journal}{\emph{Multimedia Tools and Applications}}
  \bibinfo{volume}{77}, \bibinfo{number}{22} (\bibinfo{year}{2018}),
  \bibinfo{pages}{29799--29810}.
\newblock


\bibitem[\protect\citeauthoryear{Yan, Yap, and Mori}{Yan et~al\mbox{.}}{2015}]%
        {yan2015multi}
\bibfield{author}{\bibinfo{person}{W. Yan}, \bibinfo{person}{J. Yap}, {and}
  \bibinfo{person}{G. Mori}.} \bibinfo{year}{2015}\natexlab{}.
\newblock \showarticletitle{Multi-task transfer methods to improve one-shot
  learning for multimedia event detection}. In
  \bibinfo{booktitle}{\emph{British Machine Vision Conference}}.
\newblock


\bibitem[\protect\citeauthoryear{Yang, He, and Porikli}{Yang
  et~al\mbox{.}}{2018}]%
        {yang2018one}
\bibfield{author}{\bibinfo{person}{H. Yang}, \bibinfo{person}{X. He}, {and}
  \bibinfo{person}{F. Porikli}.} \bibinfo{year}{2018}\natexlab{}.
\newblock \showarticletitle{One-shot action localization by learning sequence
  matching network}. In \bibinfo{booktitle}{\emph{Conference on Computer Vision
  and Pattern Recognition}}. \bibinfo{pages}{1450--1459}.
\newblock


\bibitem[\protect\citeauthoryear{Yao, Wang, Jair, Guyon, Hu, Li, Tu, Yang, and
  Yu}{Yao et~al\mbox{.}}{2018}]%
        {quanming2018taking}
\bibfield{author}{\bibinfo{person}{Q. Yao}, \bibinfo{person}{M. Wang},
  \bibinfo{person}{E.~H. Jair}, \bibinfo{person}{I. Guyon},
  \bibinfo{person}{Y.-Q. Hu}, \bibinfo{person}{Y.-F. Li},
  \bibinfo{person}{W.-W. Tu}, \bibinfo{person}{Q. Yang}, {and}
  \bibinfo{person}{Y. Yu}.} \bibinfo{year}{2018}\natexlab{}.
\newblock \showarticletitle{Taking human out of learning applications: A survey
  on automated machine learning}.
\newblock \bibinfo{journal}{\emph{arXiv preprint arXiv:1810.13306}}
  (\bibinfo{year}{2018}).
\newblock


\bibitem[\protect\citeauthoryear{Yao, Xu, Tu, and Zhu}{Yao
  et~al\mbox{.}}{2020}]%
        {yao2020efficient}
\bibfield{author}{\bibinfo{person}{Q. Yao}, \bibinfo{person}{J. Xu},
  \bibinfo{person}{W.-W. Tu}, {and} \bibinfo{person}{Z. Zhu}.}
  \bibinfo{year}{2020}\natexlab{}.
\newblock \showarticletitle{Efficient neural architecture search via proximal
  iterations}. In \bibinfo{booktitle}{\emph{AAAI Conference on Artificial
  Intelligence}}.
\newblock


\bibitem[\protect\citeauthoryear{Yoo, Fan, Boddeti, and Kitani}{Yoo
  et~al\mbox{.}}{2018}]%
        {yoo2018efficient}
\bibfield{author}{\bibinfo{person}{D. Yoo}, \bibinfo{person}{H. Fan},
  \bibinfo{person}{V.~N. Boddeti}, {and} \bibinfo{person}{K.~M. Kitani}.}
  \bibinfo{year}{2018}\natexlab{}.
\newblock \showarticletitle{Efficient k-shot learning with regularized deep
  networks}. In \bibinfo{booktitle}{\emph{AAAI Conference on Artificial
  Intelligence}}.
\newblock


\bibitem[\protect\citeauthoryear{Yoon, Kim, Dia, Kim, Bengio, and Ahn}{Yoon
  et~al\mbox{.}}{2018}]%
        {yoon2018bayesian}
\bibfield{author}{\bibinfo{person}{J. Yoon}, \bibinfo{person}{T. Kim},
  \bibinfo{person}{O. Dia}, \bibinfo{person}{S. Kim}, \bibinfo{person}{Y.
  Bengio}, {and} \bibinfo{person}{S. Ahn}.} \bibinfo{year}{2018}\natexlab{}.
\newblock \showarticletitle{Bayesian model-agnostic meta-learning}. In
  \bibinfo{booktitle}{\emph{Advances in Neural Information Processing
  Systems}}. \bibinfo{pages}{7343--7353}.
\newblock


\bibitem[\protect\citeauthoryear{Yu, Guo, Yi, Chang, Potdar, Cheng, Tesauro,
  Wang, and Zhou}{Yu et~al\mbox{.}}{2018}]%
        {yu2018diverse}
\bibfield{author}{\bibinfo{person}{M. Yu}, \bibinfo{person}{X. Guo},
  \bibinfo{person}{J. Yi}, \bibinfo{person}{S. Chang}, \bibinfo{person}{S.
  Potdar}, \bibinfo{person}{Y. Cheng}, \bibinfo{person}{G. Tesauro},
  \bibinfo{person}{H. Wang}, {and} \bibinfo{person}{B. Zhou}.}
  \bibinfo{year}{2018}\natexlab{}.
\newblock \showarticletitle{Diverse few-shot text classification with multiple
  metrics}. In \bibinfo{booktitle}{\emph{Conference of the North American
  Chapter of the Association for Computational Linguistics: Human Language
  Technologies}}. \bibinfo{pages}{1206--1215}.
\newblock


\bibitem[\protect\citeauthoryear{Zhang, Butepage, Kjellstrom, and Mandt}{Zhang
  et~al\mbox{.}}{2019}]%
        {zhang2018advances}
\bibfield{author}{\bibinfo{person}{C. Zhang}, \bibinfo{person}{J. Butepage},
  \bibinfo{person}{H. Kjellstrom}, {and} \bibinfo{person}{S. Mandt}.}
  \bibinfo{year}{2019}\natexlab{}.
\newblock \showarticletitle{Advances in variational inference}.
\newblock \bibinfo{journal}{\emph{IEEE Transactions on Pattern Analysis and
  Machine Intelligence}} \bibinfo{volume}{41}, \bibinfo{number}{8}
  (\bibinfo{year}{2019}), \bibinfo{pages}{2008--2026}.
\newblock


\bibitem[\protect\citeauthoryear{Zhang, Che, Ghahramani, Bengio, and
  Song}{Zhang et~al\mbox{.}}{2018a}]%
        {zhang2018metagan}
\bibfield{author}{\bibinfo{person}{R. Zhang}, \bibinfo{person}{T. Che},
  \bibinfo{person}{Z. Ghahramani}, \bibinfo{person}{Y. Bengio}, {and}
  \bibinfo{person}{Y. Song}.} \bibinfo{year}{2018}\natexlab{a}.
\newblock \showarticletitle{MetaGAN: An adversarial approach to few-shot
  learning}. In \bibinfo{booktitle}{\emph{Advances in Neural Information
  Processing Systems}}. \bibinfo{pages}{2371--2380}.
\newblock


\bibitem[\protect\citeauthoryear{Zhang, Tang, and Jia}{Zhang
  et~al\mbox{.}}{2018b}]%
        {zhang2018fine}
\bibfield{author}{\bibinfo{person}{Y. Zhang}, \bibinfo{person}{H. Tang}, {and}
  \bibinfo{person}{K. Jia}.} \bibinfo{year}{2018}\natexlab{b}.
\newblock \showarticletitle{Fine-grained visual categorization using
  meta-learning optimization with sample selection of auxiliary data}. In
  \bibinfo{booktitle}{\emph{European Conference on Computer Vision}}.
  \bibinfo{pages}{233--248}.
\newblock


\bibitem[\protect\citeauthoryear{Zhang and Yang}{Zhang and Yang}{2017}]%
        {zhang2017survey}
\bibfield{author}{\bibinfo{person}{Y. Zhang} {and} \bibinfo{person}{Q. Yang}.}
  \bibinfo{year}{2017}\natexlab{}.
\newblock \showarticletitle{A survey on multi-task learning}.
\newblock \bibinfo{journal}{\emph{arXiv preprint arXiv:1707.08114}}
  (\bibinfo{year}{2017}).
\newblock


\bibitem[\protect\citeauthoryear{Zhao, Zhao, Yan, and Feng}{Zhao
  et~al\mbox{.}}{2018}]%
        {zhao2018dynamic}
\bibfield{author}{\bibinfo{person}{F. Zhao}, \bibinfo{person}{J. Zhao},
  \bibinfo{person}{S. Yan}, {and} \bibinfo{person}{J. Feng}.}
  \bibinfo{year}{2018}\natexlab{}.
\newblock \showarticletitle{Dynamic conditional networks for few-shot
  learning}. In \bibinfo{booktitle}{\emph{European Conference on Computer
  Vision}}.
\newblock


\bibitem[\protect\citeauthoryear{Zhou}{Zhou}{2017}]%
        {zhou2017brief}
\bibfield{author}{\bibinfo{person}{Z.-H. Zhou}.}
  \bibinfo{year}{2017}\natexlab{}.
\newblock \showarticletitle{A brief introduction to weakly supervised
  learning}.
\newblock \bibinfo{journal}{\emph{National Science Review}}
  \bibinfo{volume}{5}, \bibinfo{number}{1} (\bibinfo{year}{2017}),
  \bibinfo{pages}{44--53}.
\newblock


\bibitem[\protect\citeauthoryear{Zhu and Yang}{Zhu and Yang}{2018}]%
        {zhu2018compound}
\bibfield{author}{\bibinfo{person}{L. Zhu} {and} \bibinfo{person}{Y. Yang}.}
  \bibinfo{year}{2018}\natexlab{}.
\newblock \showarticletitle{Compound memory networks for few-shot video
  classification}. In \bibinfo{booktitle}{\emph{European Conference on Computer
  Vision}}. \bibinfo{pages}{751--766}.
\newblock


\bibitem[\protect\citeauthoryear{Zhu}{Zhu}{2005}]%
        {zhu2005semi}
\bibfield{author}{\bibinfo{person}{X.~J. Zhu}.}
  \bibinfo{year}{2005}\natexlab{}.
\newblock \bibinfo{booktitle}{\emph{Semi-supervised learning literature
  survey}}.
\newblock \bibinfo{type}{{T}echnical {R}eport}.
  \bibinfo{institution}{University of Wisconsin-Madison Department of Computer
  Sciences}.
\newblock


\bibitem[\protect\citeauthoryear{Zoph and Le}{Zoph and Le}{2017}]%
        {zoph2017neural}
\bibfield{author}{\bibinfo{person}{B. Zoph} {and} \bibinfo{person}{Q.~V. Le}.}
  \bibinfo{year}{2017}\natexlab{}.
\newblock \showarticletitle{Neural architecture search with reinforcement
  learning}. In \bibinfo{booktitle}{\emph{International Conference on Learning
  Representations}}.
\newblock


\end{thebibliography}

\end{document}